\documentclass[nohyperref]{article}

\usepackage{microtype}
\usepackage{graphicx}
\usepackage{subfigure}
\usepackage{booktabs} 

\usepackage{hyperref}

\usepackage[accepted]{icml2023}

\usepackage{amsmath}
\usepackage{amssymb}
\usepackage{mathtools}
\usepackage{amsthm}
\usepackage{array}

\usepackage[capitalize,noabbrev]{cleveref}

\theoremstyle{plain}

\theoremstyle{definition}

\theoremstyle{remark}

\usepackage[textsize=tiny]{todonotes}

\icmltitlerunning{L-SA: Learning Under-Explored Targets in Multi-Target Reinforcement Learning}

\begin{document}

\twocolumn[
\icmltitle{L-SA: Learning Under-Explored Targets\\in Multi-Target Reinforcement Learning}



\icmlsetsymbol{equal}{*}

\begin{icmlauthorlist}
\icmlauthor{Kibeom Kim}{yyy,comp}
\icmlauthor{Hyundo Lee}{yyy}
\icmlauthor{Min Whoo Lee}{yyy}
\icmlauthor{Moonheon Lee}{yyy}
\icmlauthor{Minsu Lee}{equal,yyy}
\icmlauthor{Byoung-Tak Zhang}{equal,yyy}

\end{icmlauthorlist}

\icmlaffiliation{yyy}{AI Institute, Seoul National University}
\icmlaffiliation{comp}{Surromind}

\icmlcorrespondingauthor{Kibeom Kim}{kbkim@bi.snu.ac.kr}

\icmlkeywords{Machine Learning, ICML}

\vskip 0.3in
]



\printAffiliationsAndNotice{}  

\begin{abstract}
Tasks that involve interaction with various targets are called multi-target tasks. 
When applying general reinforcement learning approaches for such tasks, certain targets that are difficult to access or interact with may be neglected throughout the course of training – a predicament we call Under-explored Target Problem (UTP). 
To address this problem, we propose L-SA (Learning by adaptive Sampling and Active querying) framework that includes adaptive sampling and active querying.
In the L-SA framework, adaptive sampling dynamically samples targets with the highest increase of success rates at a high proportion, resulting in curricular learning from easy to hard targets.
Active querying prompts the agent to interact more frequently with under-explored targets that need more experience or exploration.
Our experimental results on visual navigation tasks show that the L-SA framework improves sample efficiency as well as success rates on various multi-target tasks with UTP.
Also, it is experimentally demonstrated that the cyclic relationship between adaptive sampling and active querying effectively improves the sample richness of under-explored targets and alleviates UTP.
\end{abstract}

\section{Introduction}
\label{intro}
Research in reinforcement learning has been striving to approach the ever-growing demand for human support. In particular, many human service applications in the real world are multi-target tasks \cite{kim2021goal}, where the agent must execute the given instructions regarding various targets, such as objects or destinations. 
Typical examples of such instructions include “Fetch me a tumbler” and “Bring me a tennis ball”, as well as many others that revolve around humans’ daily lives.

\begin{figure}[t]
\begin{center}
\centerline{\includegraphics[width=\columnwidth]{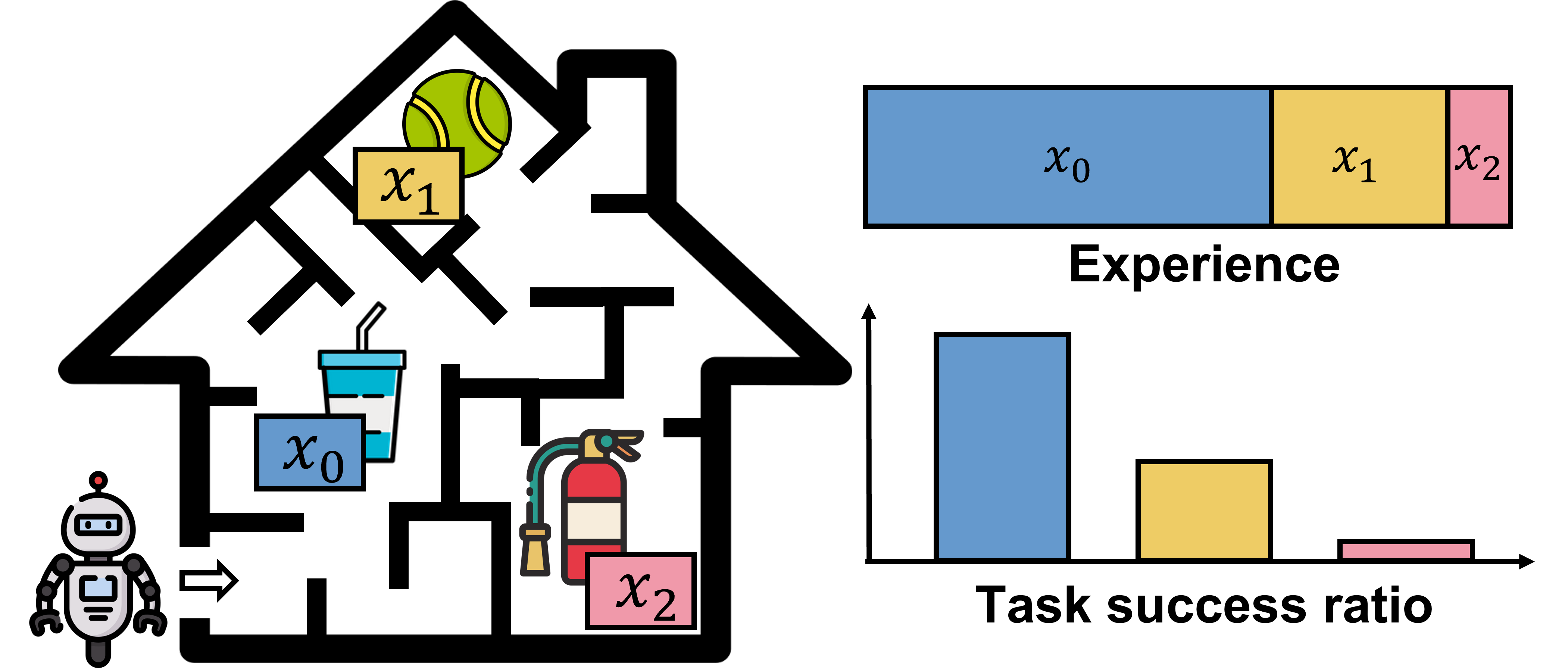}}
\caption{Example environment that illustrates Under-explored Target Problem. Suppose a task has easy targets (e.g. tumbler and ball) that are generated close to the agent’s initial location and a hard target (e.g. fire extinguisher) that appears far away from the agent’s location. Due to the high success rates of easy targets and the rarity of successful trajectories with the hard target, the latter is barely learned.}
\label{fig:UTP}
\end{center}
\end{figure}

As an important challenge to be dealt with in multi-target tasks, we bring attention to the Under-explored Target Problem (UTP), which we formulate in this study.
In the real world, visual target search for rare items among various targets is recognized as a difficult problem even for humans \cite{wolfe2005rare,mitroff2014ultra}. 
Similar to the visual target search problem, the UTP occurs when the difficulties of accessing or interacting with respective targets vary substantially.
As shown in Figure~\ref{fig:UTP}, when easy targets ($x_0$, $x_1$) and a difficult target ($x_2$) coexist, the agent’s successful experiences may likely be dominated by experiences corresponding to the easier targets.
Consequently, the more difficult target is under-explored, leading to infrequent and insufficient learning of, or even a practical exclusion of, interaction with such target.
The multi-target task is a type of multi-task learning, but studies on multi-target RL are almost rare.
Studies for multi-task \cite{sharma2017learning,liu2020adaptive,yao2021meta} propose a method for dynamically scheduling multi-tasks.
These studies require a reference score for tasks, or the method of measuring task difficulty is specialized in each field.
For this reason, the application as a multi-target task is limited, or prior knowledge is required. 
Additionally, in the case of curriculum learning \cite{florensa2017reverse,fang2019curriculum,zhang2020automatic,sharma2021autonomous} or goal-relabeling methods \cite{andrychowicz2017hindsight,fang2018dher,fang2019curriculum} try to tackle highly difficult tasks. 
However, these methods assume that the goal or initial states can be adjusted, and do not consider learning about tasks with various difficulties simultaneously.

Studies on the visual target search for rare items propose to alleviate the problem by using repeated experiences \cite{biggs2014rare,mitroff2014ultra} or by increasing the frequency \cite{hout2015failures} on under-explored targets.
Our study draws inspiration from these prior works.

To address UTP, we propose a Learning by adaptive Sampling and Active querying (L-SA) framework with a cyclic relationship between adaptive sampling and active querying. 
Through trial and error, the goal states are collected in the goal storage based on the success reward. 
For auxiliary representation learning for policy learning, the targets with the highest increase in success rate are adaptively sampled from the goal storage at a high proportion. 
Based on the storage data distribution, active querying prompts the agent to pursue targets that need more trial and error frequently. 
Our framework forms the cyclic structure that actively queries targets learned insufficiently, collects goal states in goal storage, and learns representation through adaptive sampling from the storage.
We evaluate our framework in multi-target tasks with UTP, showing that L-SA significantly outperforms competitive methods in terms of success rate and sample efficiency.
Additionally, we formulate and investigate the sample richness of each target to measure the sufficiency of collected data to be sampled for training.
The sample richness analysis confirmed that our method is suitable for solving UTP.

Our contributions are the following:
1) UTP: We formulate the Under-explored Target Problem, where hard targets tend to be excluded from learning in multi-target tasks that require learning targets of easy and hard difficulty together.
2) L-SA framework: We propose an L-SA framework with a virtuous cycle mechanism with adaptive sampling and active querying.
Our framework does not require additional learnable parameters or prior knowledge.
3) Our proposed framework shows the state-of-the-art success rate on various multi-target tasks with UTP and improves sample efficiency. 
We demonstrate the effectiveness of adaptive sampling and active querying and show that L-SA improves the sample richness for under-explored targets with a virtuous cycle.

\section{Related Work}

\subsection{Active Learning for Multi-Task}
Multi-task learning \cite{caruana1997multitask,liu2020adaptive,crawshaw2020multi,zhang2021survey} simultaneously improves the performance over multiple tasks using a single shared network.
Sharing learned representations of related tasks enables knowledge transfer and bolsters computational efficiency.
However, if the difference in difficulty between multiple tasks is large, the learning may be inefficient or fail to converge \cite{crawshaw2020multi,zhang2021survey}.

To overcome this problem, task scheduling methods \cite{sharma2017learning,liu2020adaptive,yao2021meta,matsumoto2022robust} have been proposed as active learning, and optimized task scheduling can improve performance \cite{bengio2009curriculum}.
These studies suggest ways to schedule based on reference scores \cite{sharma2017learning} or loss-based methods \cite{yao2021meta} for meta-learning.
These studies have limitations in that they require a reference score in advance or extra parameters.
In contrast, our proposed method does not need additional parameters and prior knowledge, since it performs adaptive sampling and active queries based on samples through trial and error.

\subsection{Multi-Target Task}
There are a variety of studies in RL that aim to solve tasks for diverse goals or targets.
First of all, there are studies on multi-target tasks where the instruction specifies an object or an indoor room to navigate to \cite{savva2017minos,anderson2018evaluation,wu2018building,chaplot2020object,kim2021goal}.
To solve these tasks, methods that learn rewarding or goal states have been proposed, but these methods merely recognize such rewarding states, rather than handling situations where the agent shows fewer successes with certain targets.

In addition, there have been studies that use object detection \cite{cheng2018reinforcement}, build maps \cite{chaplot2020object,emmons2020sparse}, or conduct scene-driven navigation \cite{zhu2017target,mousavian2019visual,devo2020towards}, but these studies require prior knowledge of the tasks.
Unlike previous methods, our method adapts using goal states from experience, thus tackling the UTP without prior knowledge.

Similarly, research related to subgoals \cite{florensa2018automatic,sharma2021autonomous,chane2021goal} or intermediate goals \cite{sukhbaatar2018intrinsic} is also being actively conducted.
These tasks serve as gateways or help the agent learn before reaching the ultimate goal.
It is difficult to apply these studies to our research since the targets are spawned at random locations and the agent receives a sparse reward in our experiments.

Multi-goal RL tasks \cite{dhiman2018learning,plappert2018multi,zhao2019maximum,pitis2020maximum} require interaction with given non-characteristic goals such as coordinates.
Unlike these studies, multi-target tasks of our scope prompt interaction with visually characteristic targets so it is necessary to learn representation of each target.  

\section{Preliminaries}
\label{sec:pre}
\subsection{Multi-Target Reinforcement Learning}
We define multi-target MDP as a tuple  $\left(\mathcal{S}, \mathcal{A}, \mathcal{R}, \mathcal{P}, \gamma, \mathcal{I}\right)$, where $\mathcal{S}$ denotes the state space, $\mathcal{A}$ the action space, $\mathcal{R}$ the reward function, $\mathcal{P}$ the transition probability function, and $\gamma \in [0,1)$ the discount factor.
$\mathcal{I}$ denotes ``instruction set'' $\mathcal{I} = \{I^{1}, I^{2}, \dots, I^{N}\}$ for $N$ targets. 
When an instruction $I^x$ is given for each episode (where $x \in \{1, 2, ..., N\}$), the reward function is conditioned on the target $x$ to reward the agent for reaching $x$, while the transition probability function remains unchanged.
State-value function $V(s^x_t) = \mathbb{E}[R_{t}| s^x_t]$, where $s^x_t = (s_t, I^x)$ is state $s_t$ with time $t$ conditioned on $x$.
$R_t$ denotes the sum of decayed rewards from time step $t$ to terminal step $T$.

For the base reinforcement learning algorithm, we use A3C \cite{mnih2016asynchronous}.
In this method, the policy gradient for the actor function and the loss gradient for the critic function are defined as Eq. \ref{eq:actor} and \ref{eq:critic} respectively:
\begin{align}
    &\nabla_\theta \mathcal{L}_{RL} = - \nabla_\theta \log \pi_\theta (a_t|s^x_t)(R_t-V_\phi(s^x_t))\label{eq:actor}\\&\qquad \qquad \qquad - \beta \nabla_\theta H(\pi_\theta(\cdot|s^x_t)) \nonumber\\
    &\nabla_\phi \mathcal{L}_{RL} = \nabla_\phi (R_t-V_\phi(s^x_t))^2 \label{eq:critic} 
\end{align}
where $H(\cdot)$ and $\beta$ denote the entropy term and its coefficient respectively.
Overall, the loss function $\mathcal{L}_{RL}$ is minimized to update the actor and the critic of the RL agent.

There are various studies on multi-target reinforcement learning. 
Among them, \citet{kim2021goal} proposes representation learning for discriminating targets through auxiliary learning.
Upon success, the goal state is collected, labeled as the given instruction, and stored in \textit{goal storage}. The goal storage is additionally explained in Appendix~\ref{app:l-sa}.
RGB-D observation is provided as input to the feature extractor to obtain the encoding features.
Then, the agent action is obtained by policy with the features as the input.
This results in the RL agent loss $\mathcal{L}_{RL}$ based on environmental reward.

\subsection{Representation Learning}

For representation learning of targets, we apply SupCon \cite{khosla2020supervised}.
SupCon is a supervised contrastive learning method that uses the same-labeled data as positive pairs and others as negative pairs.
In our case, given $\langle$goal state, instruction$\rangle$ pairs from the goal storage, we treat the instructions as the labels for the goal states.
With these positive and negative pairs, we train the feature extractor with the SupCon loss function $\mathcal{L}_S$, defined below:
\begin{equation}
\label{eq:supcon}
\mathcal{L}_{S} = \sum_{j \in J} {-1 \over |P(j)|}\sum_{p \in P(j)} \log {\exp(g_j \cdot g_p / \tau_s)\over \sum\limits_{h \in J\setminus\{j\}} \exp(g_j \cdot g_h/\tau_s)}
\end{equation}
where $J$ is the set of indices of goal states in the batch, $P(j)$ is the set of all positive pair indices corresponding to the $j$-th goal state (where $j \notin P(j)$),
$|P(j)|$ is the cardinality of $P(j)$, $g_j$ is the output of the feature extractor for $j$-th goal state, and $\tau_s$ is temperature as a hyperparameter.

On top of the RL loss, auxiliary learning is conducted based on SupCon loss $\mathcal{L}_{S}$, for representation learning of different targets.
The final loss function is $\mathcal{L}_{total} = \mathcal{L}_{RL} + \eta \mathcal{L}_{S}$, with coefficient $\eta$ as a hyperparameter.


\subsection{Task Definition and Under-Explored Targets}
For multi-target tasks, we conduct experiments in the visual navigation domain.
In every episode, the initial location of the agent is set to the center of the map.
Various targets are created at random locations, and instruction $I^x$ is given randomly.
A success reward is given when the target appropriate for the instruction is reached; otherwise, the agent receives a timeout penalty upon exceeding the maximum time step $T$ or a failure penalty upon reaching a non-instructed target.

Empirically, we find that the random agent in our ``\textbf{Studio-2N 2H}'' map (in Sec.~\ref{sec:exp}) achieves 7-8\% success rate for normal-difficulty targets, while it attains as little as 0.2\% for more difficult targets.
When there is such an extreme difference in success rates, successful trajectories or states are rarely collected for the hard-difficulty targets, making it virtually excluded from learning.
We refer to the problem caused by the hard-difficulty targets that are excluded from learning as the under-explored target problem.
It is assumed that the difficulty of each target is not known in advance, and all targets are reachable by the agent.

\begin{figure}[t]
\begin{center}
\centerline{\includegraphics[width=0.95\columnwidth]{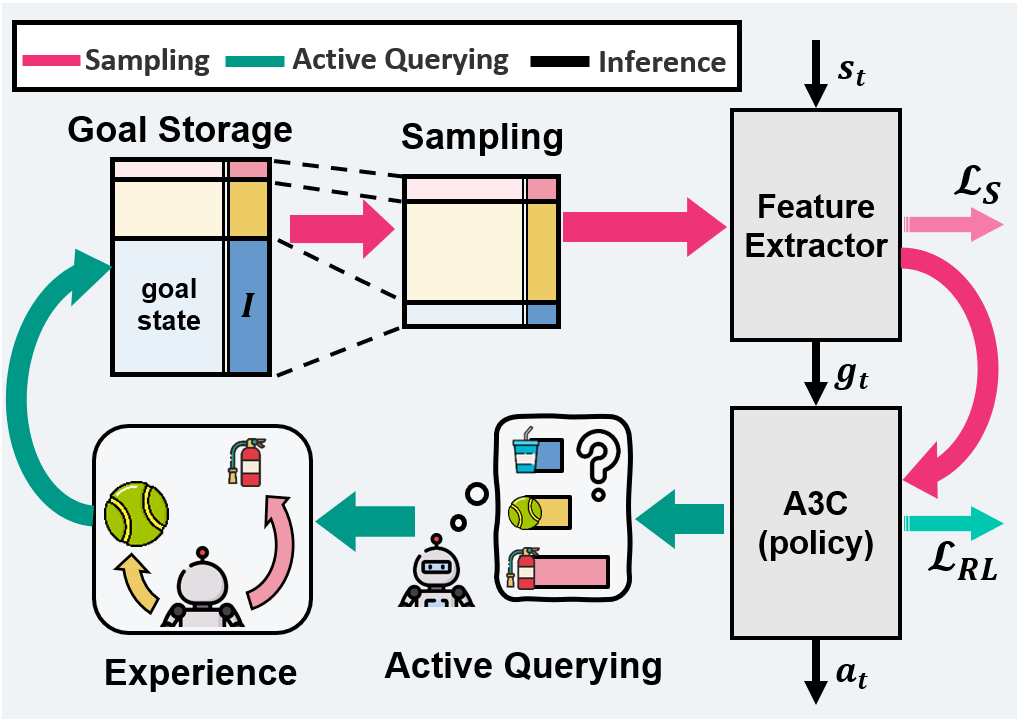}}
\caption{The overall architecture of L-SA framework.
    Our framework has a cyclical structure of active querying for targets required for learning, collecting goal states in goal storage, and adaptive sampling from the storage.
    The magenta line shows the computational flow when calculating $\mathcal{L}_S$, and the feature extractor is updated by sampling from goal storage.
    The green line indicates the active querying flow during \textit{training}, using the instruction determined by the active querying.
    The black line represents the inference flow during \textit{testing}. 
    }
\label{fig:arch}
\end{center}
\end{figure}

\section{L-SA Framework: Learning by Adaptive Sampling and Active Querying}

In this section, we present the L-SA framework to resolve the under-explored target problem in multi-target reinforcement learning.

\subsection{L-SA Framework}
\label{sec:l-sa}


The L-SA framework, which consists of adaptive sampling and active querying, is a cyclic mechanism, as shown in Figure~\ref{fig:arch} and Algorithm~\ref{alg:l-sa} in the Appendix \ref{app:algo}.
Prior to adaptive sampling, the agent collects goal states into goal storage through trial and error.
When performing representation learning for goals, the agent performs adaptive sampling from the goal storage.
The agent's policy and feature extractor are updated via SupCon loss $\mathcal{L}_S$ to better discern the targets \cite{jaderberg2016reinforcement,kim2021goal}.
The policy builds experience through trial and error on targets that require further training, which are prompted as instructions $I^x$ by active querying.
From these experiences, the goal storage is expanded by collecting success states corresponding to the instructions $I^x$.
Through this process, adaptive sampling and active querying have a cyclical relationship.
The sampling and active querying can be used independently of each other, but when used together, they form a virtuous cycle and alleviate UTP in multi-target tasks.

\subsection{Adaptive Sampling for Representation Learning}
\label{sec:sampling}

\begin{figure}[t]
    \centering
    \includegraphics[width=0.9\columnwidth]{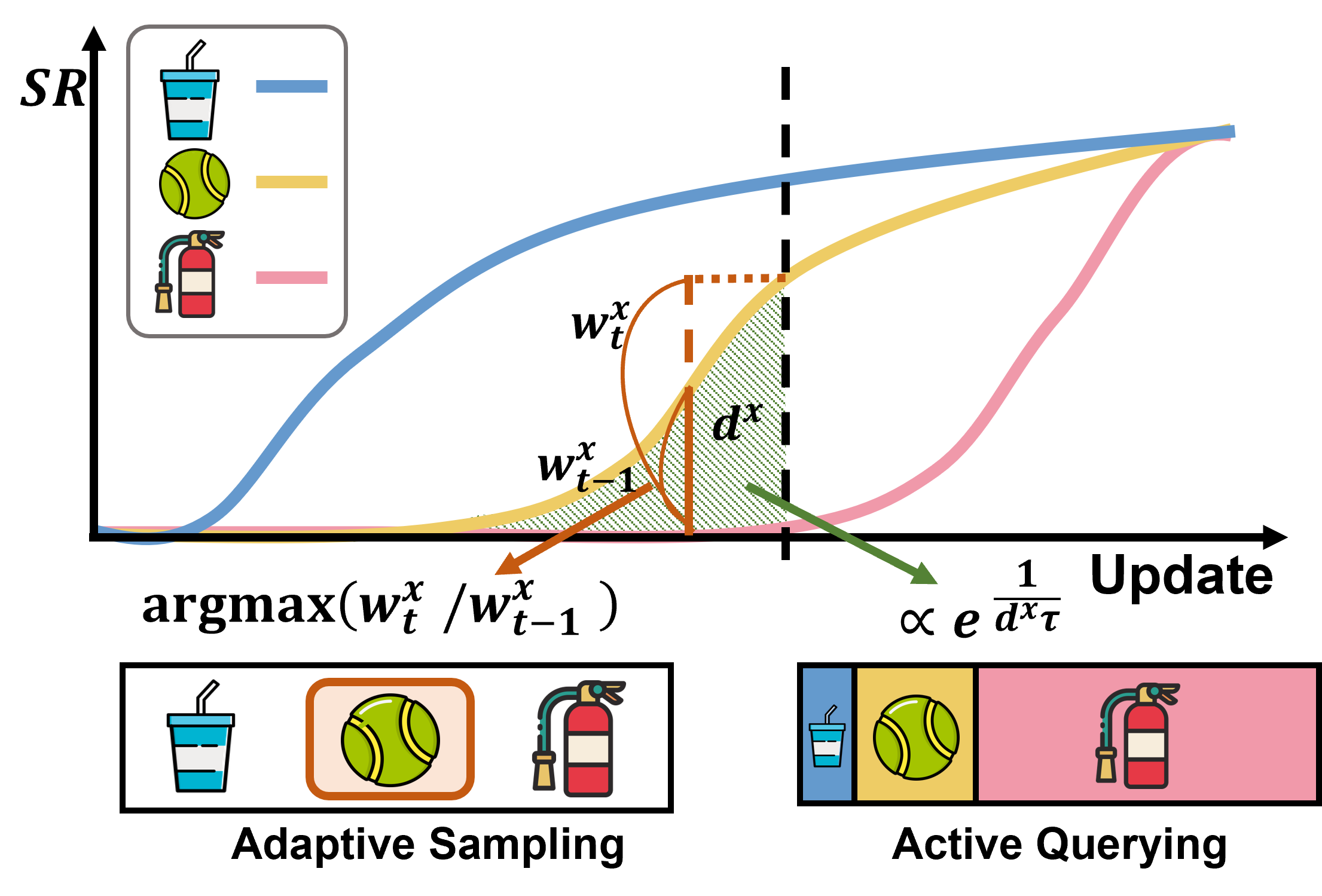} 
    \caption{Methods for adaptive sampling and active querying in L-SA framework.
    i) Adaptive sampling method allocates a higher portion of the batch to a target with a greater increase in success rate.
    ii) Active querying sets the instruction with a higher probability to a target with lower cumulative goal storage, in order to promote further attempts at under-explored targets.
    }
    \label{fig:method}
\end{figure}

In a multi-target task where targets vary in difficulty, most of the goal storage tends to be occupied by easier targets, and thus the random sampling virtually excludes hard targets.
This provides even less opportunity for the difficult under-explored target to be learned.
To resolve such UTP, we propose adaptive sampling for the L-SA framework, making efficient use of the goal storage data. 
In detail, we sample data with the emphasis on one specific target that we want to focus the representation learning on, which we call the \textit{focused target}.
To choose the focused target, we recognize three observations which are illustrated in Figure \ref{fig:method}: (1) the success rate for each target increases dramatically as the learning commences; (2) the time when the increasing success rate starts varies according to the difficulty of the target; (3) the success rate saturates and the rate of change decreases for the targets that the agent has practically finished learning.

Based on these traits, we determine the focused target $\tilde{x}_t$ according to $\tilde{x}_t = \operatorname*{argmax}_{x \in \mathcal{X}} w_t^x / w_{t-1}^x$  
where $\mathcal{X}$ is the set of all target classes, and $w^x_t$ is the success rate of target $x$ at $t$-th update.
In other words, the focused target is selected as the target whose relative increase in success rate at $t$-th update is the greatest.
Subsequently, the representation learning of the focused target is boosted by sampling it with a higher portion for $\mathcal{L}_{S}$ updates.
To be specific, the sampling ratio $B_t(x)$ for each target $x$ to be used in $\mathcal{L}_{S}$ updates is calculated as $B_t(x) = m\times\boldsymbol{1}_{\tilde{x}_t=x} + \frac{1 - m}{N}$ 
where $N$ is the number of target classes, $m\leq 1$ is a hyperparameter that determines the weight between focused target and uniform sampling, and $\boldsymbol{1}$ is the indicator function.

In effect, this method initially increases the sampling ratio of easy targets that can be readily collected.
After that, when saturation is reached, the main sampling ratio is shifted towards the target of the next difficulty, and ultimately, the hard target can be efficiently learned.
As the focused target is determined based on the change in the success rate, learning proceeds with targets chosen adaptively. 

\begin{figure*}[t]
\centering
    \subfigure[\footnotesize
    \label{fig:studio}Studio]{\includegraphics[width=0.32\textwidth]{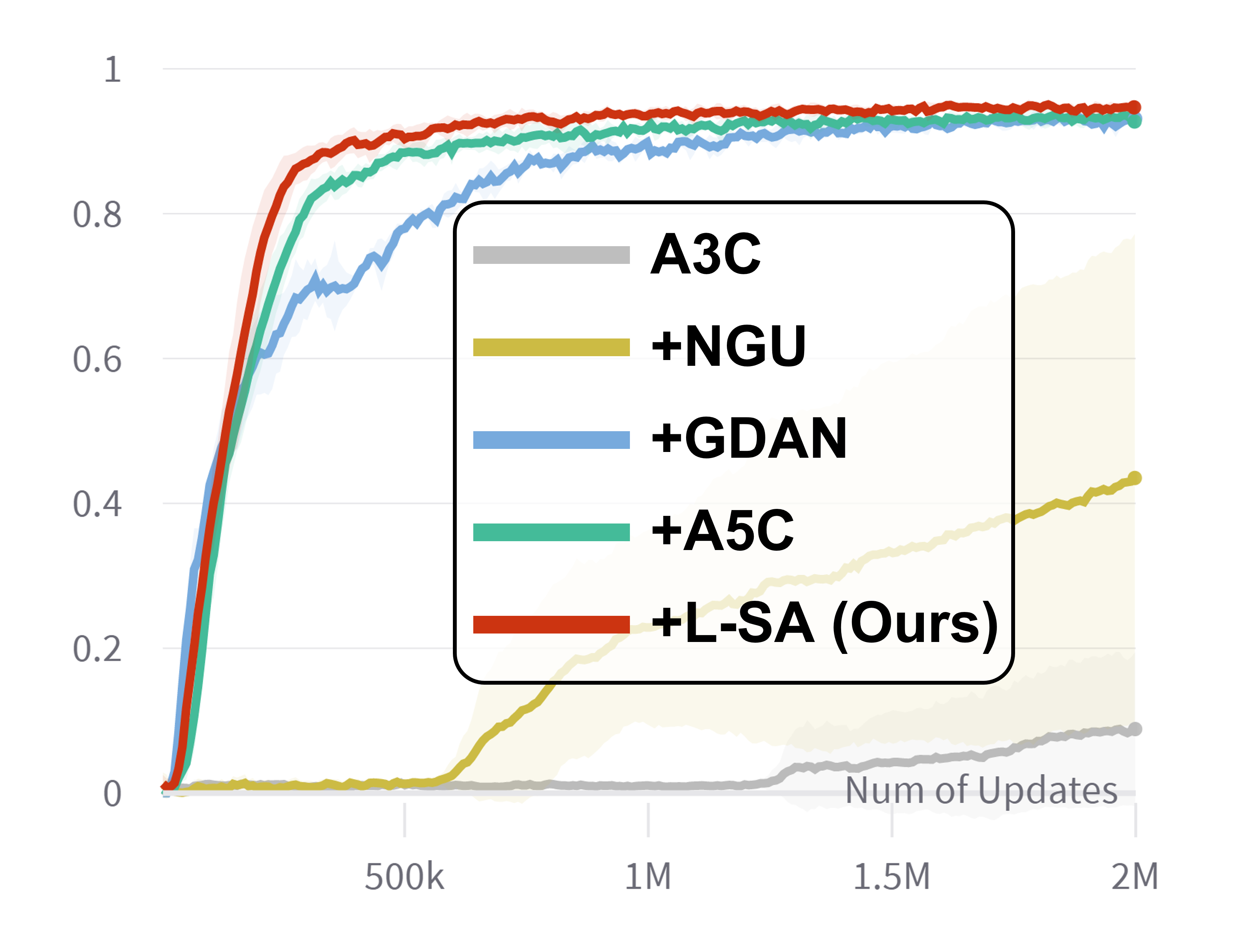}}
    \centering
    \subfigure[\footnotesize
    \label{fig:studio2e2h}Studio-2N 2H]{\includegraphics[width=0.32\textwidth]{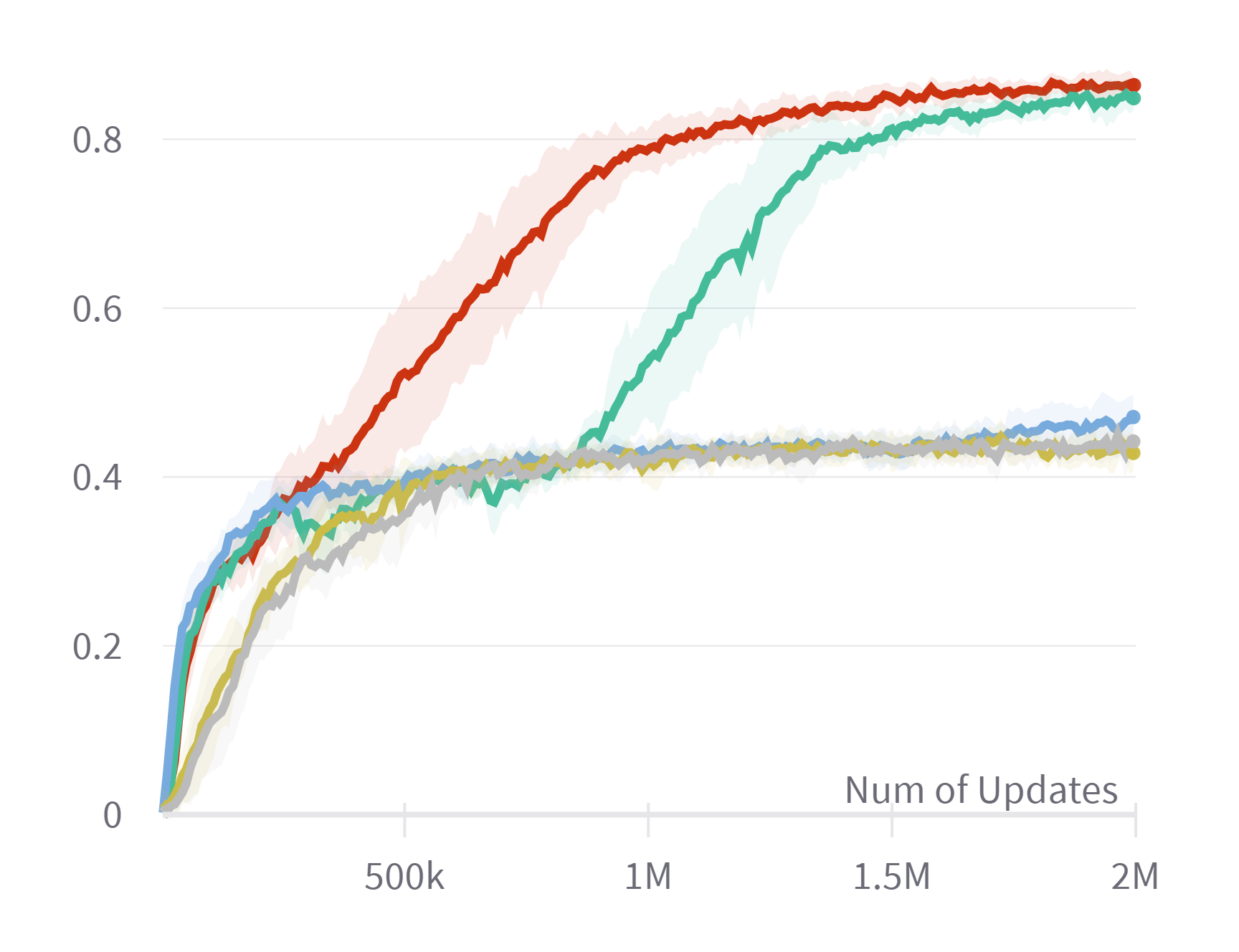}}
    \centering
    \subfigure[\footnotesize
    \label{fig:maze2e2h}Maze-2N 2H]{\includegraphics[width=0.32\textwidth]{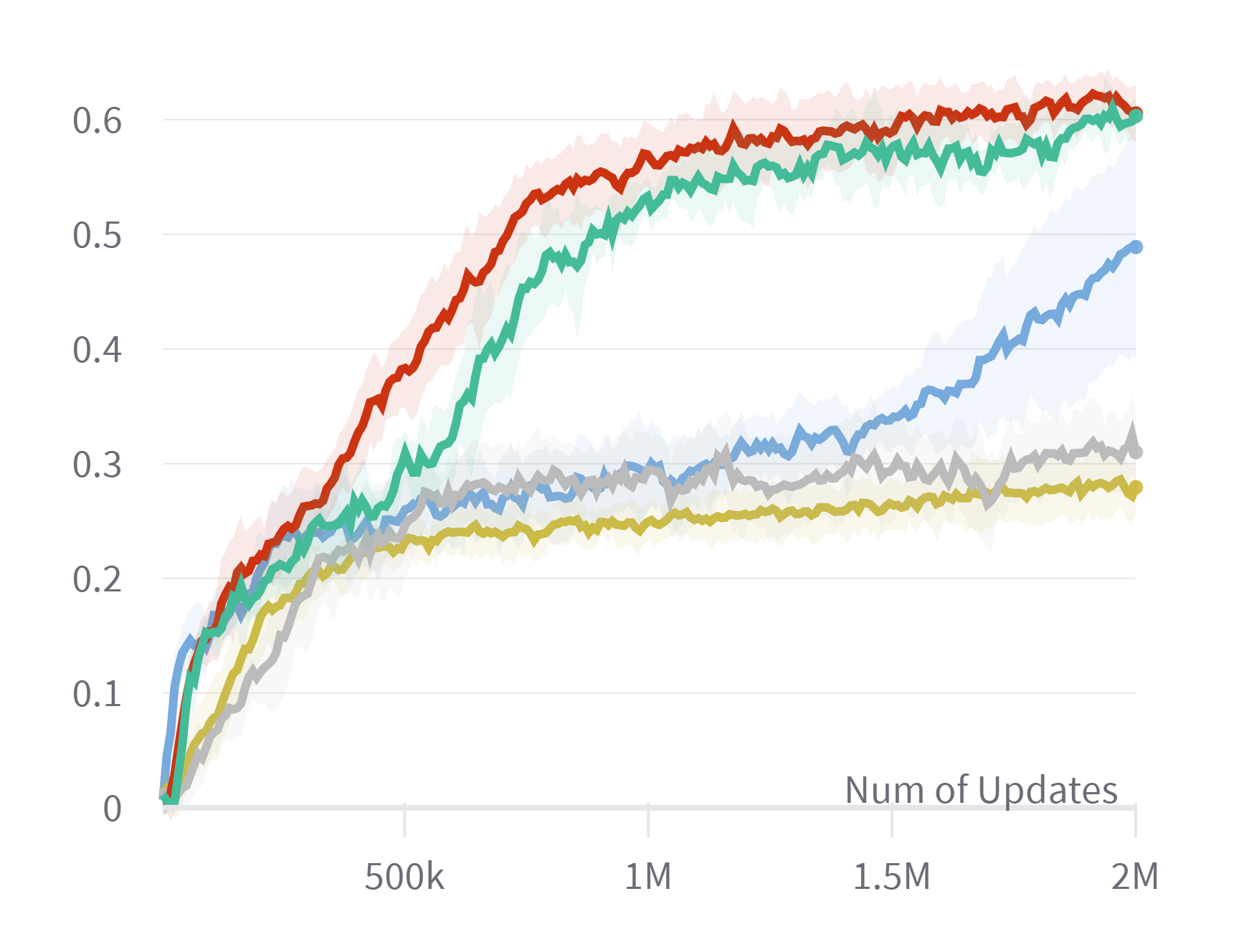}}
    \centering
    \\
    \subfigure[\footnotesize
    \label{fig:studio6e}Studio-6N]{\includegraphics[width=0.32\textwidth]{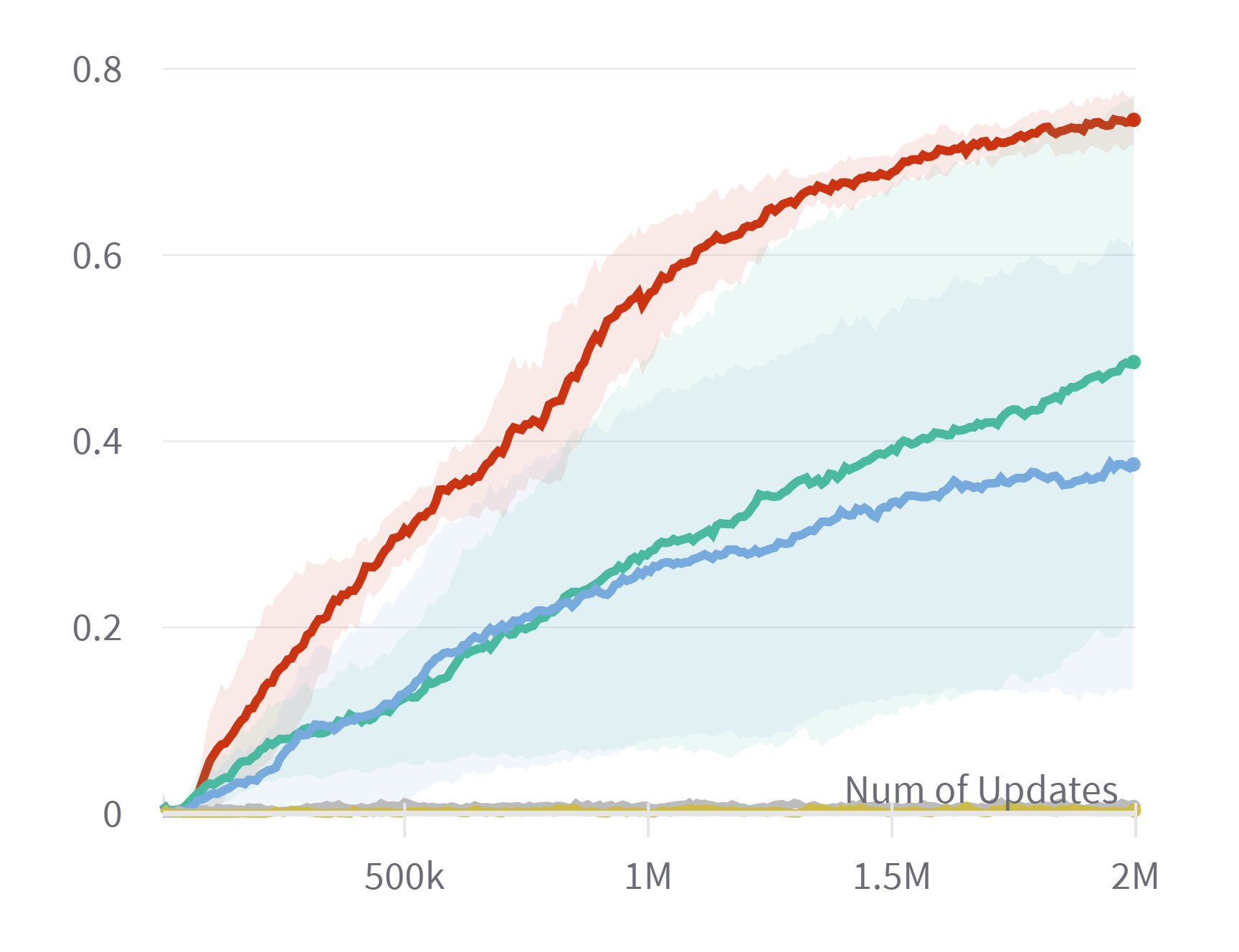}}
    \centering
    \subfigure[\footnotesize
    \label{fig:studio3e3h}Studio-3N 3H]{\includegraphics[width=0.32\textwidth]{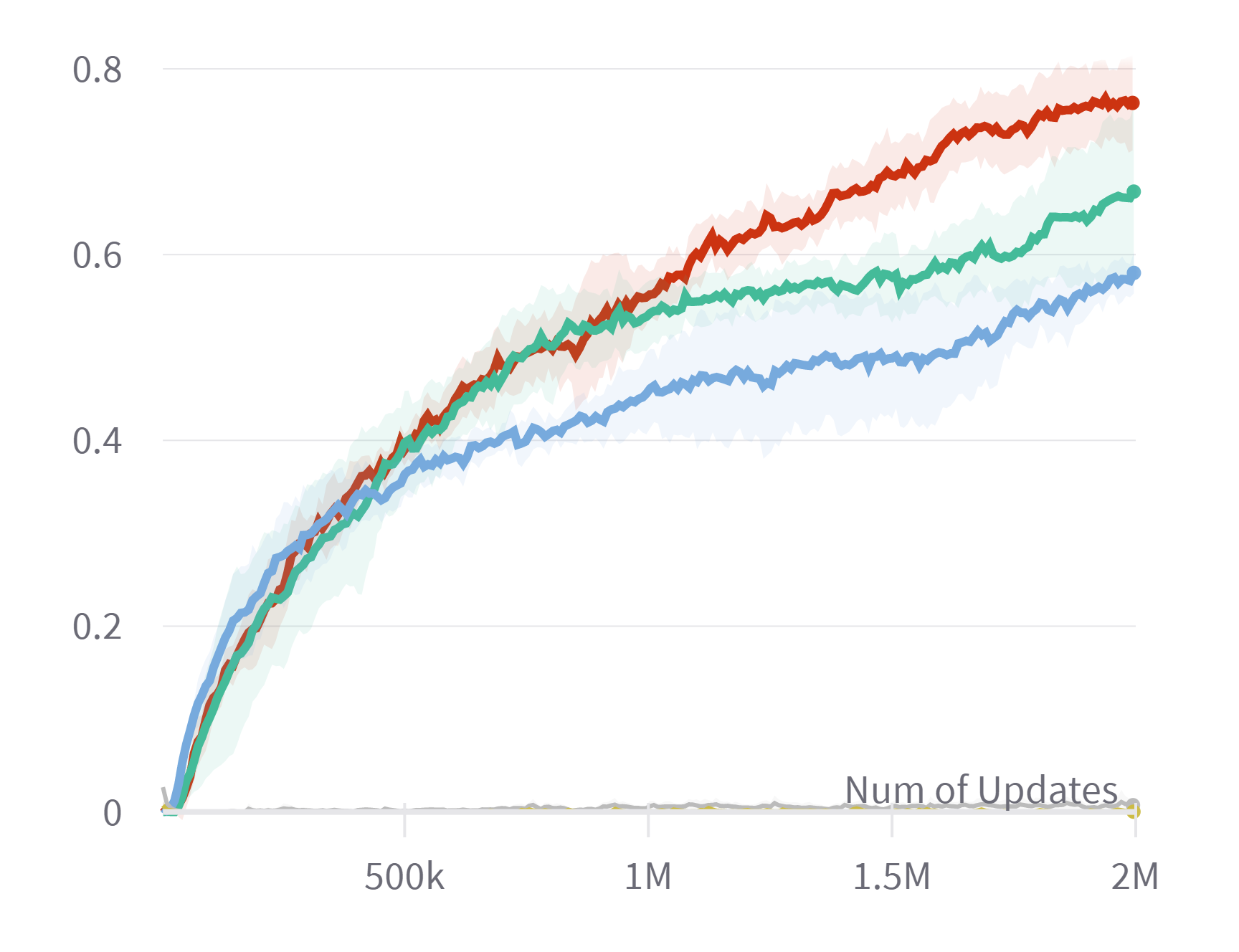}}
    \centering
    \subfigure[\footnotesize
    \label{fig:studio8e}Studio-8N]{\includegraphics[width=0.32\textwidth]{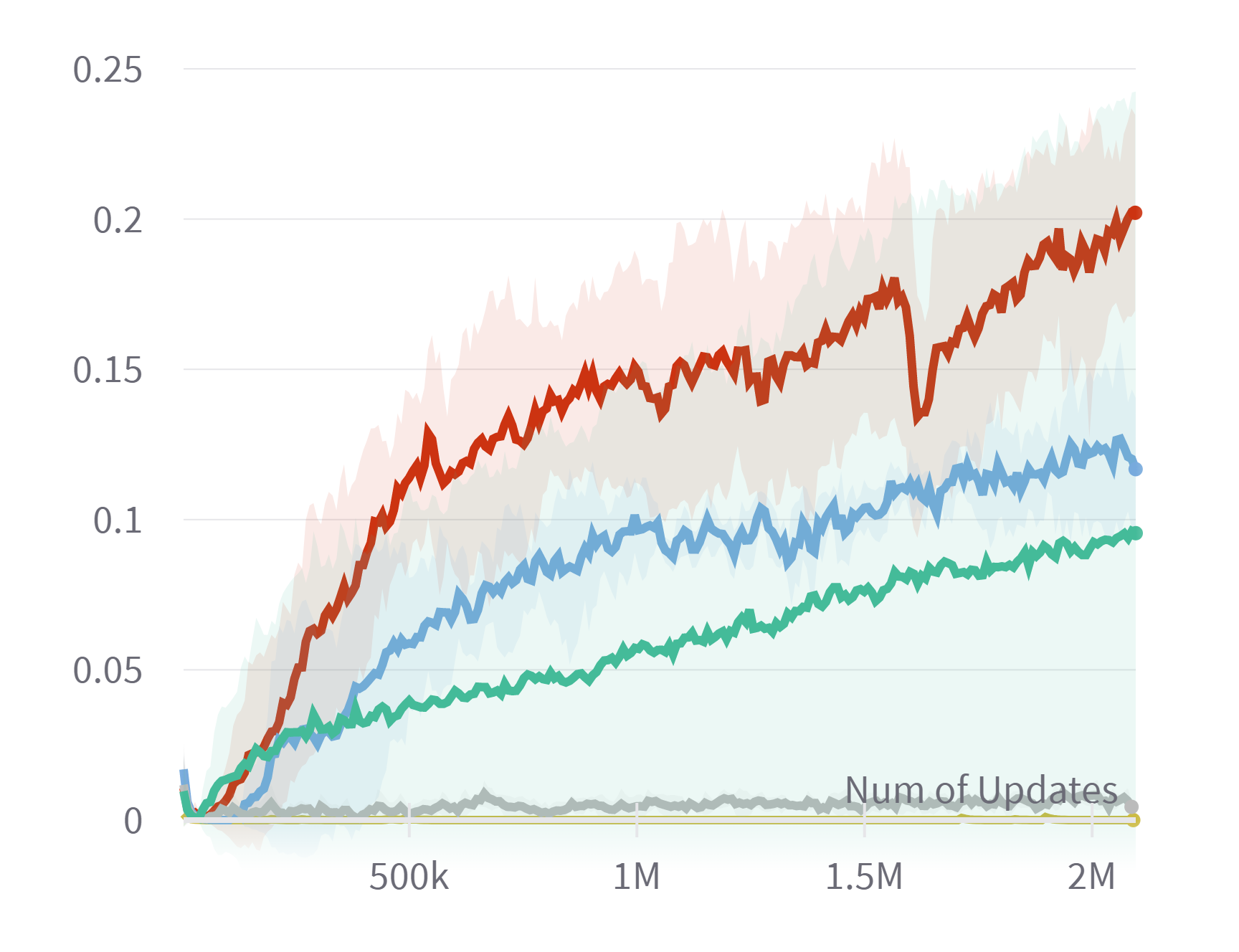}}
    \centering
    \caption{Our experiments in studio and maze environments with UTP.
    Based on the studio setting with 4 normal targets without UTP (a), we use a variant with 2 normal (N) and 2 hard (H) targets (b) or a maze environment with 2 N and 2 H targets (c).
    Studio with 6 N targets (d), 3 N and 3 H targets (e), and 8 N targets (f) are also included as highly complicated tasks.
    Our proposed method shows remarkable results in all experiments.
    The x-axis is the number of updates and the y-axis is the success rate.
    All experiments are repeated 5 times or more, and each curve indicates mean $\pm$ standard deviation.
    }
\label{fig:main_exp}
\end{figure*}

\subsection{Active Querying for Experiences}
\label{sec:query}

It is crucial for an agent to actively seek interactions with various targets in a multi-target task, rather than being restricted to only performing externally given instructions.
In particular, in situations where success experiences between targets differ significantly coordination of trial and error is necessary.

%

Specifically, one of the targets is specified via the instruction $I^x$.
The active querying sets the instruction $I^{x}$ to schedule trial-and-error for each target based on goal storage, via the scheme $A(x) = \cfrac{\exp\left(\{d^x\times\tau_a\}^{-1}\right)}{\Sigma_{x' \in \mathcal{X}} \exp\left(\{d^{x'}\times\tau_a\}^{-1}\right)}$ 
where $d^{x}$ is the proportion of the target $x$ in goal storage. 
In $A(x)$, the smaller the temperature $\tau_a$, the higher the sensitivity.
Instruction $I^x$ is multinomially determined by the ratio obtained through $A(x)$.
We use the exponential of the reciprocal $\exp(\{d^x\}^{-1})$ to make the proportion $d^x$ of the target data $x$ occupy a smaller ratio compared to the total amount in the goal storage.

With this method, instructions are set more often for the targets that occupy a small proportion of the goal states collected in the goal storage as illustrated in Figure~\ref{fig:method}.

\section{Experiments}
\label{sec:exp}
In this section, we show that the L-SA framework is effective in alleviating under-explored target problems by comparing L-SA with several baselines in various experimental settings.

\subsection{Task Details}
We conduct extensive experiments to investigate the performance of our proposed method in a task with UTP. To do so, we set up multi-target navigation tasks using ViZDoom \cite{kempka2016vizdoom} in Figure~\ref{fig:app_item} visualized in Appendix~\ref{app:env}. Given egocentric vision as observation, the agent can select one of three discrete actions (TurnLeft, TurnRight, and GoStraight). 

Targets in these tasks are different types of objects that can be reached by the agent.
We adjust each target's difficulty via the distance of its spawn position from the agent's initial position.
To be specific, normal-difficulty targets are spawned randomly at any distance from the agent, while hard-difficulty targets are always generated far away from the agent.
All tests in our experiments use random instructions, and the success rate is calculated as the average across 500 episodes and are repeated at least five times.
More details of these tasks can be found in the Environmental Details of the Appendix \ref{app:env}.
We list the maps and settings used for our experiments below:
\begin{itemize}
    \item \textbf{Studio}: This experiment is the same as the V1 setting of the study by \citet{kim2021goal}, and four targets of normal-difficulty are generated in a blank room.
    \item \textbf{Studio-2N 2H}: This experiment is the same as the Studio setting, except with two normal-difficulty targets and two hard-difficulty targets.
    
    \item \textbf{Maze-2N 2H}: This map is structured like a maze, consisting of two normal- and two hard-difficulty targets.
    \item \textbf{Studio-6N}, \textbf{Studio-3N 3H} and \textbf{Studio-8N}: These tasks consist of six normal-, three normal- and three hard-, and eight normal-difficulty targets in the Studio, respectively.
\end{itemize}

\subsection{Experiments with UTP}
We compare the performances of the following methods:
\begin{itemize}
    \item A3C: Base RL algorithm, proposed in \cite{mnih2016asynchronous}, with network architectures proposed for visual navigation by \cite{wu2018building} using LSTM and gated-attention \cite{chaplot2018gated}.
    \item + NGU: This method \cite{badia2020never} pursues exploration by giving an intrinsic reward for visiting various states based on the experience. This is the recent study that breaks the record of difficult tasks in Atari.
    \item + GDAN \cite{kim2021goal}: This is the current state-of-the-art method in a multi-target task, and we adopted their idea to have the agent learn to discriminate targets in our study.
    \item + A5C \cite{sharma2017learning}: This is a method on multi-task learning, which performs learning on tasks in different environments by score-based sampling.
    A5C increases the sampling ratio for tasks with large differences between the current score and their reference scores.
    In the experiments, this method is applied in place of each of sampling and active querying.
    \item + \textbf{L-SA} (ours): This is our proposed framework that pursues experience and efficiency through adaptive sampling and active querying.
\end{itemize}


In Figure~\ref{fig:main_exp}, learning curves for various tasks including UTP are shown.
In all tasks, our method achieves the steepest learning curves as well as the highest success rate.
In experiments with four normal targets, as shown in Figure~\ref{fig:studio}, the gap between methods designed for multi-target tasks (e.g. GDAN, A5C, and L-SA) and those that do not consider multiple targets (e.g. A3C, NGU) appears large.
In Figure~\ref{fig:studio2e2h} and \ref{fig:maze2e2h}, only L-SA and A5C successfully learn including hard targets, while all the baselines learn well only for two normal targets.
This shows that even applying the A5C method to sampling and active querying methods in our framework can overcome the limitations of the general learning methods.

In addition, Figure~\ref{fig:studio2e2h} and Table~\ref{tab:results-2E2H} in Appendix~\ref{app:exp} display superior sample efficiency of our method, achieving 490\% Sample-Efficiency Improvement (SEI\footnote{$SEI=n_A/n_B$ where $n_A$ and $n_B$ are the number of updates when method $A$ and $B$ reach $B$'s the best success rate respectively.}) over A3C.
Furthermore, as shown in Figure~\ref{fig:studio6e} and Table~\ref{tab:sei-6E}, our method improves the success rate by about two times and attains 289.6\% SEI compared to GDAN, the state-of-the-art method on multi-target tasks.
In Figure~\ref{fig:studio6e}, \ref{fig:studio3e3h}, and \ref{fig:studio8e}, methods that do not consider multiple targets show near-zero performance because the task has become more difficult due to more diverse targets.
In other words, our framework, based on adaptive sampling and active querying, is essential for successful and efficient learning in complicated multi-target tasks.
We speculate that the instability of the baseline methods in the Studio-6N task is due to distractions caused by the six targets of similar difficulties.
In contrast, our method is robust in this task, because adaptive sampling assigns a curriculum from easy to hard targets.

\begin{table}[t]
\caption{Success rate (SR) and Sample-Efficiency Improvement (SEI) compared to GDAN in \textbf{Studio-6N}.}
\vskip 0.1in
\centering
\begin{tabular}{l|c|c|c}
\toprule[1pt]
Algorithm  & SR (\%) & \# of Updates &  SEI (\%) \\
\midrule
\midrule
A3C             & 1.2 $\pm$ 0.9 & - & -\\
+ NGU           & 1.6 $\pm$ 0 & - & - \\
+ GDAN                  & 37.6 $\pm$ 24.1 & 1.94 M & 100\\
+ A5C                  & 48.5 $\pm$ 28.3  & 1.44 M  & 134.72 \\
+ \textbf{L-SA} (ours)       & $\mathbf{74.5 \pm 2.7}$ & 0.67 M  & $\mathbf{289.55}$ \\
\bottomrule[1pt]
\end{tabular}
\label{tab:sei-6E}
\end{table}

\begin{table}[b]
\caption{Success rate (SR) of ablation studies for each sampling and querying.
We use the \textbf{Studio-2N 2H} for these experiments.
}
\vskip 0.1in
\centering
\begin{tabular}{l|c|c}
\toprule[1pt]
Algorithm  & $\begin{matrix}\textbf{Sampling}\\ \text{SR(\%)} \end{matrix}$  & $\begin{matrix}\textbf{Querying}\\ \text{SR(\%)} \end{matrix}$ \\
\midrule
\midrule
A3C + SupCon (GDAN)                  & \multicolumn{2}{c}{44.0 $\pm$ 1.3}  \\
+Uniform Sampling                 & 49.6 $\pm$ 7.1  & -\\
+A5C Sampling               & 56.4 $\pm$ 17.6 & - \\
+\textbf{Adaptive Sampling}(ours)    & \textbf{69.7} $\pm$ \textbf{21.0}  & -\\
+A5C Querying      & -  & 57.6 $\pm$ 14.9 \\
+\textbf{Active Querying} (ours)   & -  & \textbf{65.9} $\pm$ \textbf{11.6} \\
\midrule
\textbf{L-SA} (ours)    & \multicolumn{2}{c}{\textbf{86.6} $\pm$ \textbf{1.5}} \\
\bottomrule[1pt]
\end{tabular}

\label{tab:exp_ablation}
\end{table}

\section{Analyses}
\subsection{Ablation Studies}
\begin{figure}[t]
\centering
    \subfigure[\footnotesize
    \label{fig:sample1}Normal-difficulty Target]{\includegraphics[width=0.23\textwidth]{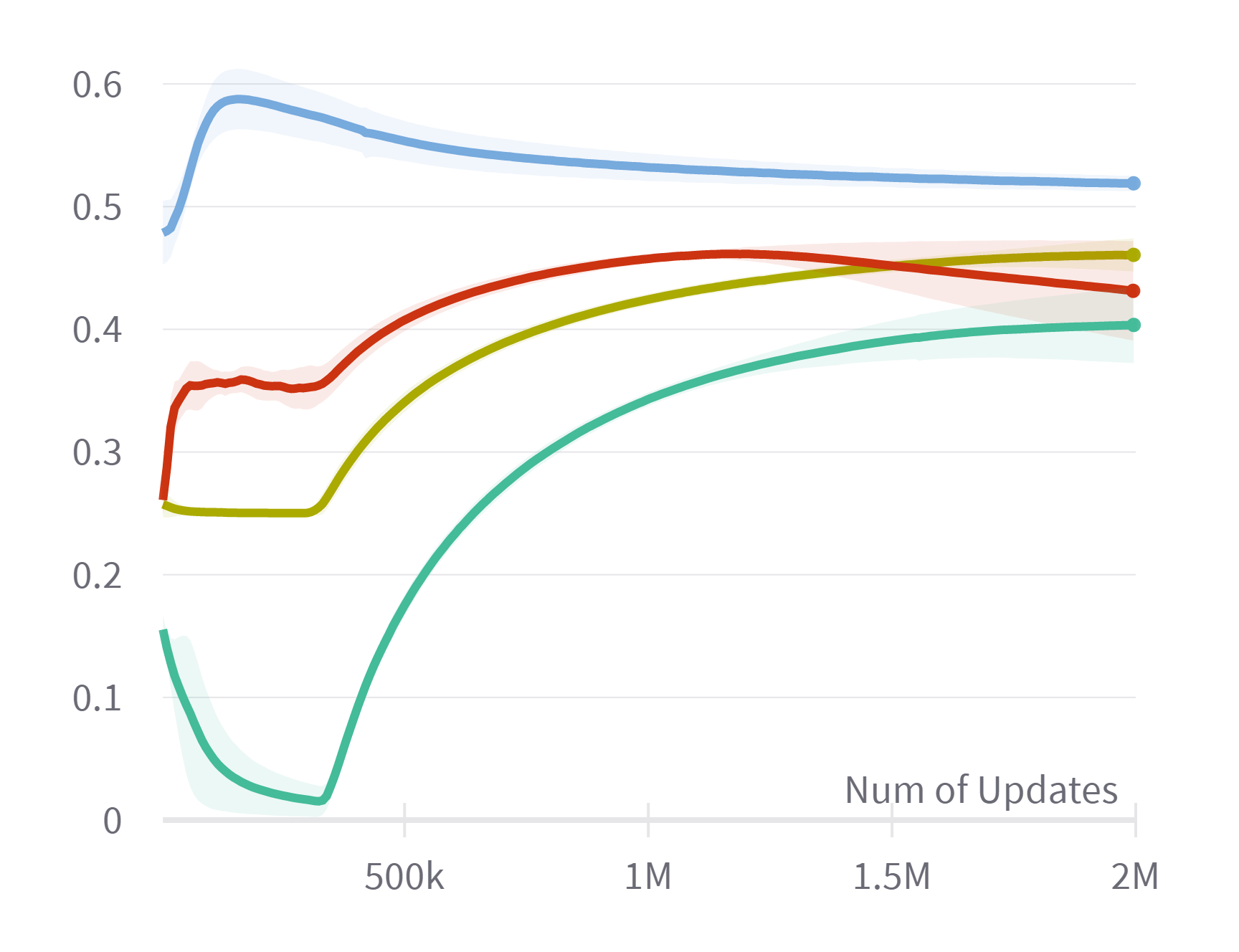}}
    \subfigure[\footnotesize
    \label{fig:sample2}Hard-difficulty Target]{\includegraphics[width=0.23\textwidth]{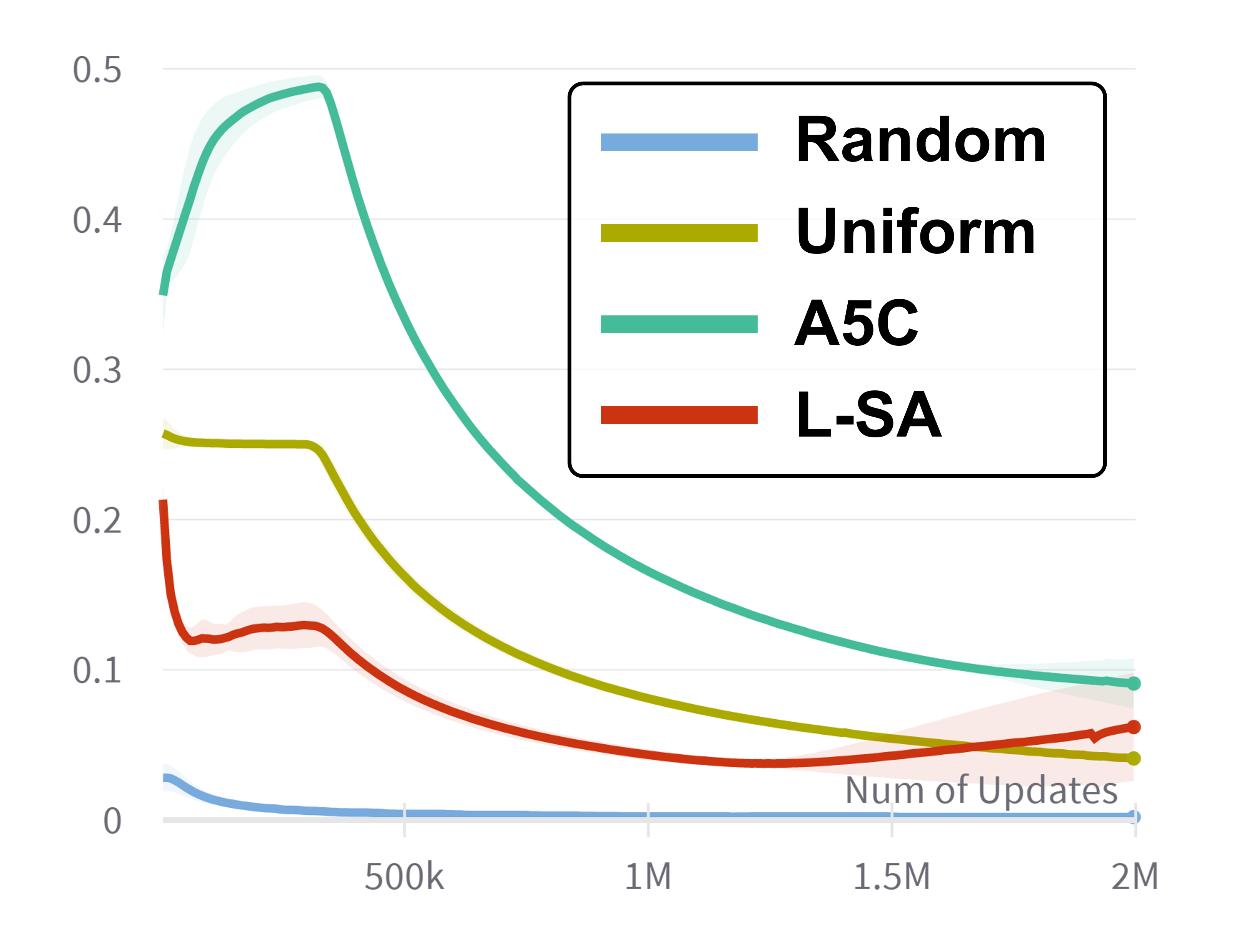}}
    \centering
    \caption{
    Ablation experiments for sampling method in Studio-2N 2H task.
    One normal target (a) and one hard target (b) are indicated.
    The vertical axis indicates the cumulative sampling ratio.
    Our method shows an increase in the cumulative sampling ratio of hard targets around 1.2 M updates in (b). 
    }
\label{fig:sample}
\end{figure}

In this section, ablation studies are conducted on \textbf{Studio-2N 2H} task to show the performance and role of sampling and active querying method. 
In this experiment, targets 0 and 1 are normal-difficulty targets and targets 2 and 3 are hard-difficulty targets. 
Due to the lack of space, only one normal target and one hard target are indicated in Figure~\ref{fig:sample}, \ref{fig:sample_value}, \ref{fig:query}, \ref{fig:sample_richness} and \ref{fig:query_storage} and the other targets are shown in Figure~\ref{fig:app_sample} and \ref{fig:app_query}, Appendix~\ref{app:exp}.

\subsubsection{Ablation Studies for Sampling Methods}
Figure~\ref{fig:sample} and Table~\ref{tab:exp_ablation} show the results of each sampling method.
During these experiments, the active querying method is set to random.

Table~\ref{tab:exp_ablation} shows the task success rates for each sampling method.
Figure~\ref{fig:sample1} and \ref{fig:sample2} show cumulative sampling ratio for normal- and hard-difficulty targets.
The blue lines in the figures refer to GDAN with random sampling, and we can confirm that sampling for the hard-difficulty target is neglected.
The uniform, A5C, and L-SA sampling methods decrease at about 320k updates because the storage reaches the maximum capacity and the storage ratio for hard-difficulty targets in Figure~\ref{fig:sample2} decreases.
The A5C, at the start of the learning, prioritizes the sampling of hard-difficulty targets that show a large difference between reference and experimental success rates.
The L-SA gradually rebounds at about 1.2M updates because the ratio of targets that are easily collected is high in the beginning, and the sampling rate for hard-difficulty targets is maintained at a steady rate minimally.
Our method of adjusting the proper proportion of the normal- and hard-difficulty targets shows that it is more effective than baseline methods.
We measure this as a sample richness in Sec~\ref{sec:rich}.

\begin{figure}[t]
\centering
    \subfigure[\footnotesize
    \label{fig:sample_value1}Normal-difficulty Target]{\includegraphics[width=0.23\textwidth]{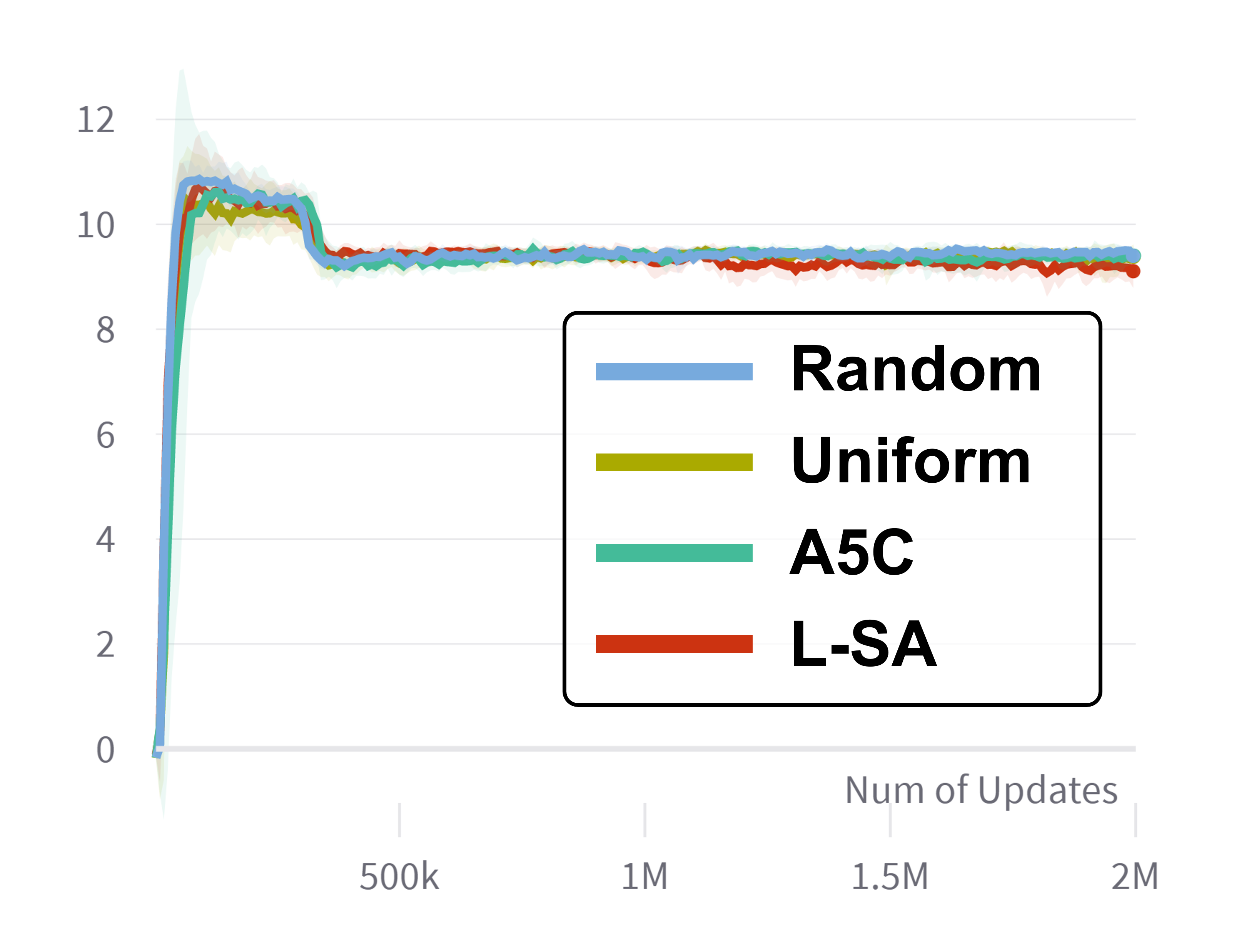}}
    \centering
    \subfigure[\footnotesize
    \label{fig:sample_value2}Hard-difficulty Target]{\includegraphics[width=0.23\textwidth]{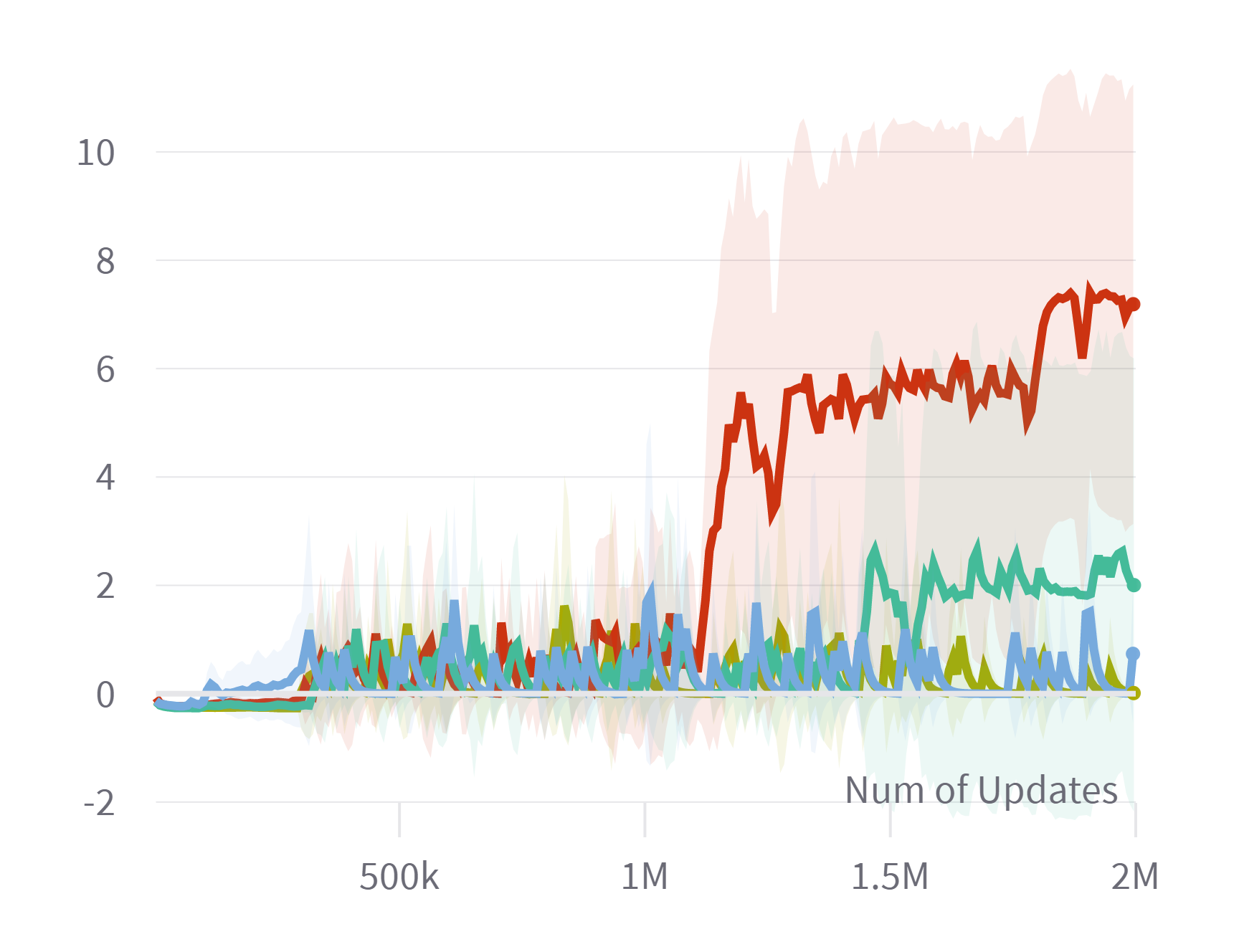}}
    \centering
    \caption{
    Value inference for each target in ablation experiments for sampling.
    In (a), all models show similar results, but in (b), which corresponds to hard-difficulty targets, only our method offers the highest value inference results.    
    }
\label{fig:sample_value}
\end{figure}

Additionally, in Figure~\ref{fig:sample_value}, to demonstrate the proper value estimation of the sampling method, we infer values for the latest ten goal states collected in the goal storage.
As a result, for the normal difficulty targets as in Figure~\ref{fig:sample_value1}, all methods show similar values, confirming that the learning is successful.
In Figure~\ref{fig:sample_value2}, our method infers the highest value, which is most similar to the saturated value in Figure~\ref{fig:sample_value1}. In other words, our method more accurately learns the discrimination of goals than other methods.

\subsubsection{Ablation Studies for Querying Methods}

Table~\ref{tab:exp_ablation} shows the ablation study results for each active querying method.
In these experiments, random and uniform methods are the same for querying, so only random is tested as the GDAN.
While performing the ablation study for active querying, the sampling method is set to random.

\begin{figure}[t]
\centering
    \subfigure[\footnotesize
    \label{fig:query1}Normal-difficulty Target]{\includegraphics[width=0.23\textwidth]{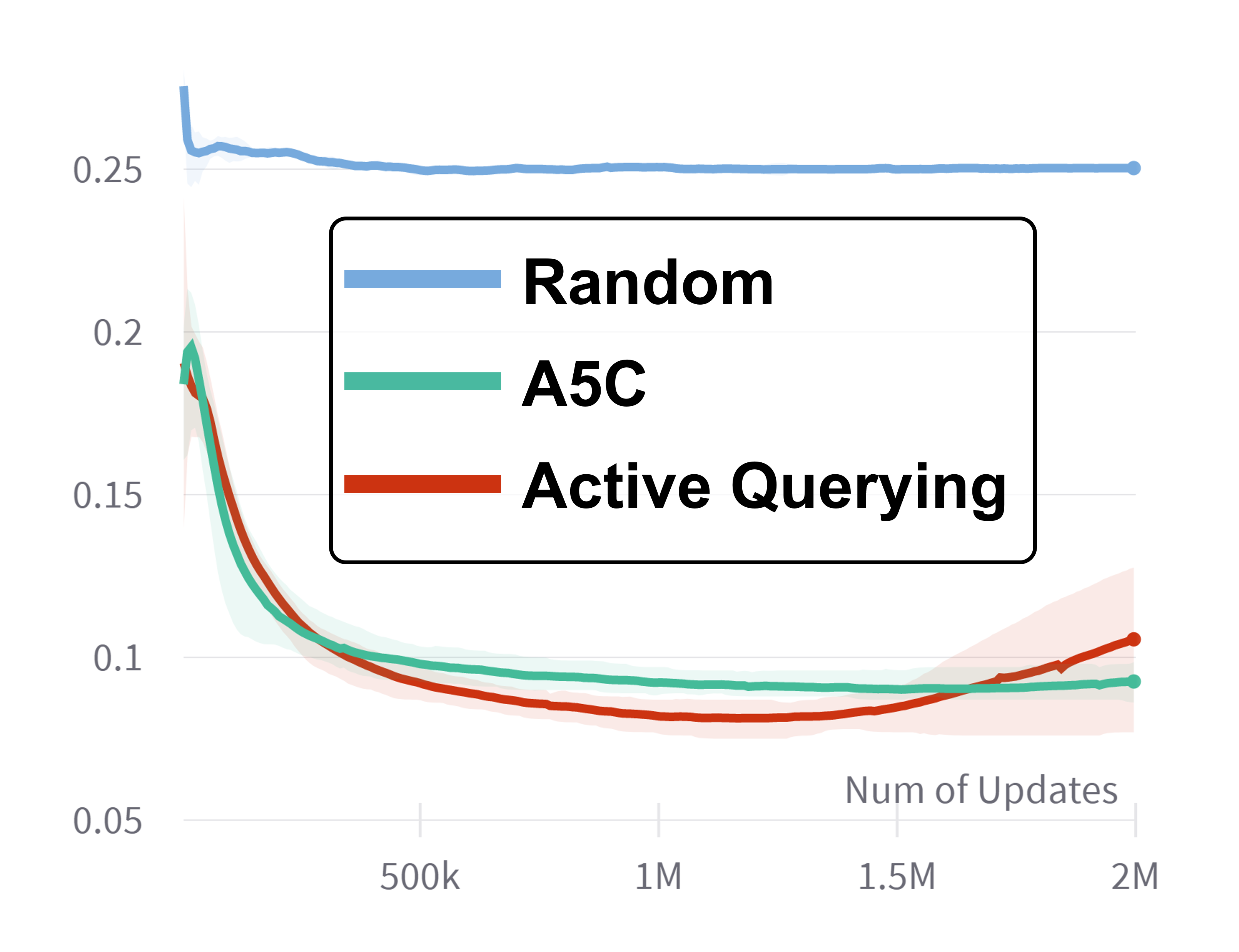}}
    \centering
    \subfigure[\footnotesize
    \label{fig:query2}Hard-difficulty Target]{\includegraphics[width=0.23\textwidth]{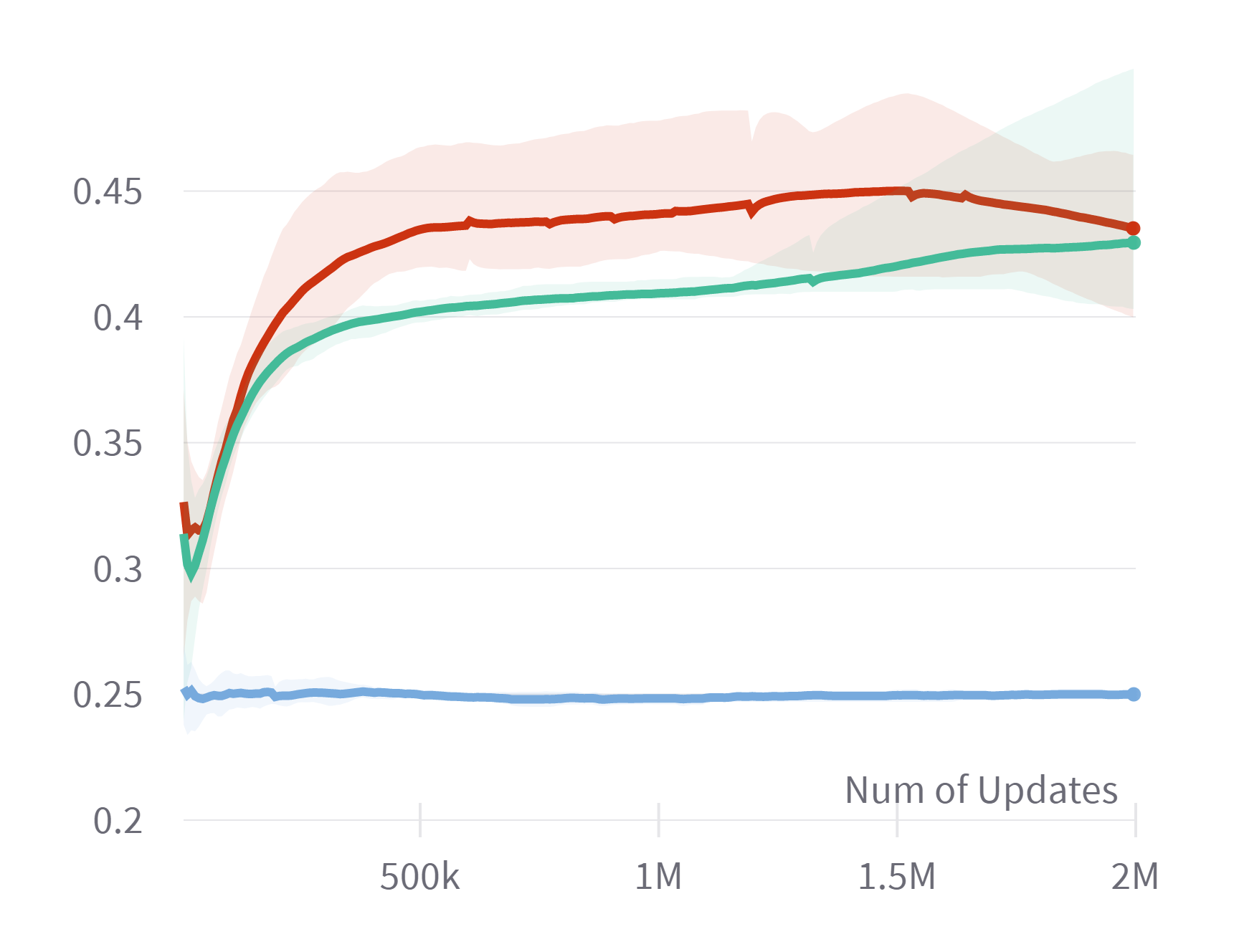}}
    \centering
    \caption{
    Ablation experiments for active querying.
    The vertical axis indicates the cumulative querying ratio for each target.
    Our method (red) shows a higher active querying ratio than the A5C method in (b). 
    Difficult tasks with low success rates can adjust the query ratio more precisely by ours rather than A5C.
    }
\label{fig:query}
\end{figure}

Figure~\ref{fig:query1} shows a normal-difficulty target, and Figure~\ref{fig:query2} shows a hard-difficulty target with the cumulative active querying rate.
We observe that the random method is not helpful for a task with hard difficulties, because it tries a uniform query regardless of the difficulty.
The A5C method performs similarly to the L-SA in Figure~\ref{fig:query1}.
However, the A5C method and active querying method of querying ratio show the largest difference in Figure~\ref{fig:query2} which corresponds to hard-difficulty targets.
The L-SA in Figure \ref{fig:query2} shows a higher querying ratio than the A5C method, which means that fewer goal states are collected in Figure~\ref{fig:query2} than in Figure~\ref{fig:query3} in Appendix~\ref{app:exp}. 
This higher active querying ratio leads to higher success rates when compared with A5C in Table~\ref{tab:exp_ablation}.
While difficult targets record low scores, our method determines targets that require more trial and error through differences in the goal states collected during training.

\begin{figure}[t]
\centering
    \subfigure[\footnotesize
    \label{fig:query_stor1}Normal-difficulty Target]{\includegraphics[width=0.23\textwidth]{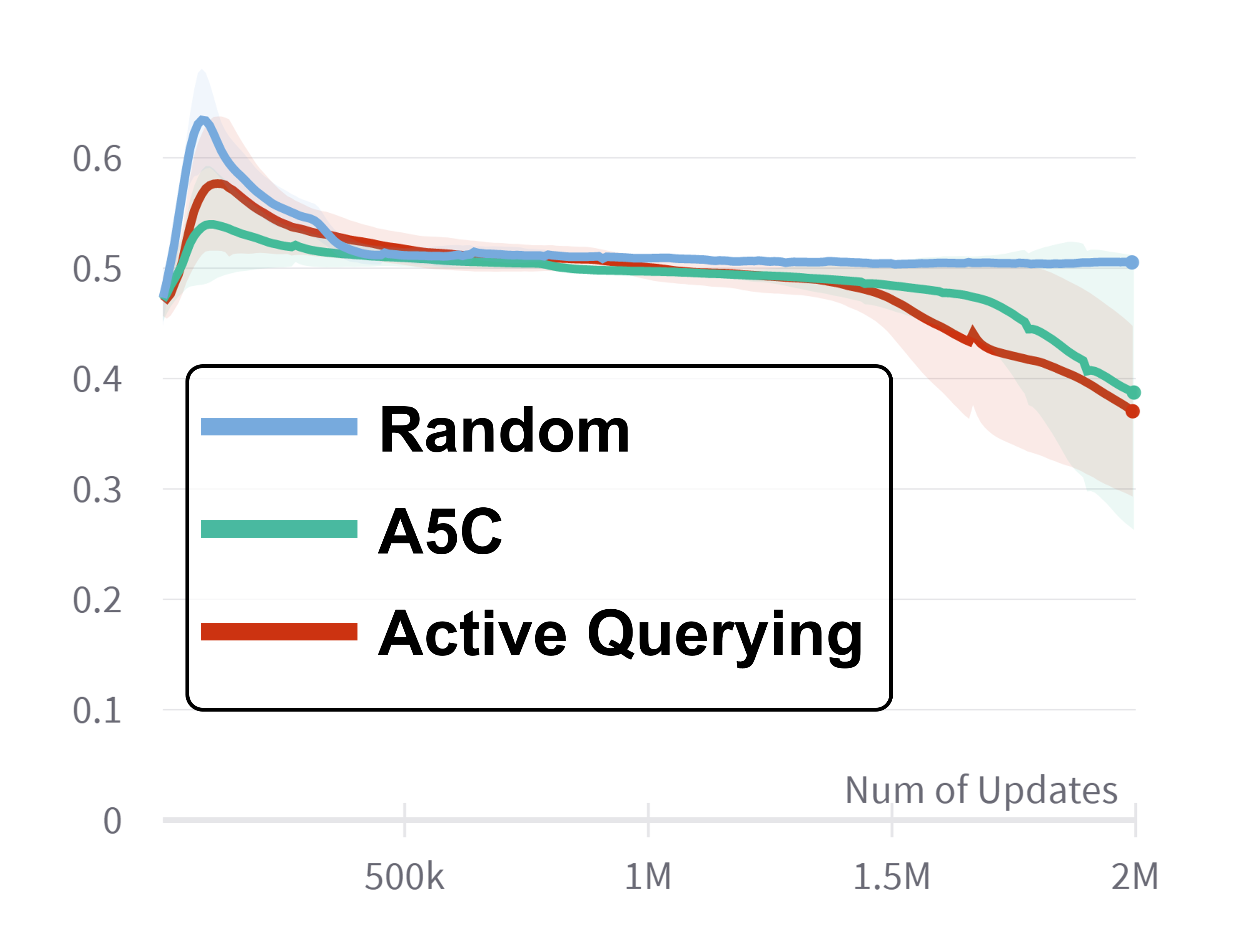}}
    \centering
    \subfigure[\footnotesize
    \label{fig:query_stor2}Hard-difficulty Target]{\includegraphics[width=0.23\textwidth]{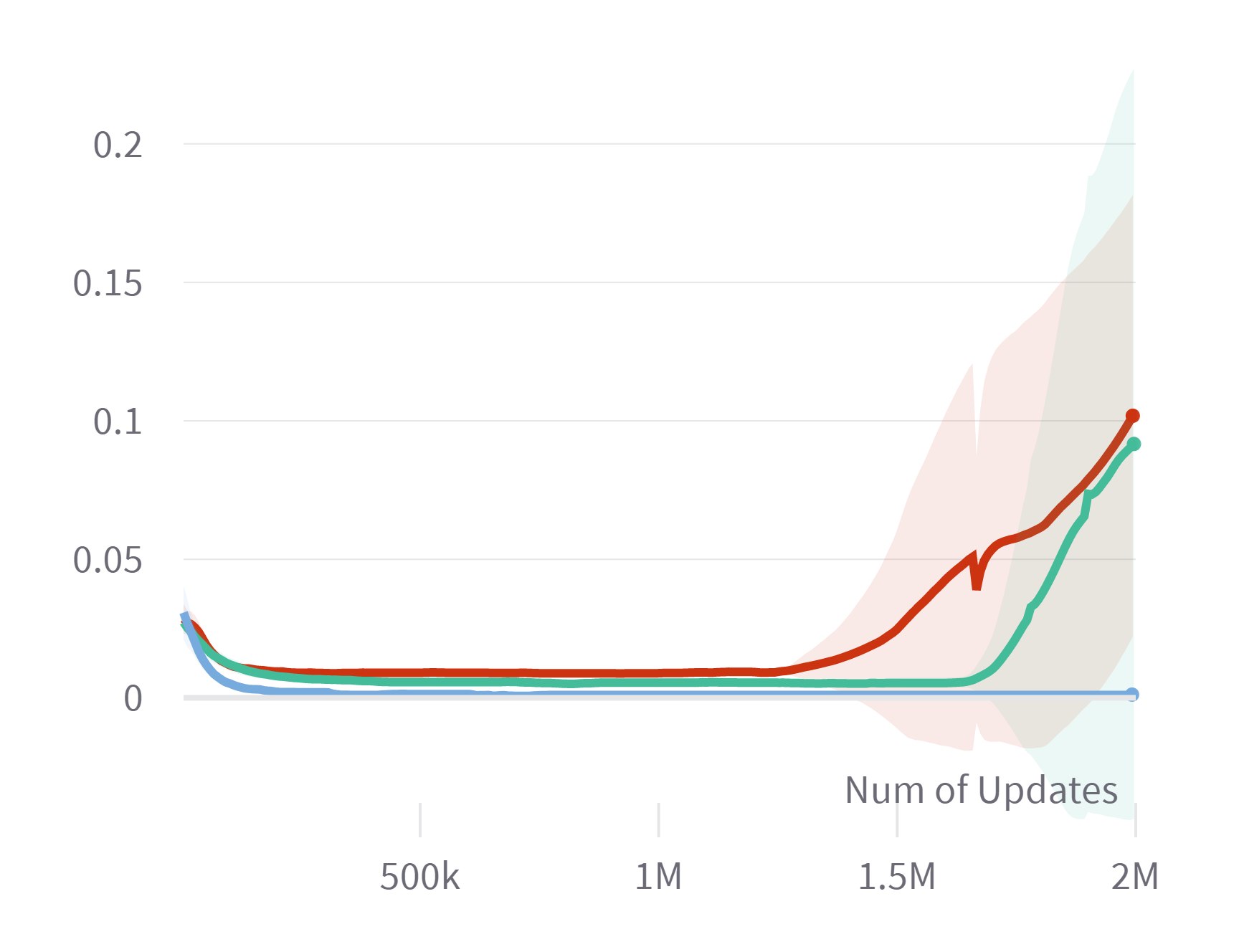}}
    \centering
    \caption{
    Storage ratio for each target in the ablation experiment for active querying.
    In (a), the normal-difficulty target shows a high storage share from the beginning, and in (b), the hard-difficulty target shows a red line first and highest share increase.
    }
\label{fig:query_storage}
\end{figure}

We visualize the proportion of goal storage for each target in Figure~\ref{fig:query_storage} to show that active querying works correctly.
In Figure~\ref{fig:query_stor1}, corresponding to the normal-difficulty target, it can be confirmed that all methods show the storage ratio of all targets with a high proportion.
In Figure~\ref{fig:query_stor2}, which corresponds to the hard-difficulty target, our method increases the collection rate first and quickly.

\subsection{Sample Richness Measurement}
\label{sec:rich}
To show the virtuous cycle of adaptive sampling and active querying that make up the L-SA framework, we measure sample richness.
The sample richness is calculated as $M_x / k_x$ where the number of states sampled for representation learning is $k_x$ and the total number of collected goal states in the goal storage is $M_x$ for target $x$.
Intuitively, the larger the richness, the more various data can be sampled compared to the amount of data needed for sampling.
Due to the characteristics of reinforcement learning that collects samples through trial and error, it is necessary to adjust the sample richness for each target.

\begin{figure}[b]
\centering
    \subfigure[\footnotesize
    \label{fig:sample_stor1}Normal-difficulty Target]{\includegraphics[width=0.23\textwidth]{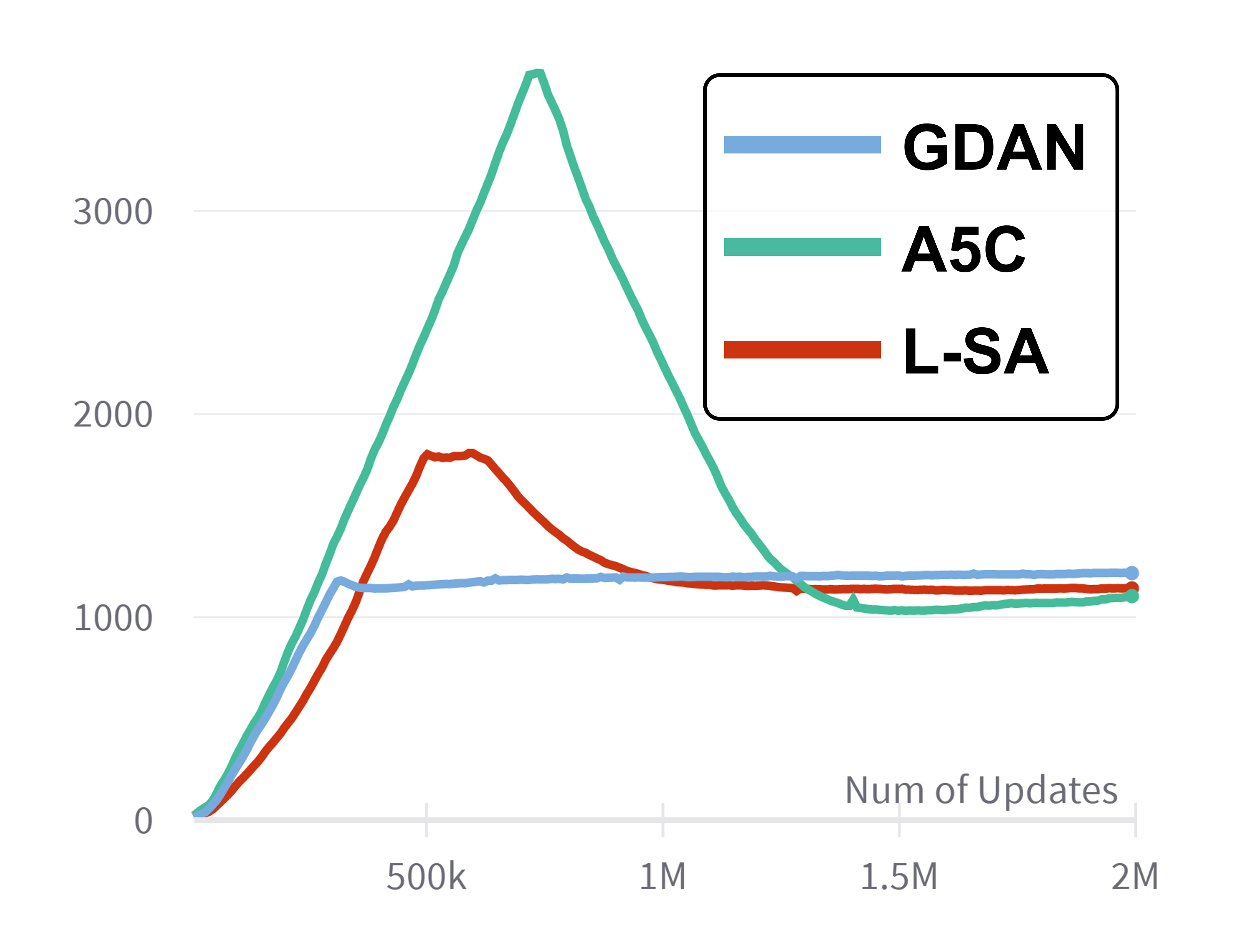}}
    \centering
    \subfigure[\footnotesize
    \label{fig:sample_stor2}Hard-difficulty Target]{\includegraphics[width=0.23\textwidth]{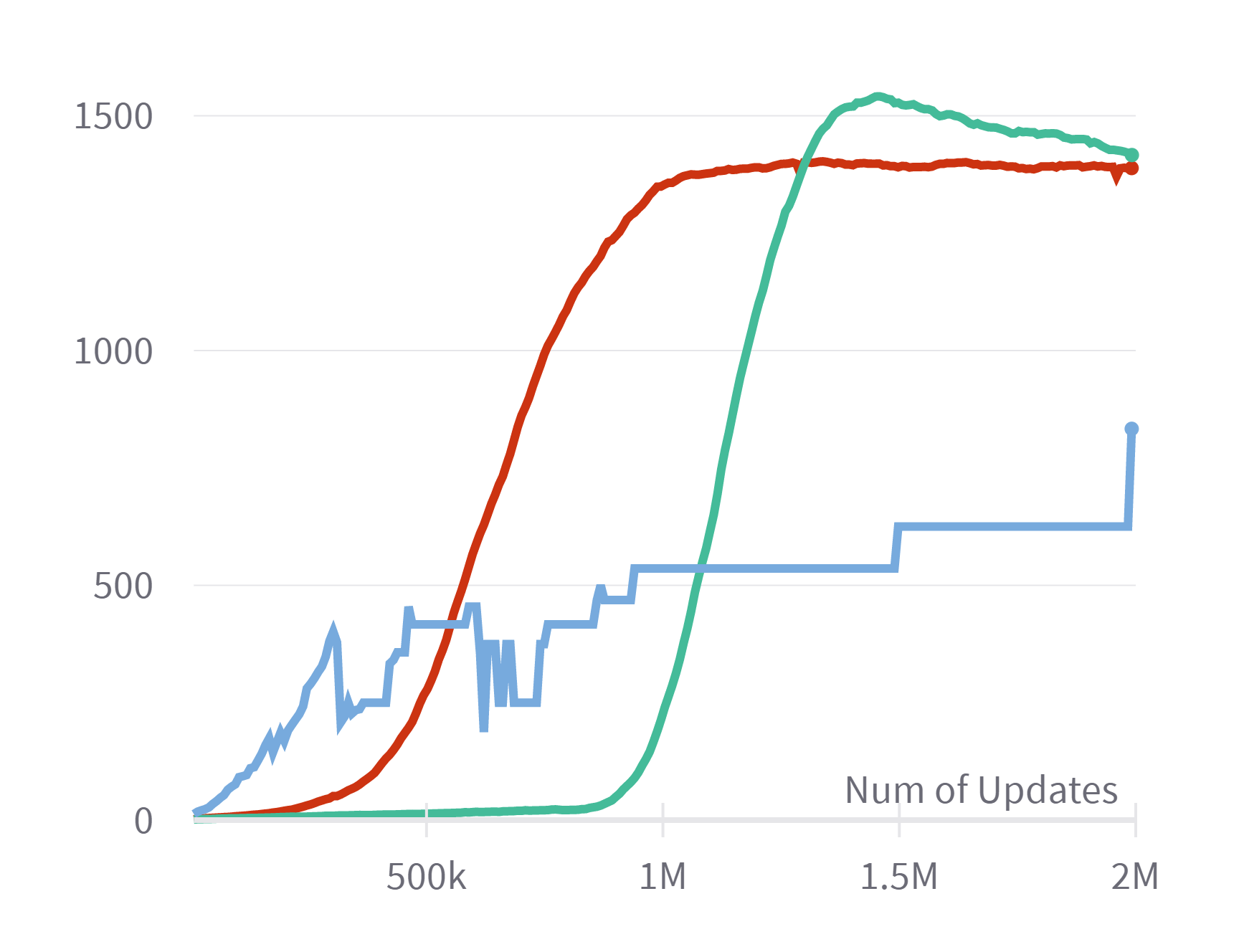}}
    \centering
    \caption{
    Sample richness measurement.
    The larger the sample richness (vertical axis), the greater the abundance of stored data compared to the amount of data to be sampled.
    If the richness is low, redundant sampling may occur due to the insufficiency of stored data.
    }
\label{fig:sample_richness}
\end{figure}

Figure~\ref{fig:sample_richness} displays the sample richness measurement in the Studio-2N 2H task.
The blue line (GDAN) consists of random sampling and random active querying.
Since normal-difficulty targets are intensively collected and sampled, hard-difficulty targets are neglected from learning.
Nonetheless, the sample richness rises because there are significantly fewer data actually sampled, as indicated by the blue line in Figure~\ref{fig:each_stor2} and \ref{fig:each_stor3} in Appendix~\ref{app:exp}.
The green line (A5C) rises the highest for normal-difficulty targets but shows a late improvement for hard-difficulty targets.
This method initially has high richness due to its low sampling ratio for normal-difficulty targets in Figure~\ref{fig:each_sample0} and \ref{fig:each_sample1} with a high storage ratio in Figure~\ref{fig:each_stor0} and \ref{fig:each_stor1}. 
On the other hand, the hard-difficulty targets yield very low richness due to the higher sampling ratio in Figure~\ref{fig:each_sample2} and \ref{fig:each_sample3} with low storage ratio early in Figure~\ref{fig:each_stor2} and \ref{fig:each_stor3}, resulting in inefficiency due to redundant sampling.

The L-SA framework initially attempts to increase the storage ratio for the hard-difficulty target by utilizing active querying in Figure~\ref{fig:each_query2} and \ref{fig:each_query3} while sampling a high proportion of normal-difficulty targets in Figure~\ref{fig:each_sample0} and \ref{fig:each_sample1}.
Unlike the A5C method, the L-SA framework rapidly increases the storage ratio for the hard-difficulty target by keeping lower sample richness of the easy-to-collect normal-difficulty targets through a high sampling ratio.
Our framework achieves these results through a virtuous cycle structure of experience and goal state collection through querying, and representation learning through adaptive sampling.

\section{Conclusion}
We propose the L-SA framework with a virtuous cycle structure through adaptive sampling and active querying.
The L-SA autonomously regulates the course of training based on the performance changes for each target and prompts experience on under-explored targets. 
Experimental results demonstrate that it alleviates UTP without prior knowledge or additional learnable parameters, showing state-of-the-art success rates and sample efficiency.
Our framework can be applied to multi-target tasks with visual navigation and effectively utilized for complex observations such as visual input.
However, this method is limited to tasks where the target is specific. Therefore, its application may be limited in tasks where the instructions do not explicitly specify targets (e.g. ``Keep going'').
Also, to apply active querying, we assume the agent can set the instruction.
In the future, we plan to expand the proposed method to tasks that require more complex actions (such as executing pick-and-place or reaching intermediate goals) to find solutions.
In addition, we intend to investigate the capability of L-SA to perform continual learning in tasks where the number of targets gradually increases.
We hope this helps to broader and more practical applications, effectively providing real-world service to humans.

\clearpage

\bibliography{example_paper}
\bibliographystyle{icml2023}

\newpage
\appendix
\onecolumn

\section{L-SA Framework}
\label{app:l-sa}
\subsection{Goal Storage}
\label{sec:storage}
Suppose that $I^x$ is set as the instruction of the episode with $x$ as the goal.
Then, during training, upon successful execution of the instruction, the (goal state, instruction) pair is stored in the goal storage, which acts as a dataset for subsequent representation learning.
Unlike the existing reinforcement learning, this method can collect goal-related data purely from reward signals. 

\subsection{L-SA}
To learn multi-processing-based A3C in our proposed framework, we create goal storage that can be shared among processes.
We calculate the sampling rates and the active querying rates in Sec.\ref{sec:sampling} and \ref{sec:query}) by saving and retrieving the success rate for each target in goal storage.
The adaptive sampling rate is periodically calculated every 50 update intervals.
From this, SupCon loss $\mathcal{L}_S$ is calculated and the backward updates are made by sampling at the calculated sampling rate for 50 updates for each process.
This is for the purpose of optimizing the learning speed by reducing the number of calculations, and we judge that it does not significantly affect the performance of learning.

\section{Experiments}
\label{app:exp}

\subsection{Additional Results}
\begin{figure}[b]
\centering
    \subfigure[\footnotesize 
    \label{fig:sample_sr} Success Rate of Sampling Methods]{\includegraphics[width=0.47\textwidth]{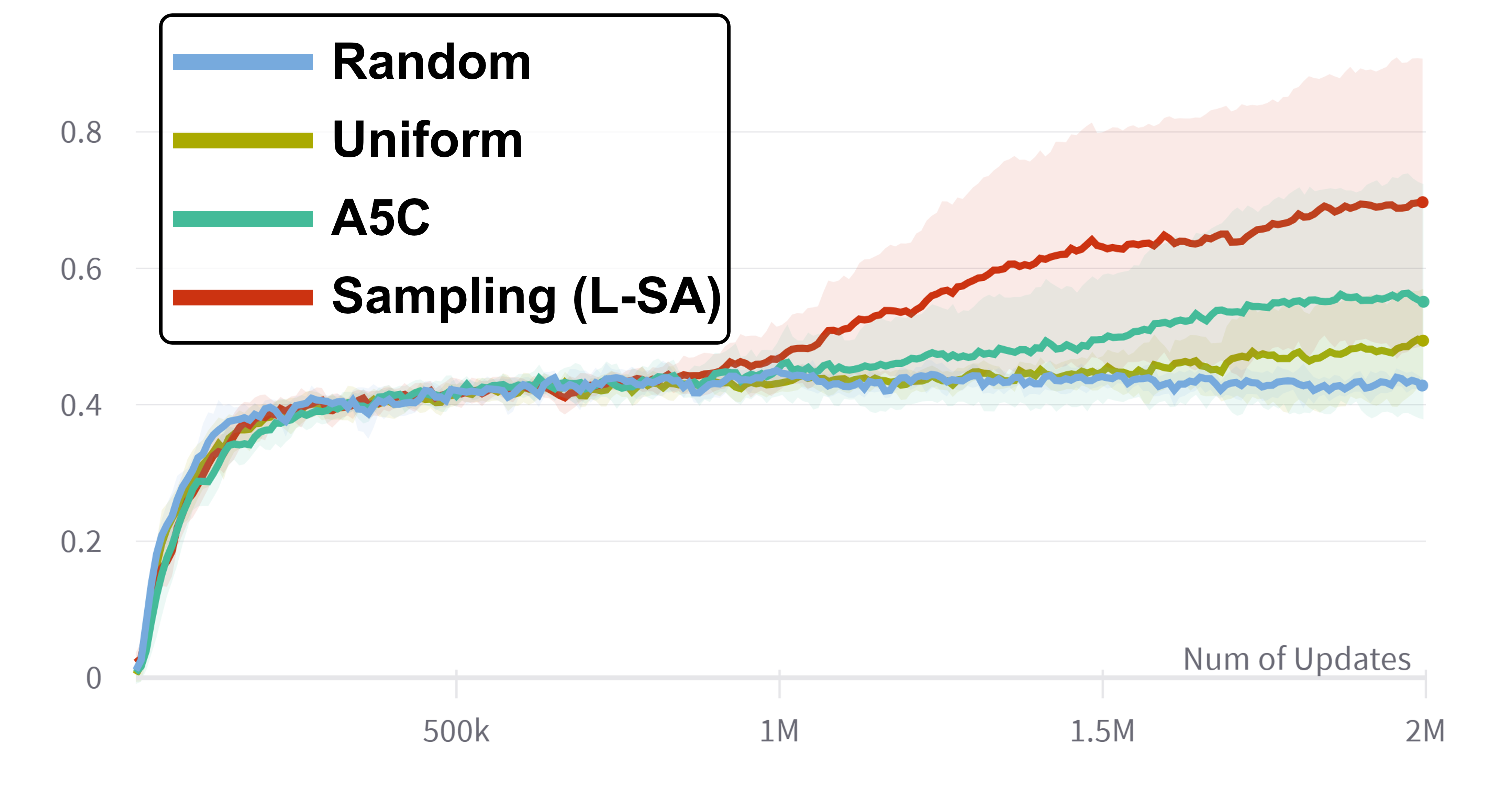}}
    \subfigure[\footnotesize
    \label{fig:sample0}Target 0 (Normal)]{\includegraphics[width=0.23\textwidth]{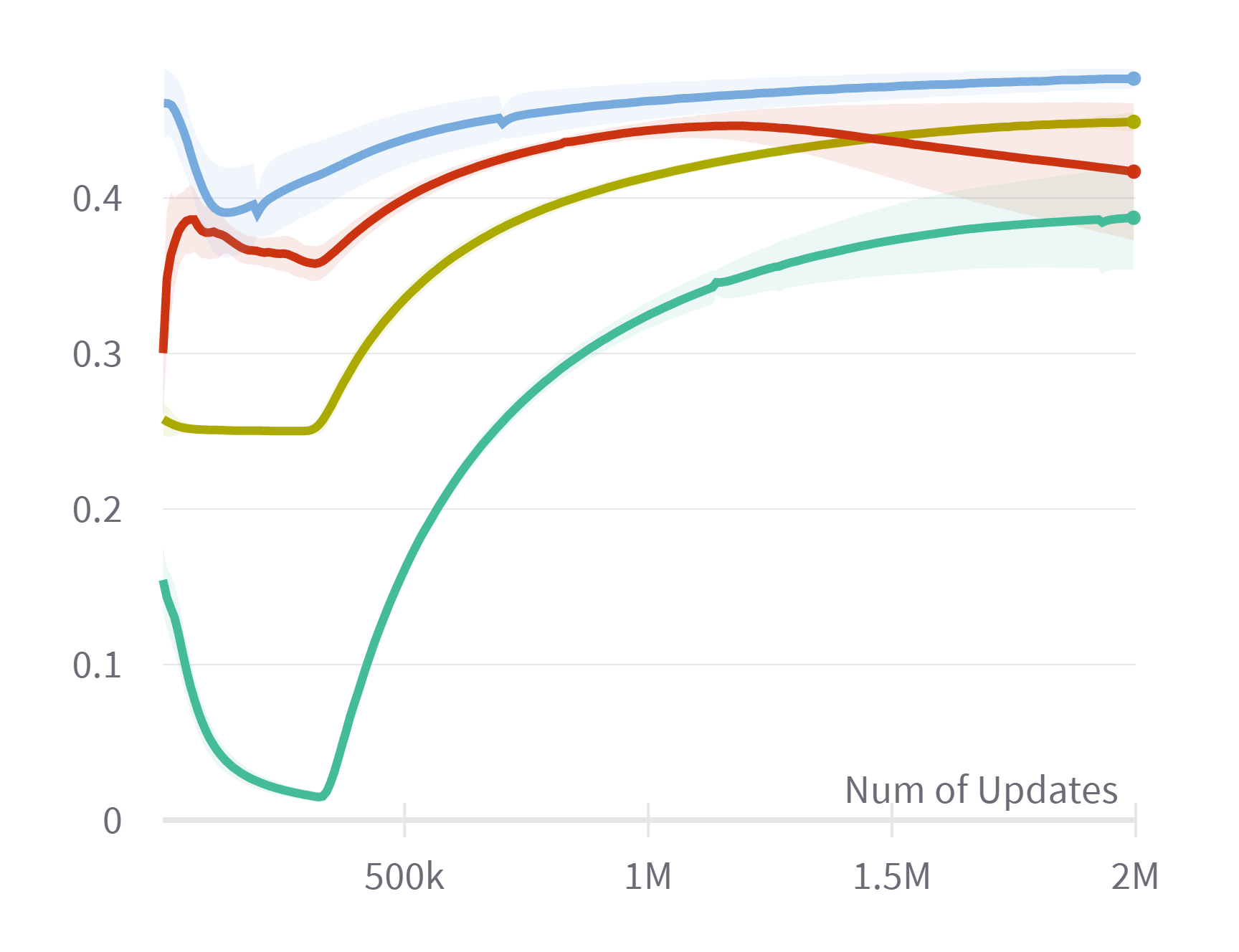}}
    \centering
    \subfigure[\footnotesize
    \label{fig:sample3}Target 3 (Hard)]{\includegraphics[width=0.23\textwidth]{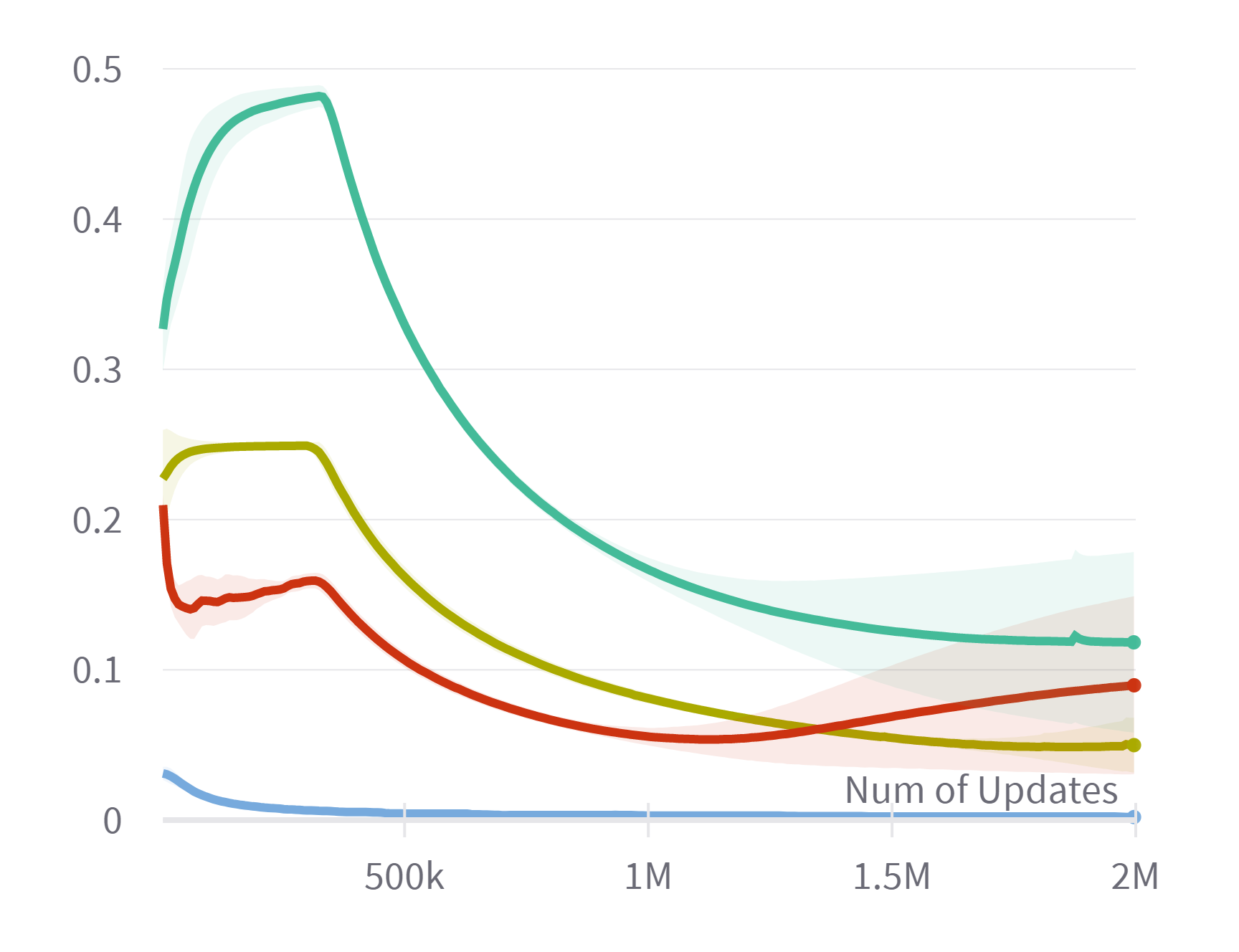}}
    \centering
    \caption{
    Ablation experiments for sampling method.
    (a) displays the curves of success rate and (b) and (c) display the curves of cumulative sampling rates for a normal-difficulty target and a hard-difficulty target, respectively.
    }
\label{fig:app_sample}
\end{figure}

\begin{figure}[b]
\centering
    \subfigure[\footnotesize 
    \label{fig:query_sr} Success Rate of Active Querying Methods]{\includegraphics[width=0.47\textwidth]{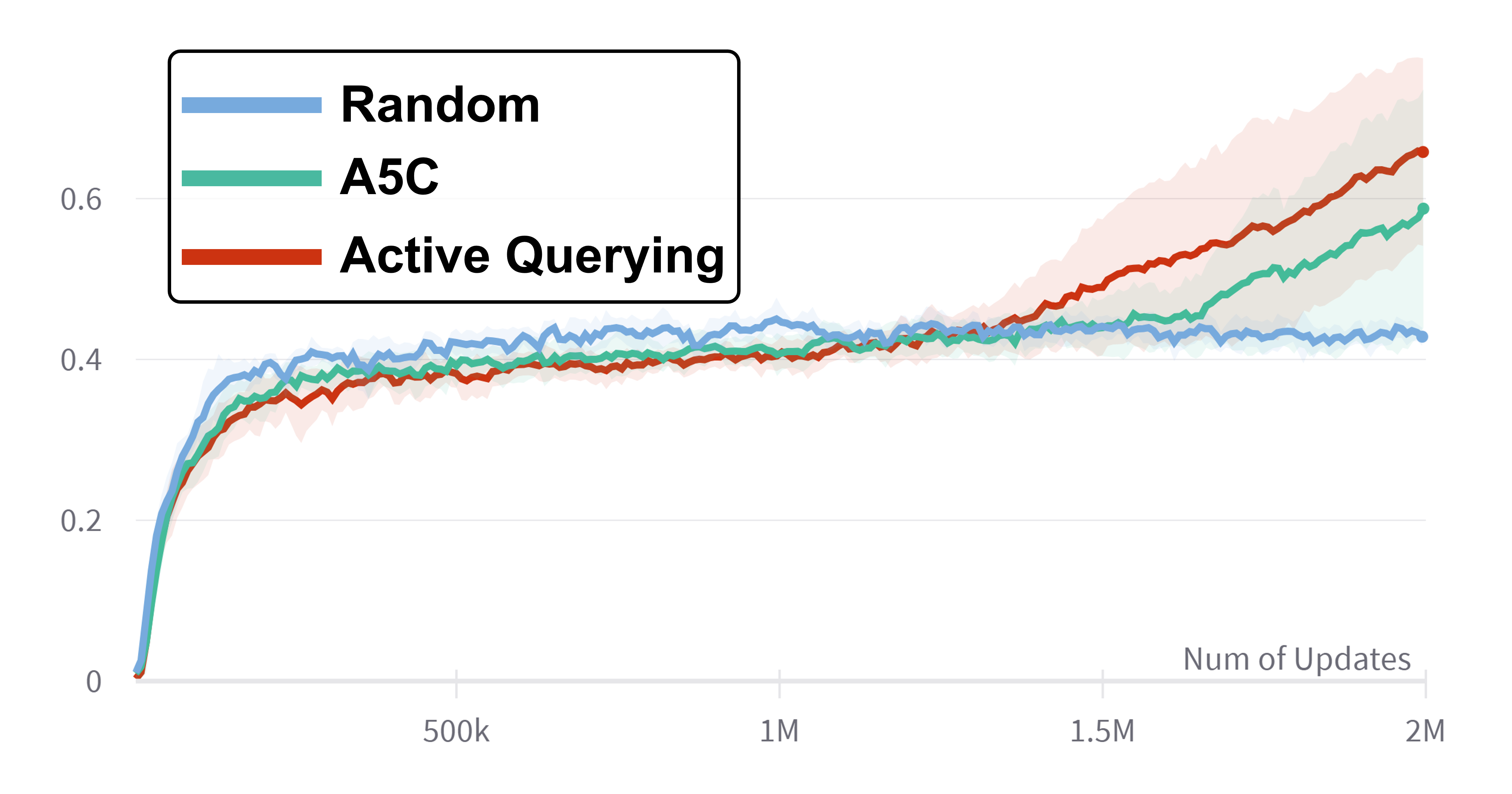}}
    \subfigure[\footnotesize
    \label{fig:query0}Target 0 (Normal)]{\includegraphics[width=0.23\textwidth]{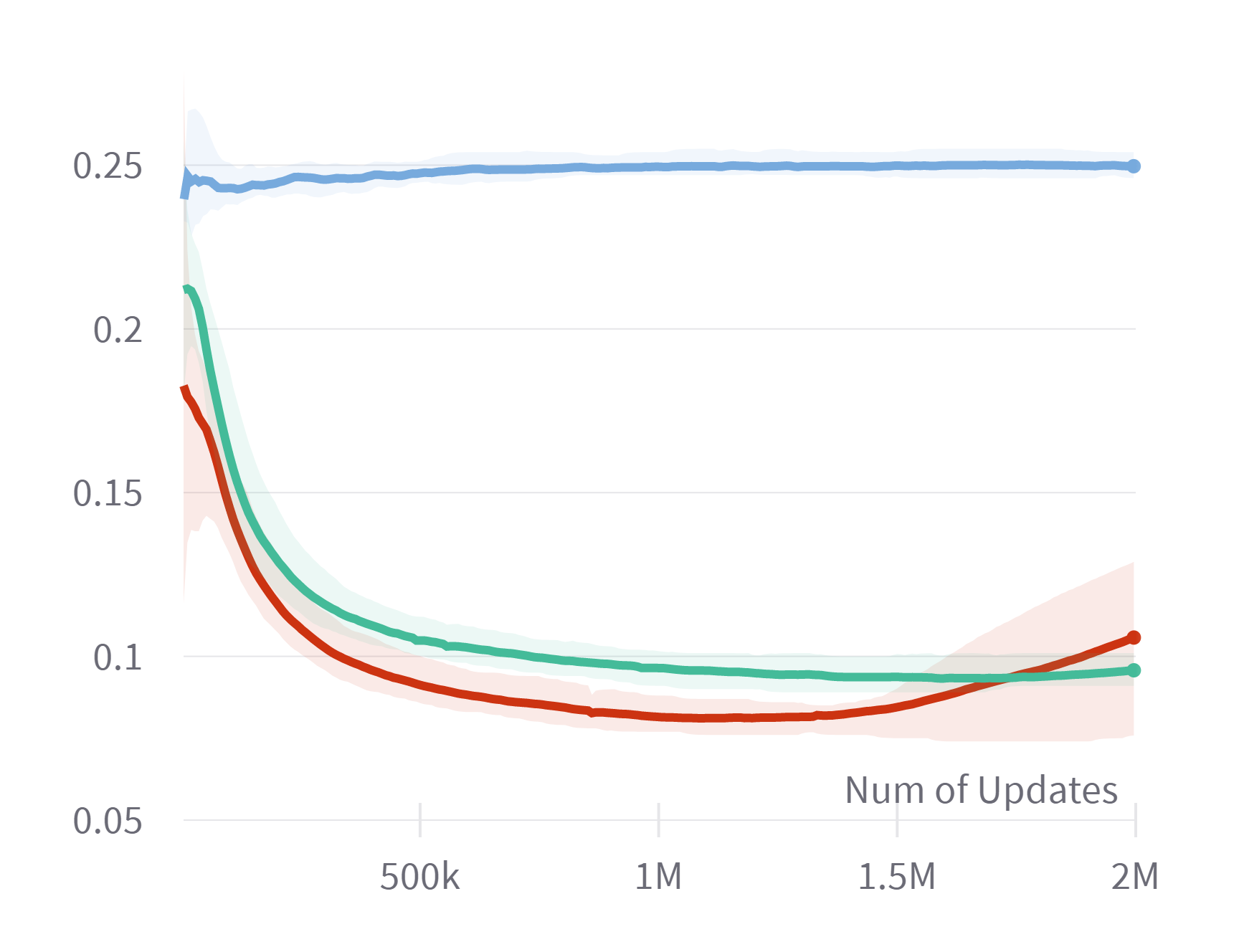}}
    \centering
    \subfigure[\footnotesize
    \label{fig:query3}Target 3 (Hard)]{\includegraphics[width=0.23\textwidth]{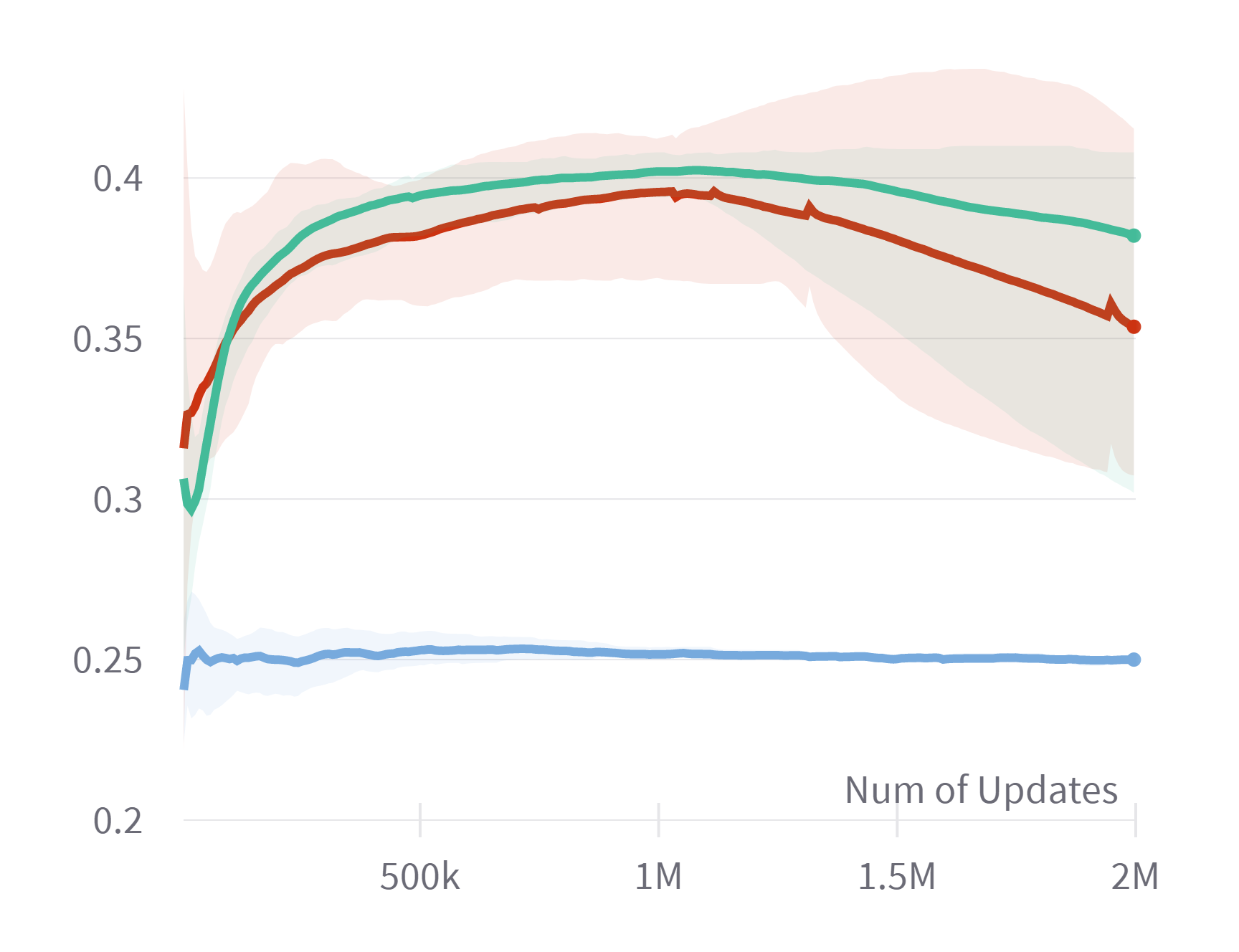}}
    \centering
    \caption{
    Ablation experiments for active querying.
    (a) displays the curves of success rate and (b) and (c) display the curves of cumulative querying rates for a normal-difficulty target and a hard-difficulty target, respectively.
    }
\label{fig:app_query}
\end{figure}

Figures~\ref{fig:app_sample} and \ref{fig:app_query} include ablation studies for sampling and querying that are not included in the main paper due to lack of space.

\begin{figure}[t]
\centering
    \subfigure[\footnotesize
    \label{fig:each_target0}Target 0 (Normal)]{\includegraphics[width=0.23\textwidth]{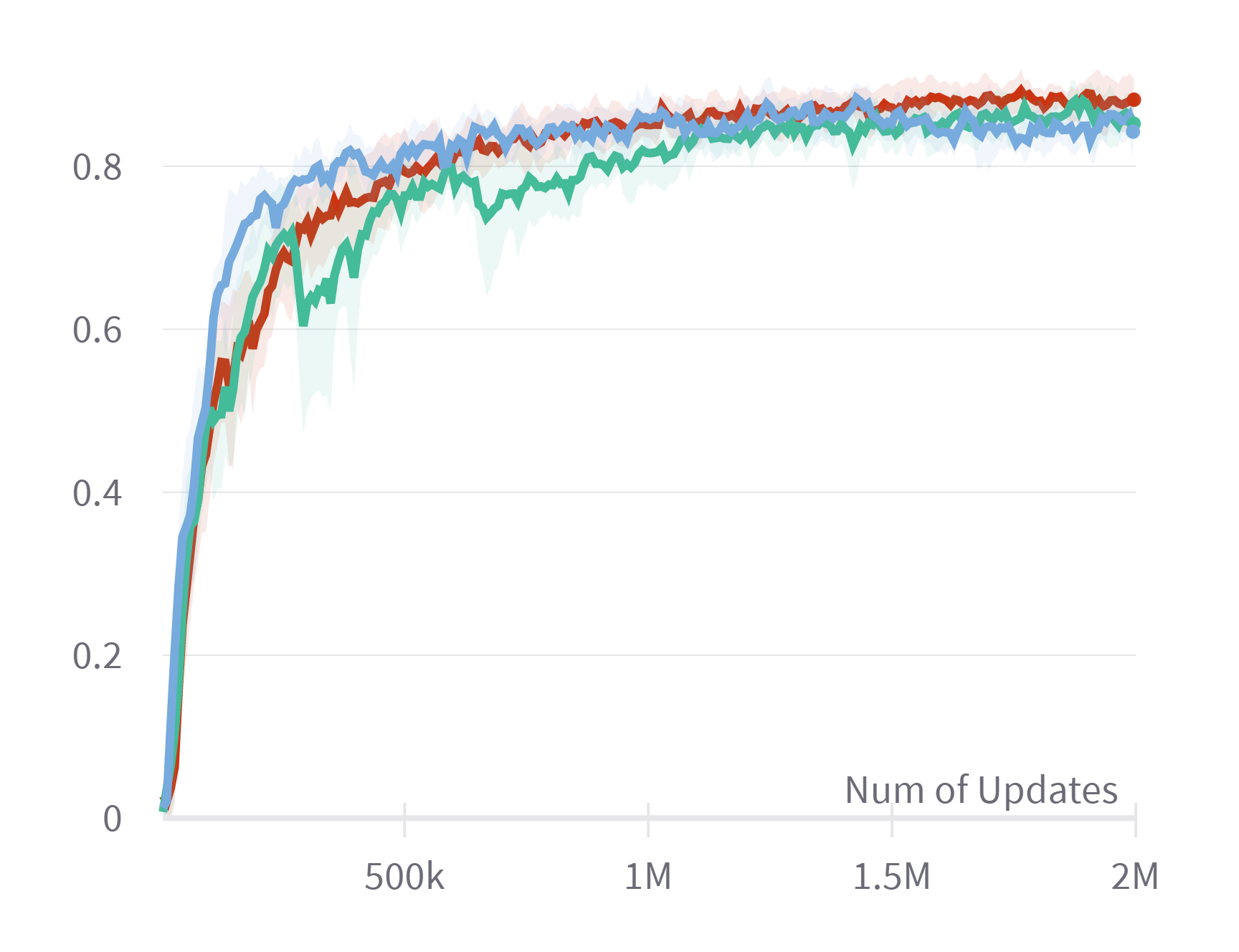}}
    \centering
    \subfigure[\footnotesize
    \label{fig:each_target1}Target 1 (Normal)]{\includegraphics[width=0.23\textwidth]{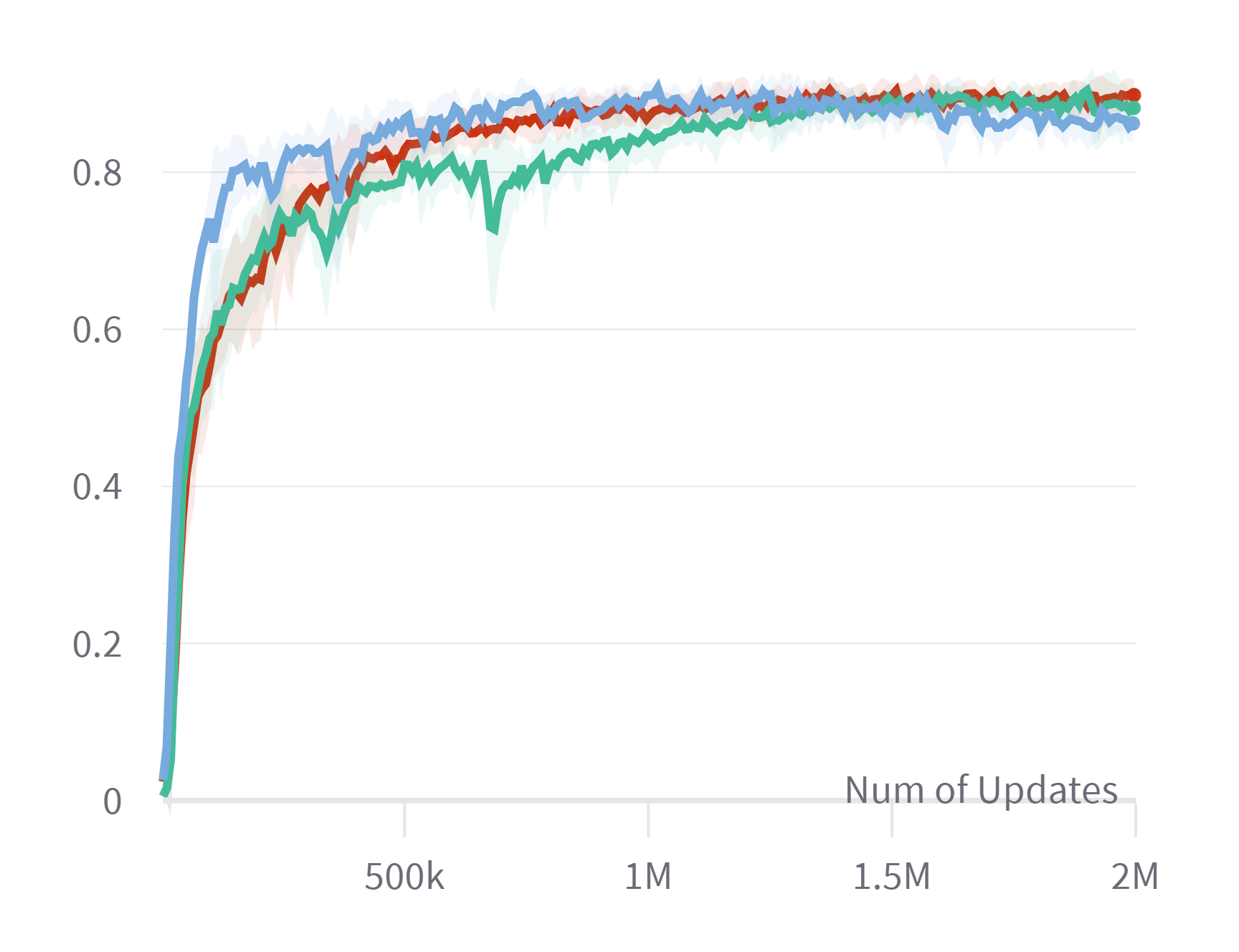}}
    \centering
    \subfigure[\footnotesize
    \label{fig:each_target2}Target 2 (Hard)]{\includegraphics[width=0.23\textwidth]{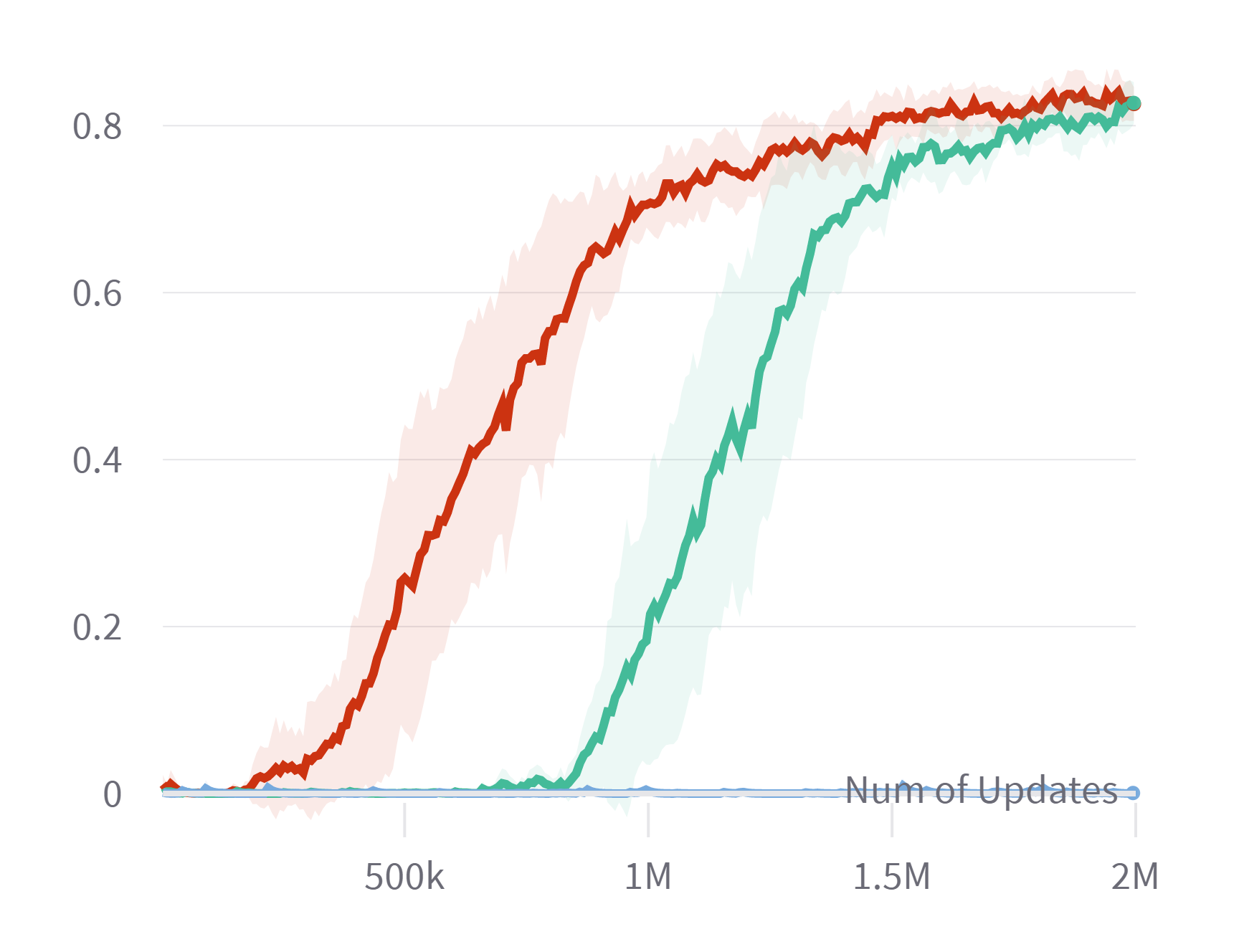}}
    \centering
    \subfigure[\footnotesize
    \label{fig:each_target3}Target 3 (Hard)]{\includegraphics[width=0.23\textwidth]{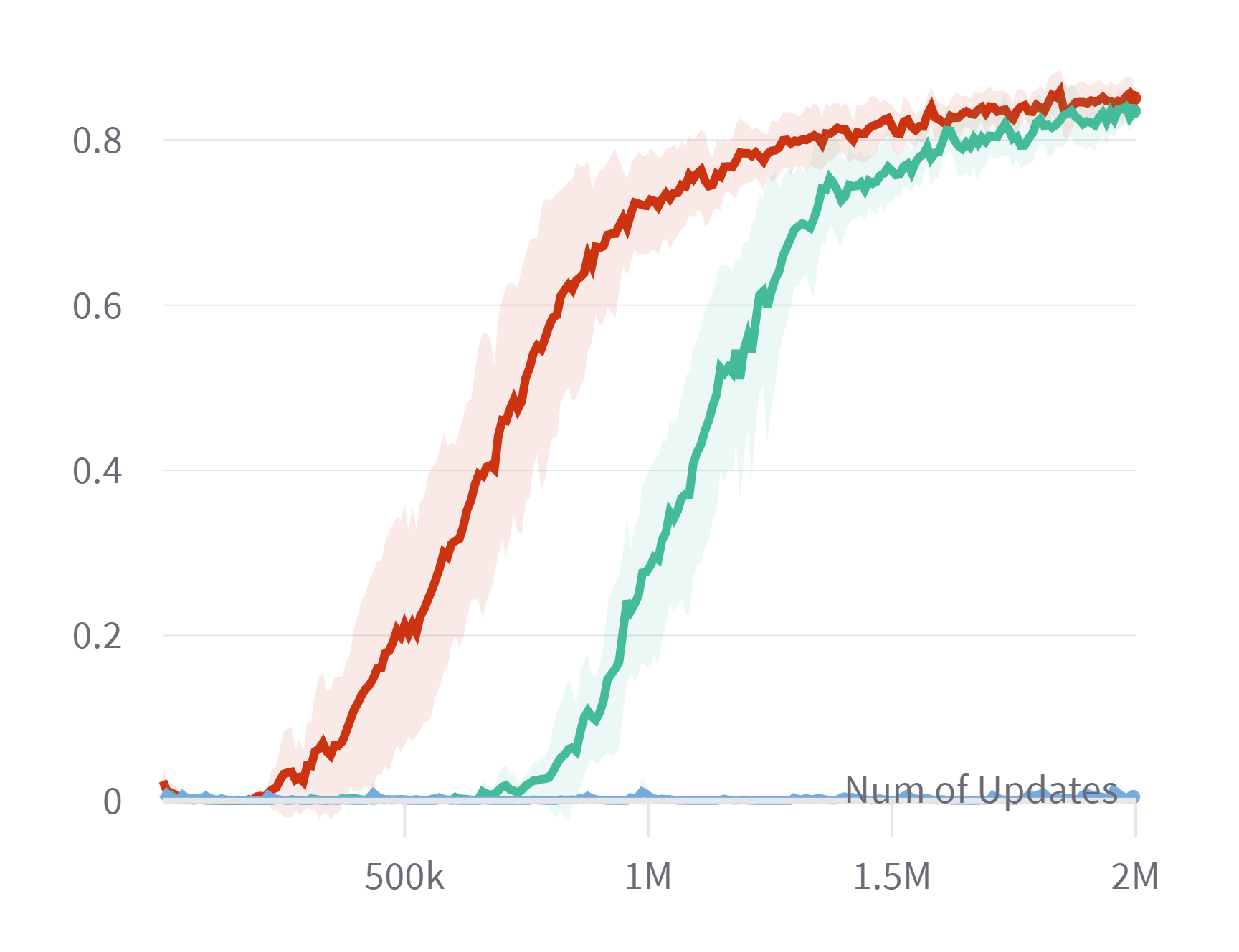}}
    \centering
    \caption{
    The success rate of each target in the Studio-2N 2H task.
    Normal-difficulty targets are learned in all baselines, but only the L-SA and A5C methods are learned in hard-difficulty targets.
    In (c)and (d), L-SA is quickly learned and is sample-efficient.
    }
\label{fig:each_target}
\end{figure}

\begin{figure}[t]
\centering
    \subfigure[\footnotesize
    \label{fig:each_query0}Target 0 (Normal)]{\includegraphics[width=0.23\textwidth]{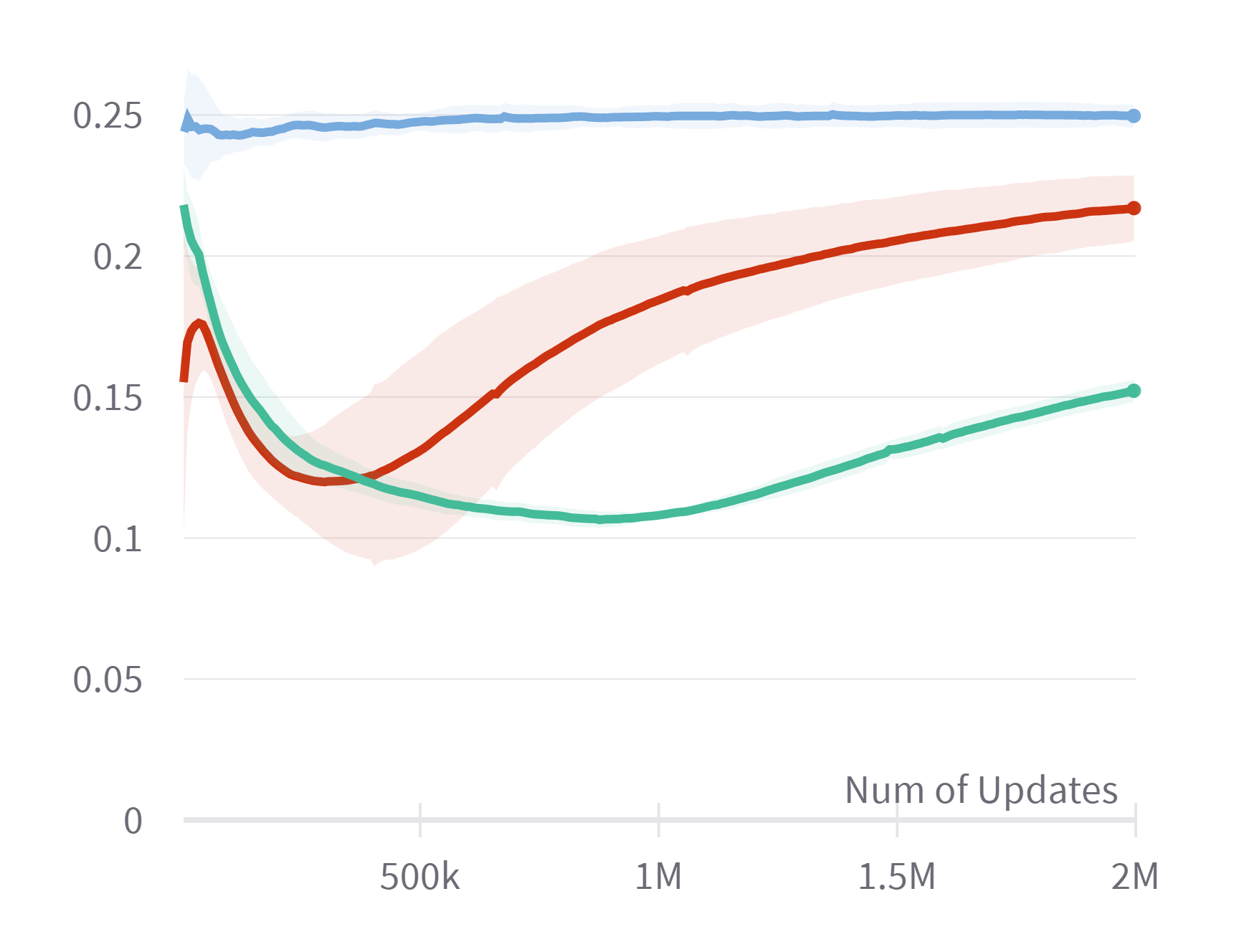}}
    \centering
    \subfigure[\footnotesize
    \label{fig:each_query1}Target 1 (Normal)]{\includegraphics[width=0.23\textwidth]{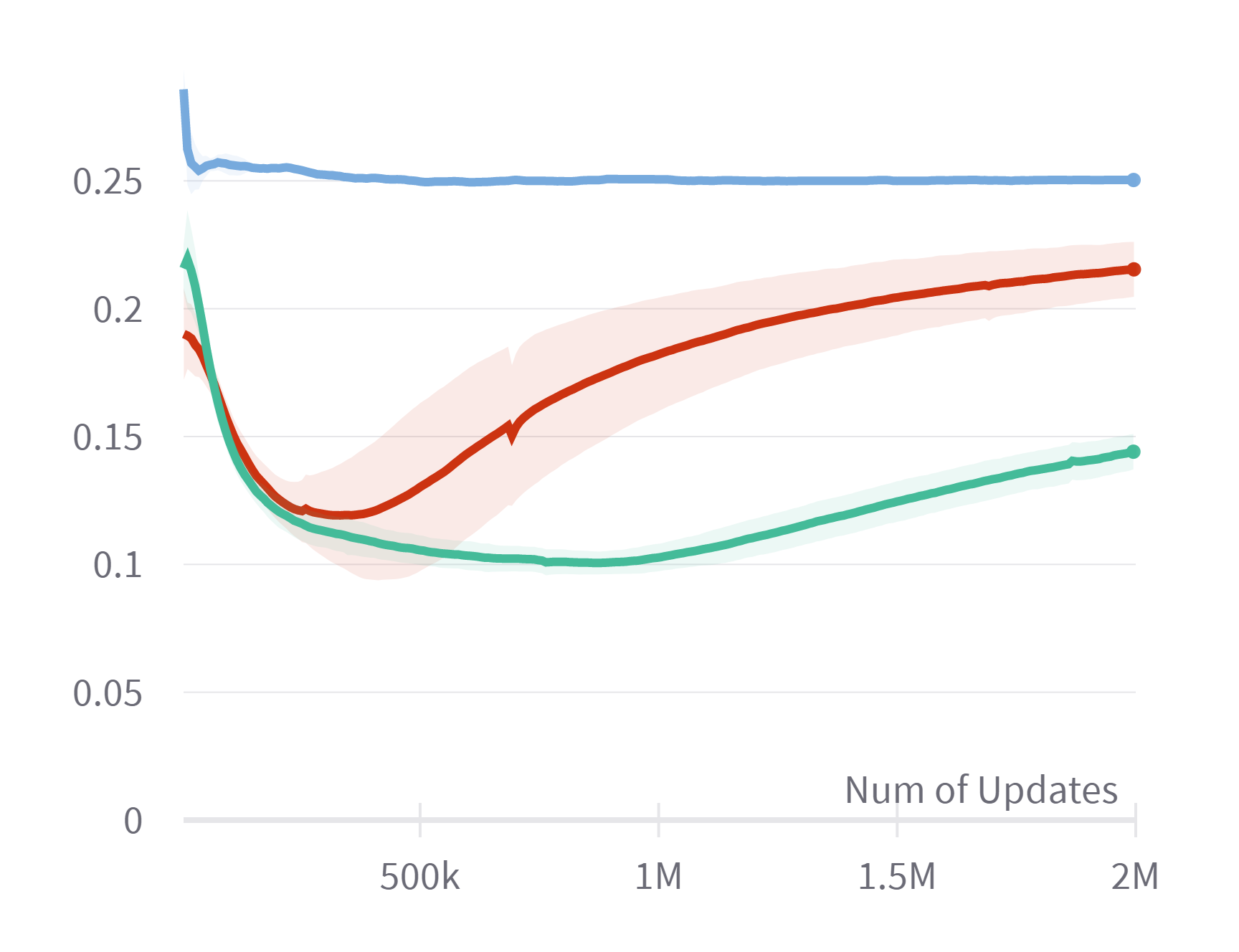}}
    \centering
    \subfigure[\footnotesize
    \label{fig:each_query2}Target 2 (Hard)]{\includegraphics[width=0.23\textwidth]{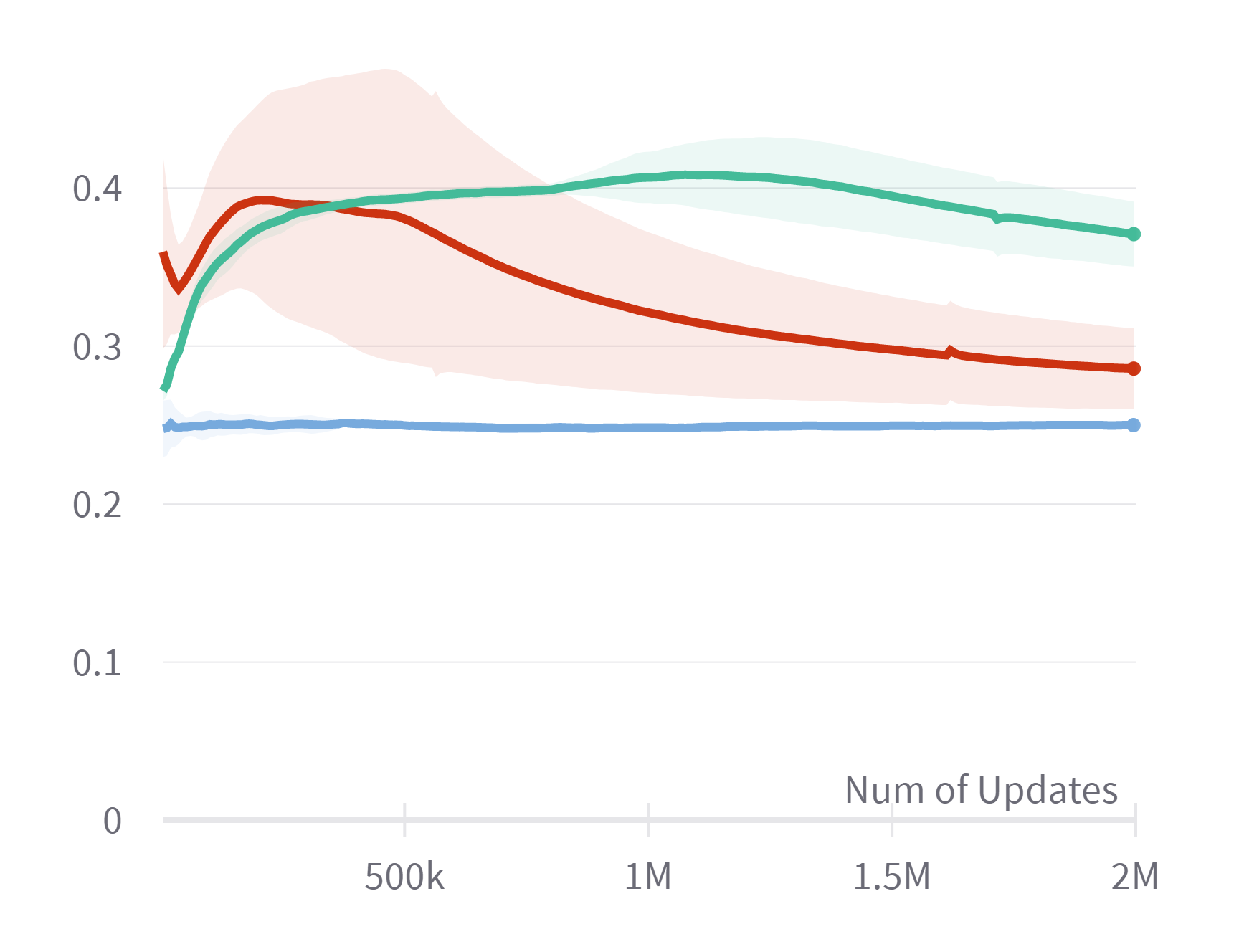}}
    \centering
    \subfigure[\footnotesize
    \label{fig:each_query3}Target 3 (Hard)]{\includegraphics[width=0.23\textwidth]{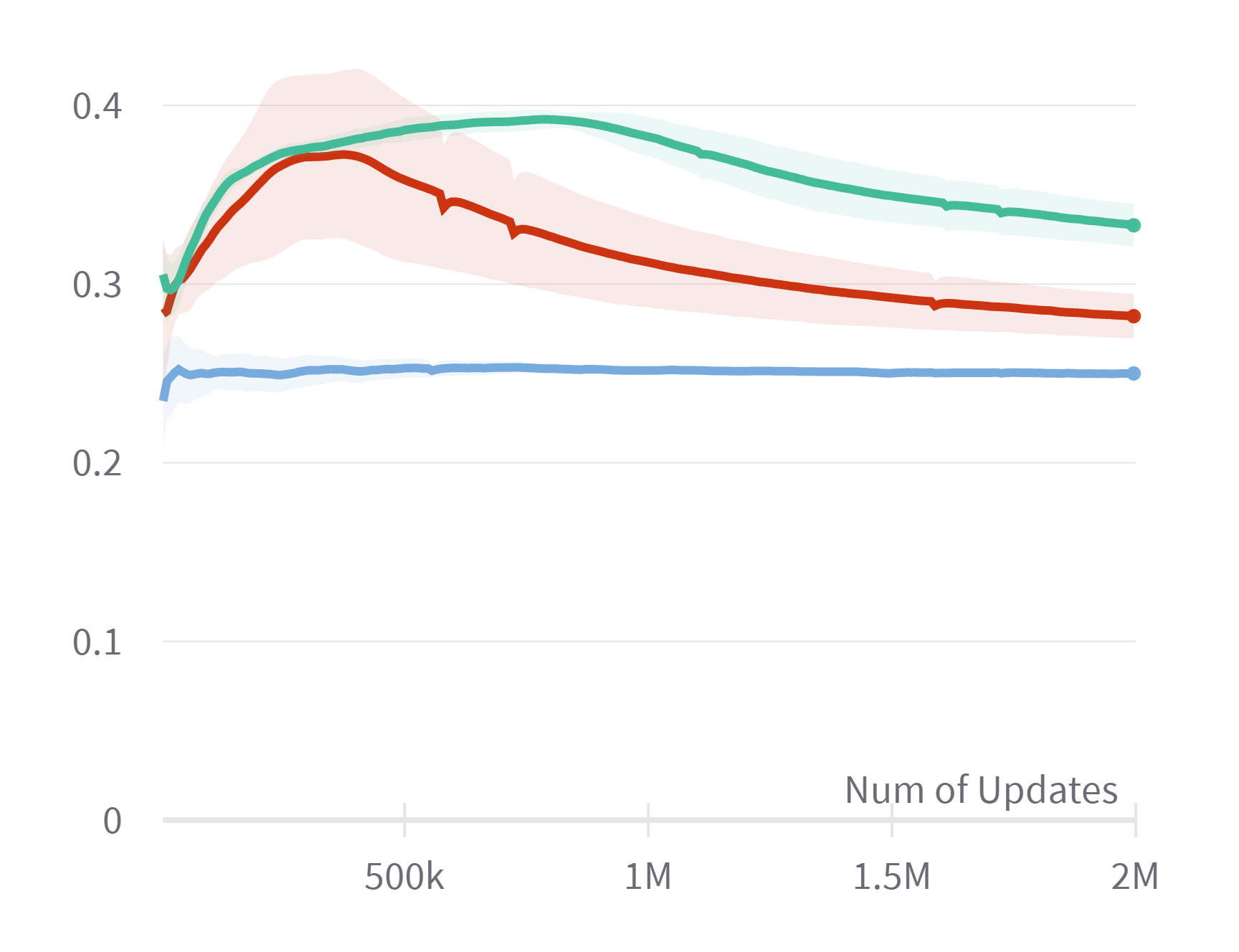}}
    \centering
    \caption{
    The cumulative active querying rate of each target in the Studio-2N 2H task.
    The L-SA framework with active querying rapidly reduces the rate at the hard target. This is because learning is taking place quickly due to the virtuous cycle structure.
    }
\label{fig:each_query}
\end{figure}

\begin{figure}[t]
\centering
    \subfigure[\footnotesize
    \label{fig:each_sample0}Target 0 (Normal)]{\includegraphics[width=0.23\textwidth]{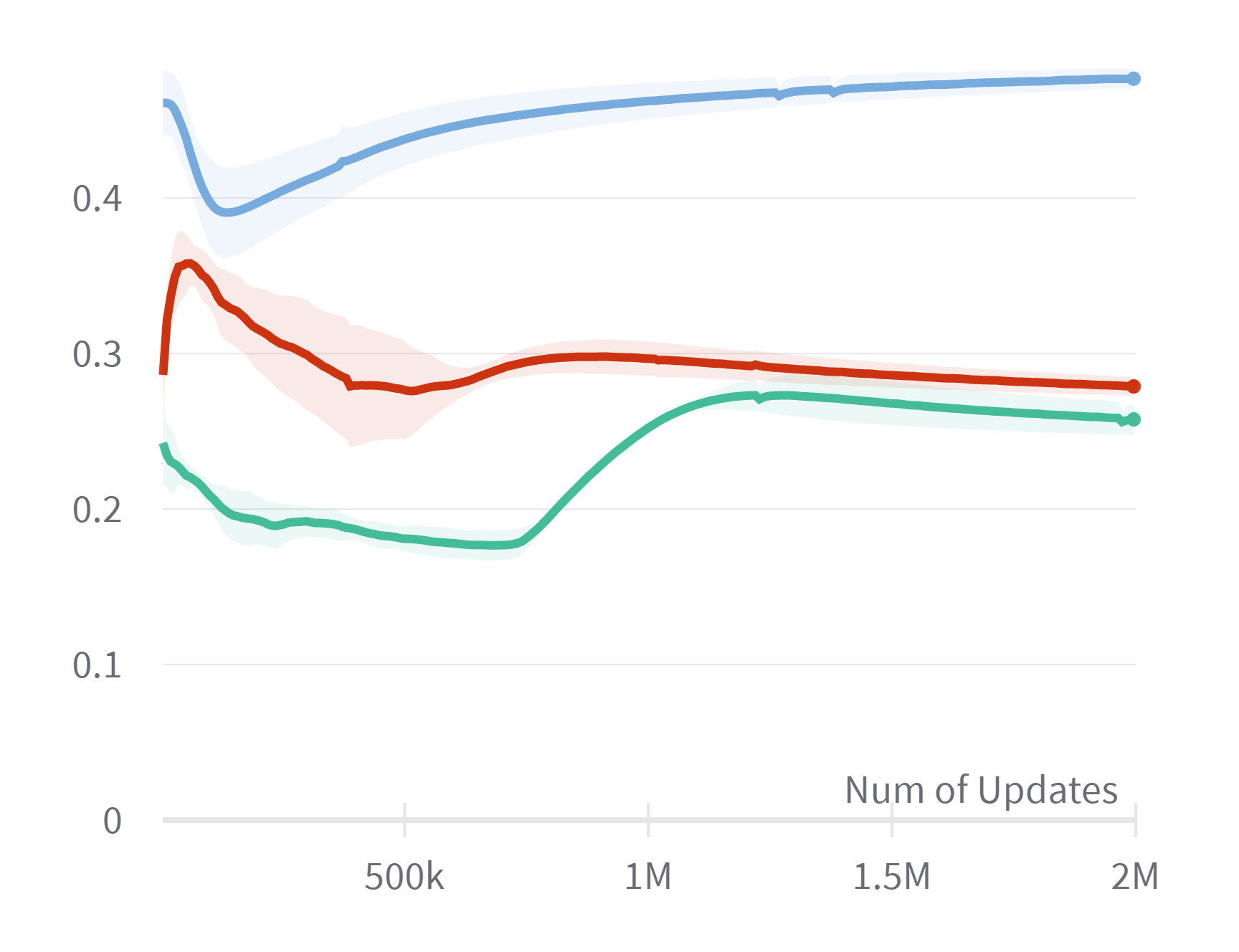}}
    \centering
    \subfigure[\footnotesize
    \label{fig:each_sample1}Target 1 (Normal)]{\includegraphics[width=0.23\textwidth]{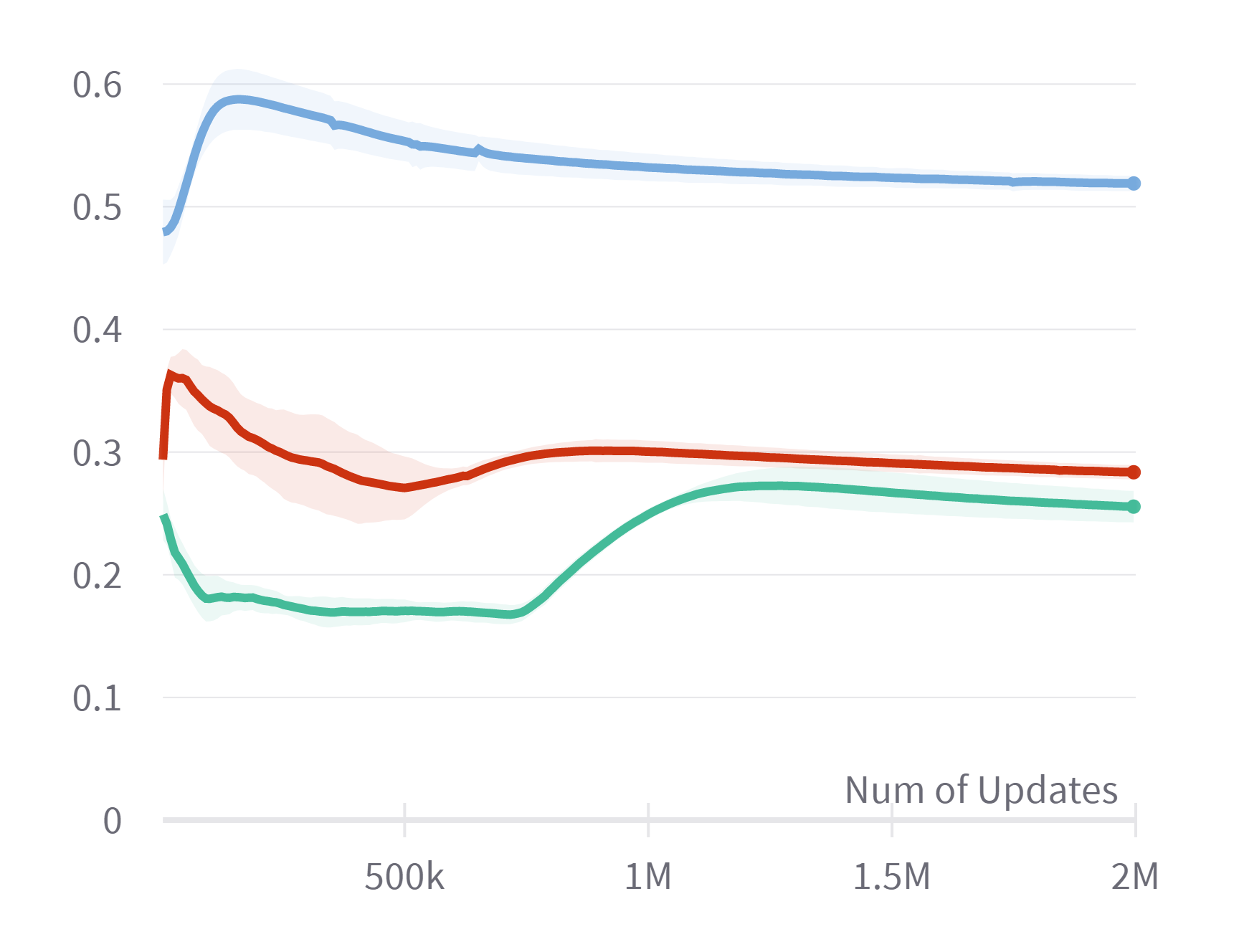}}
    \centering
    \subfigure[\footnotesize
    \label{fig:each_sample2}Target 2 (Hard)]{\includegraphics[width=0.23\textwidth]{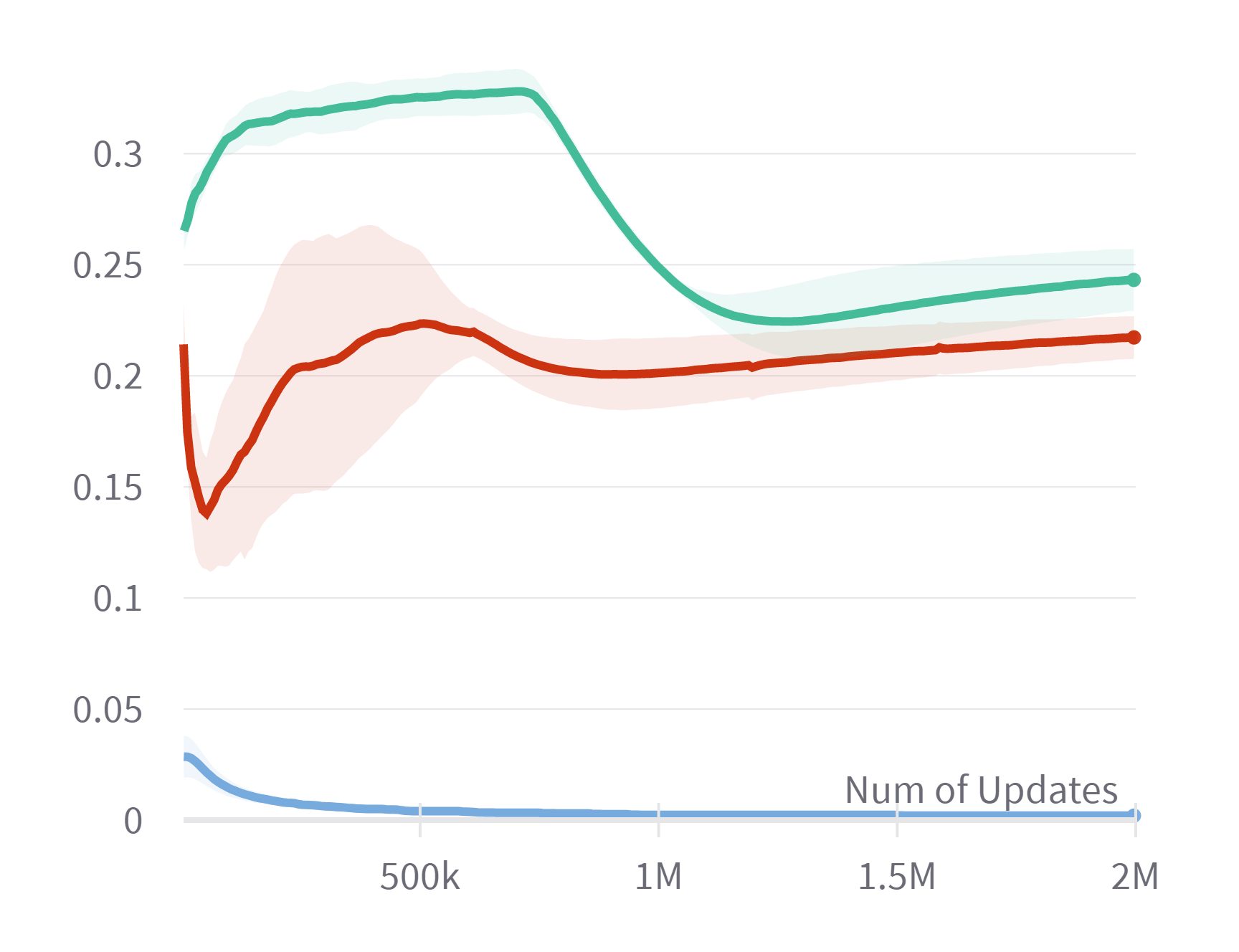}}
    \centering
    \subfigure[\footnotesize
    \label{fig:each_sample3}Target 3 (Hard)]{\includegraphics[width=0.23\textwidth]{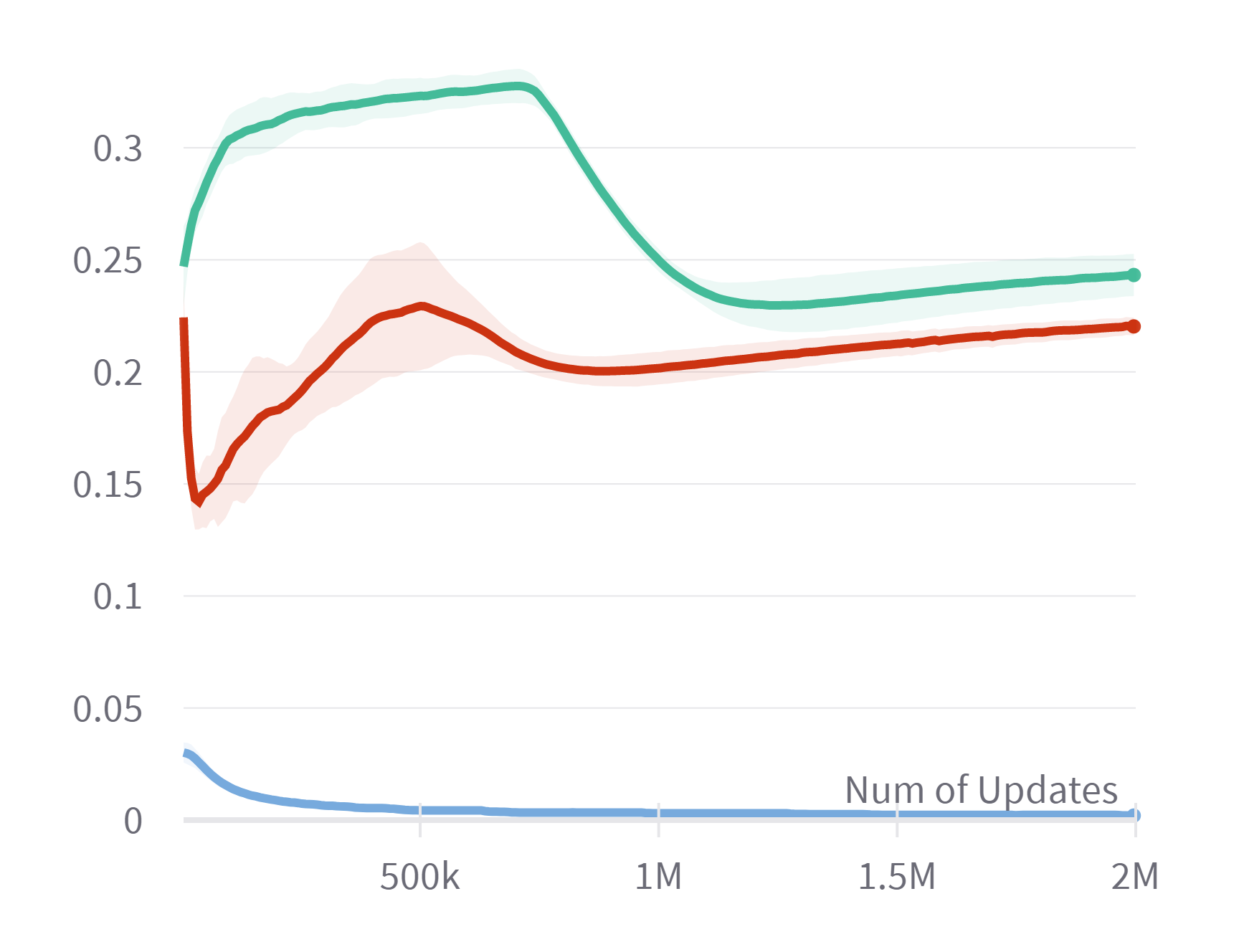}}
    \centering
    \caption{
    The cumulative sampling rate of each target in the Studio-2N 2H task.
    From the beginning of learning, green lines are sampled at a high rate for the hard targets. The red line samples a high proportion of normal-difficulty targets at the start of learning. Sample efficiency is improved by minimizing redundant sampling in the red line.
    }
\label{fig:each_sample}
\end{figure}

\begin{figure}[t]
\centering
    \subfigure[\footnotesize
    \label{fig:each_stor0}Target 0 (Normal)]{\includegraphics[width=0.23\textwidth]{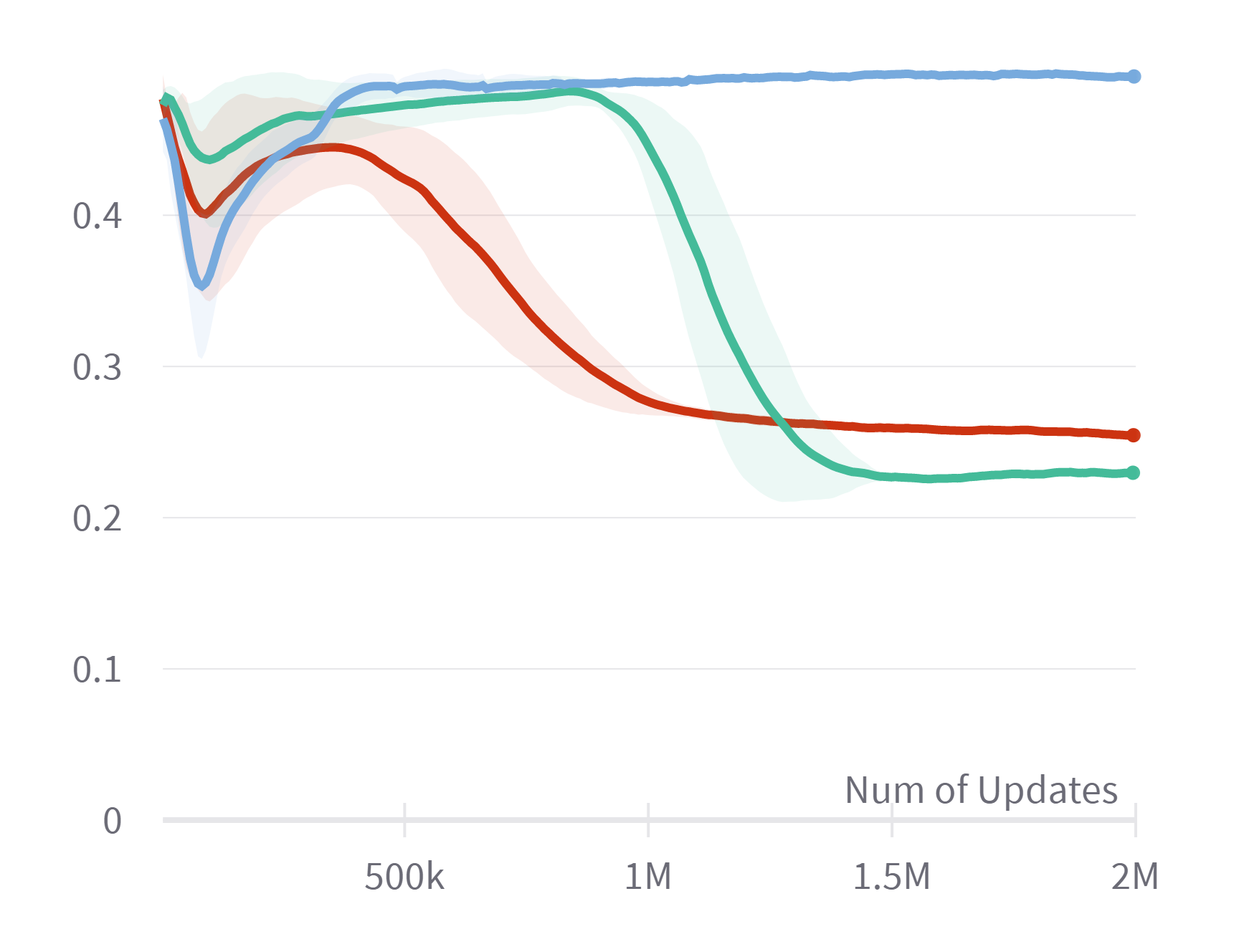}}
    \centering
    \subfigure[\footnotesize
    \label{fig:each_stor1}Target 1 (Normal)]{\includegraphics[width=0.23\textwidth]{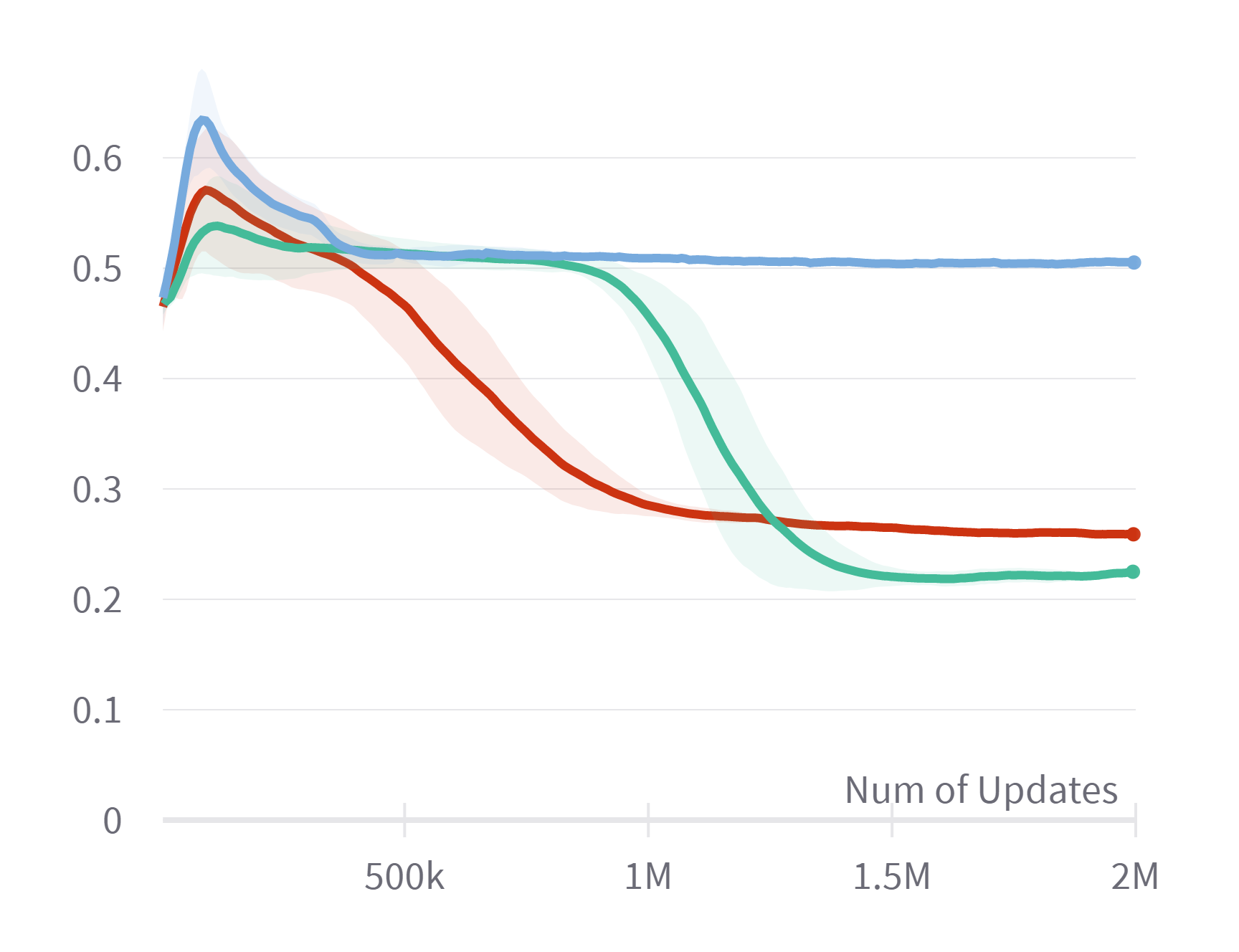}}
    \centering
    \subfigure[\footnotesize
    \label{fig:each_stor2}Target 2 (Hard)]{\includegraphics[width=0.23\textwidth]{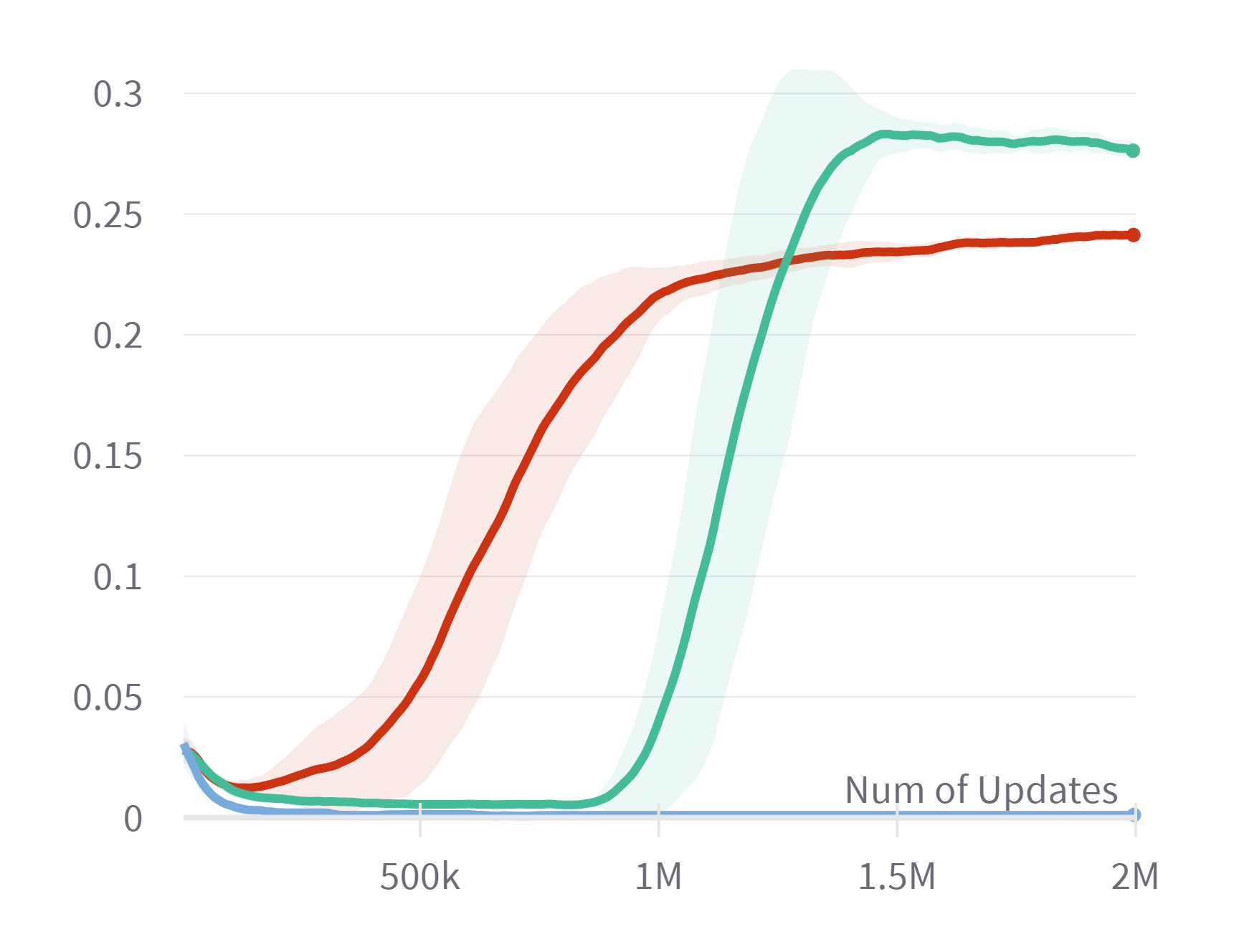}}
    \centering
    \subfigure[\footnotesize
    \label{fig:each_stor3}Target 3 (Hard)]{\includegraphics[width=0.23\textwidth]{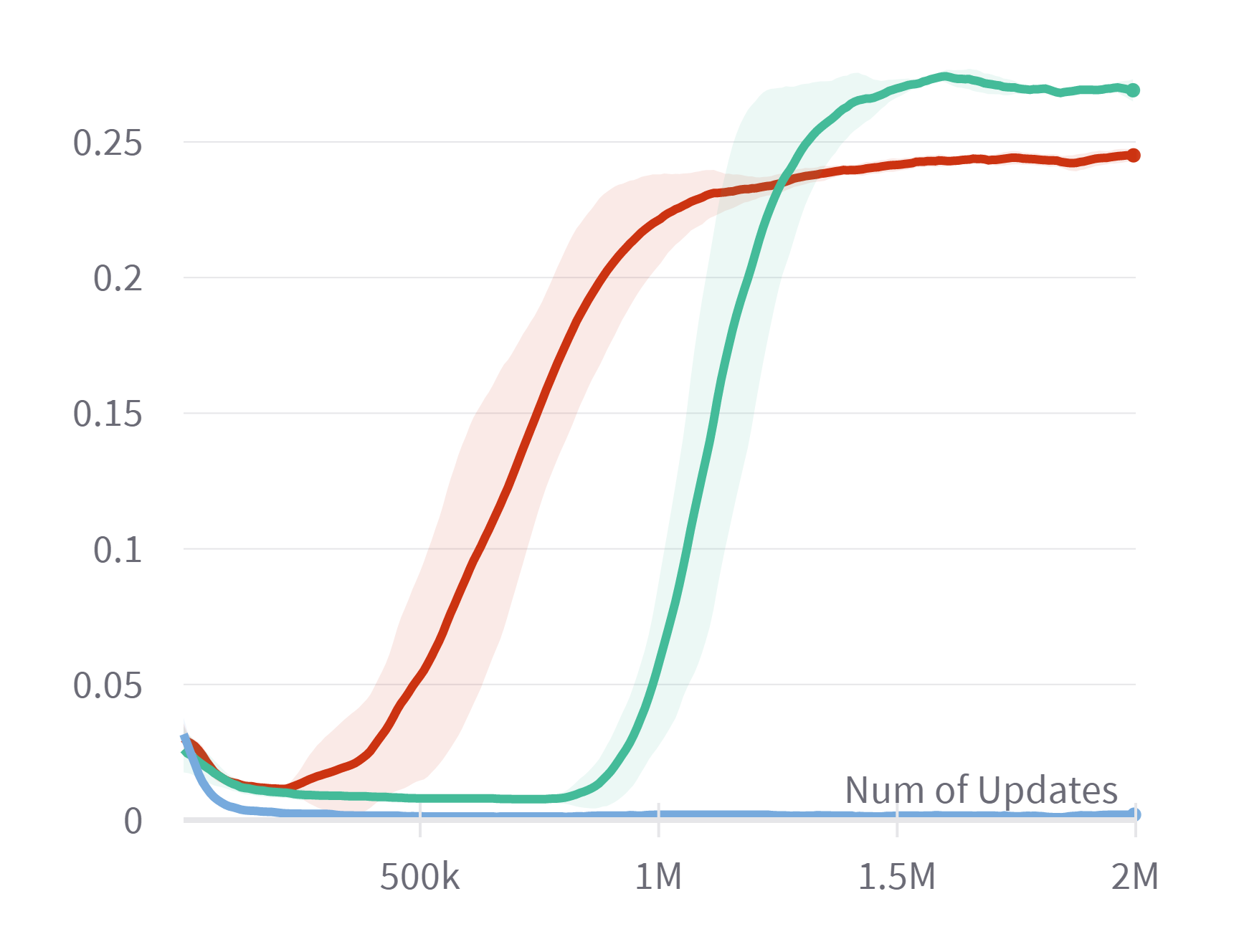}}
    \centering
    \caption{
    The storage rate of each target in the Studio-2N 2H task.
    All methods have a high ratio of normal targets at the beginning of learning. However, our method (red line) increases the rate of the hard targets quickly. This is due to the virtuous cycle structure that minimizes redundant sampling of the L-SA framework.
    }
\label{fig:each_stor}
\end{figure}

\begin{figure}[t]
\centering
    \subfigure[\footnotesize
    \label{fig:t0_6E}Target 0]{\includegraphics[width=0.3\textwidth]{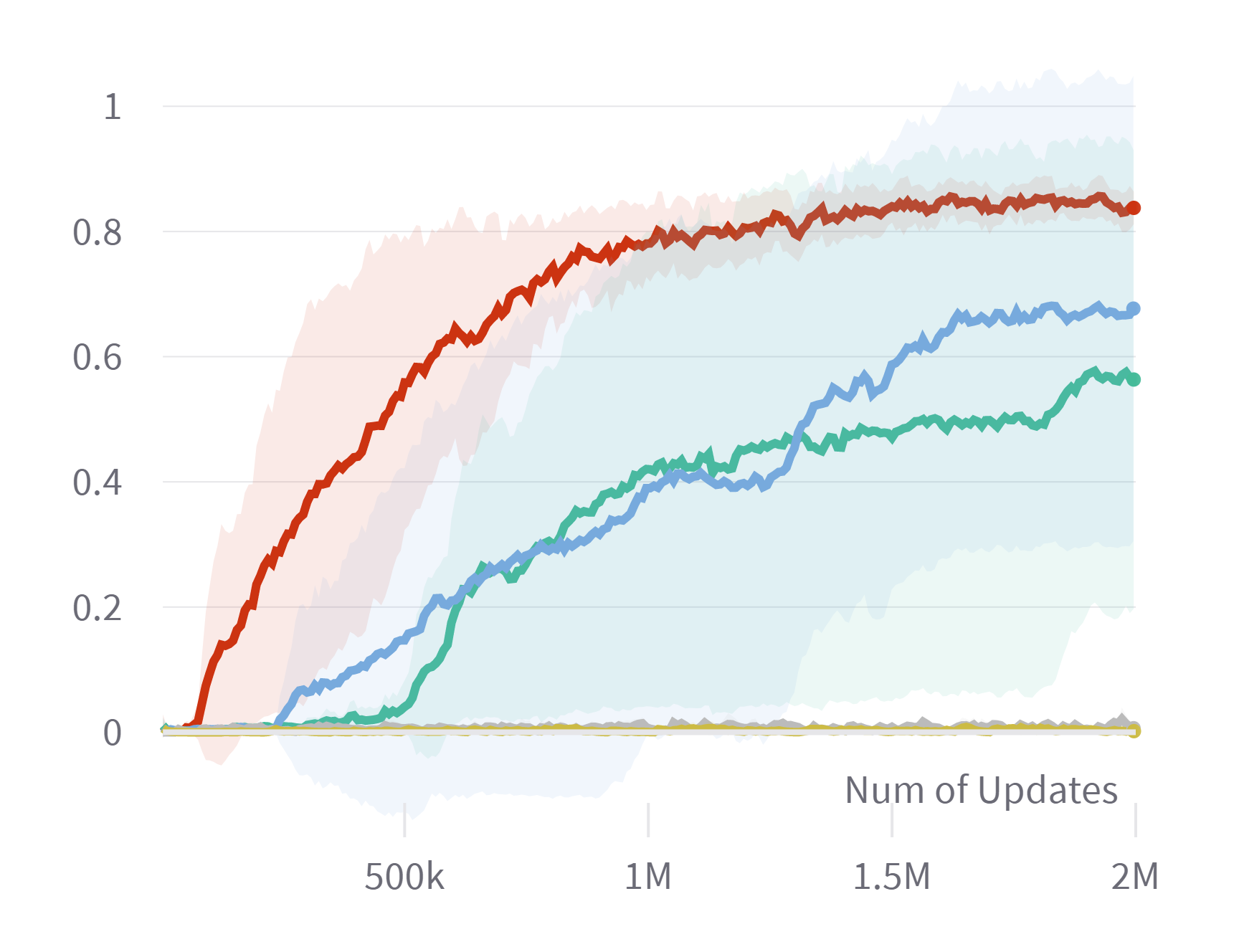}}
    \centering
    \subfigure[\footnotesize
    \label{fig:t1_6E}Target 1]{\includegraphics[width=0.3\textwidth]{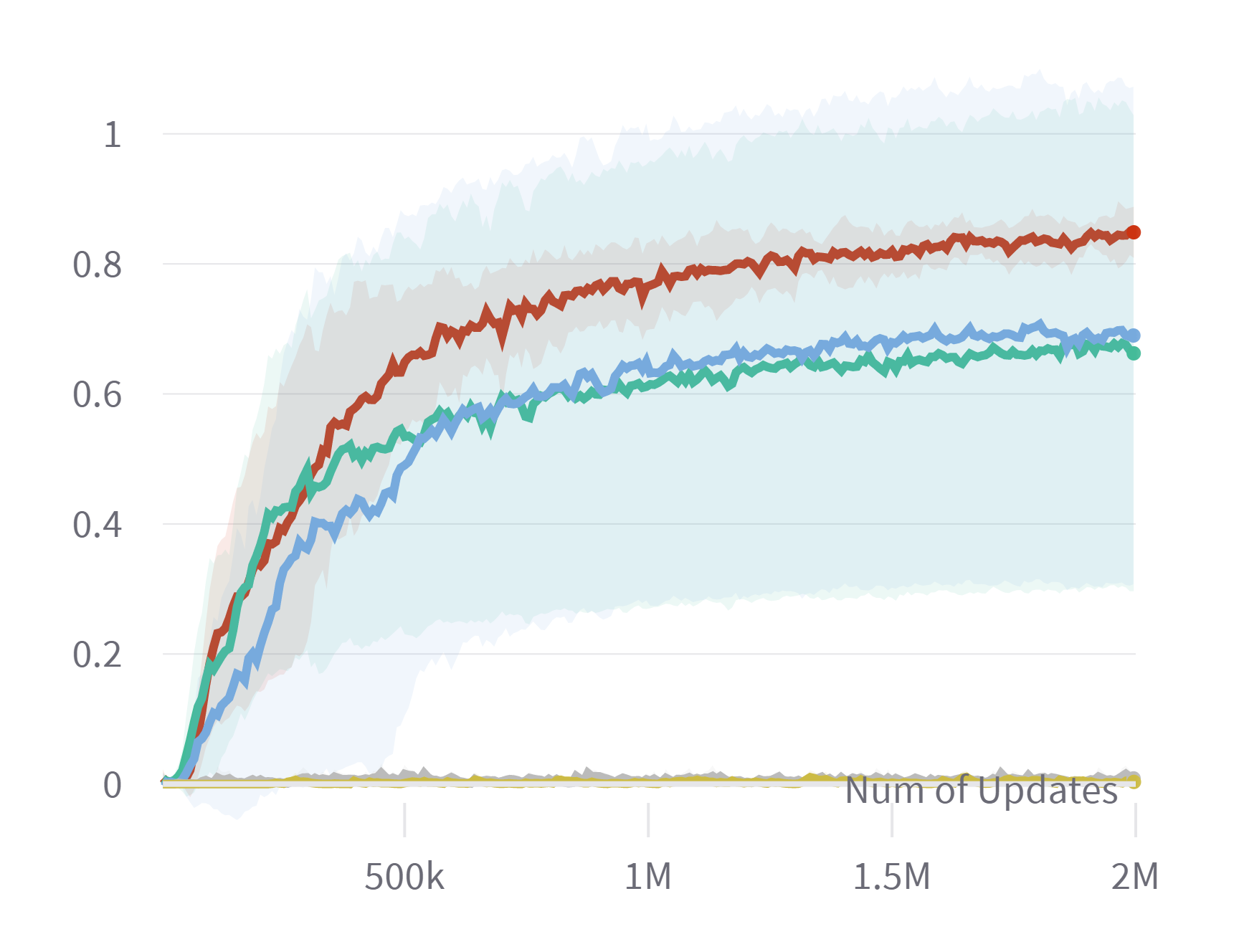}}
    \centering
    \subfigure[\footnotesize
    \label{fig:t2_6E}Target 2]{\includegraphics[width=0.3\textwidth]{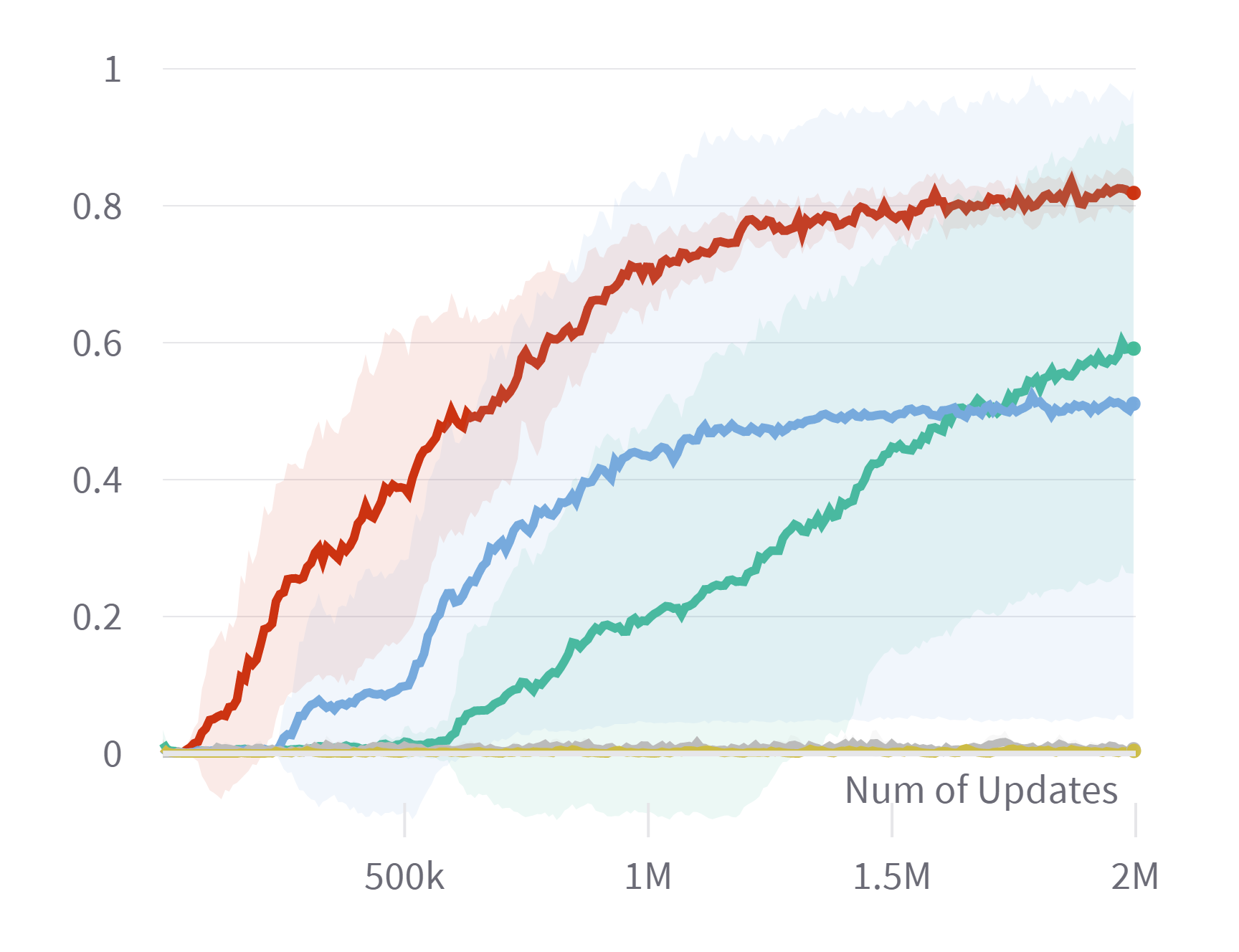}}
    \centering
    \\
    \subfigure[\footnotesize
    \label{fig:t3_6E}Target 3]{\includegraphics[width=0.3\textwidth]{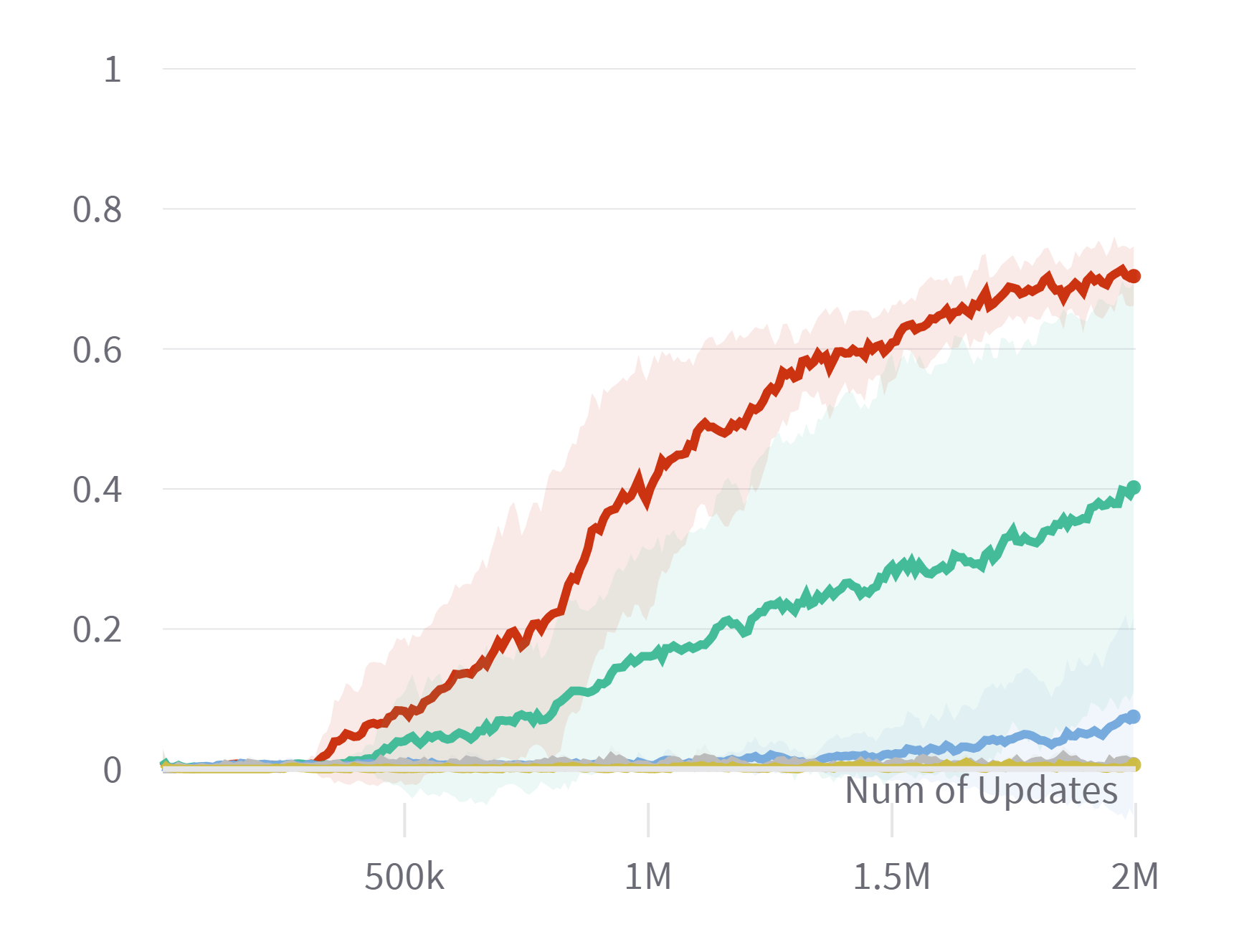}}
    \centering    
    \subfigure[\footnotesize
    \label{fig:t4_6E}Target 4]{\includegraphics[width=0.3\textwidth]{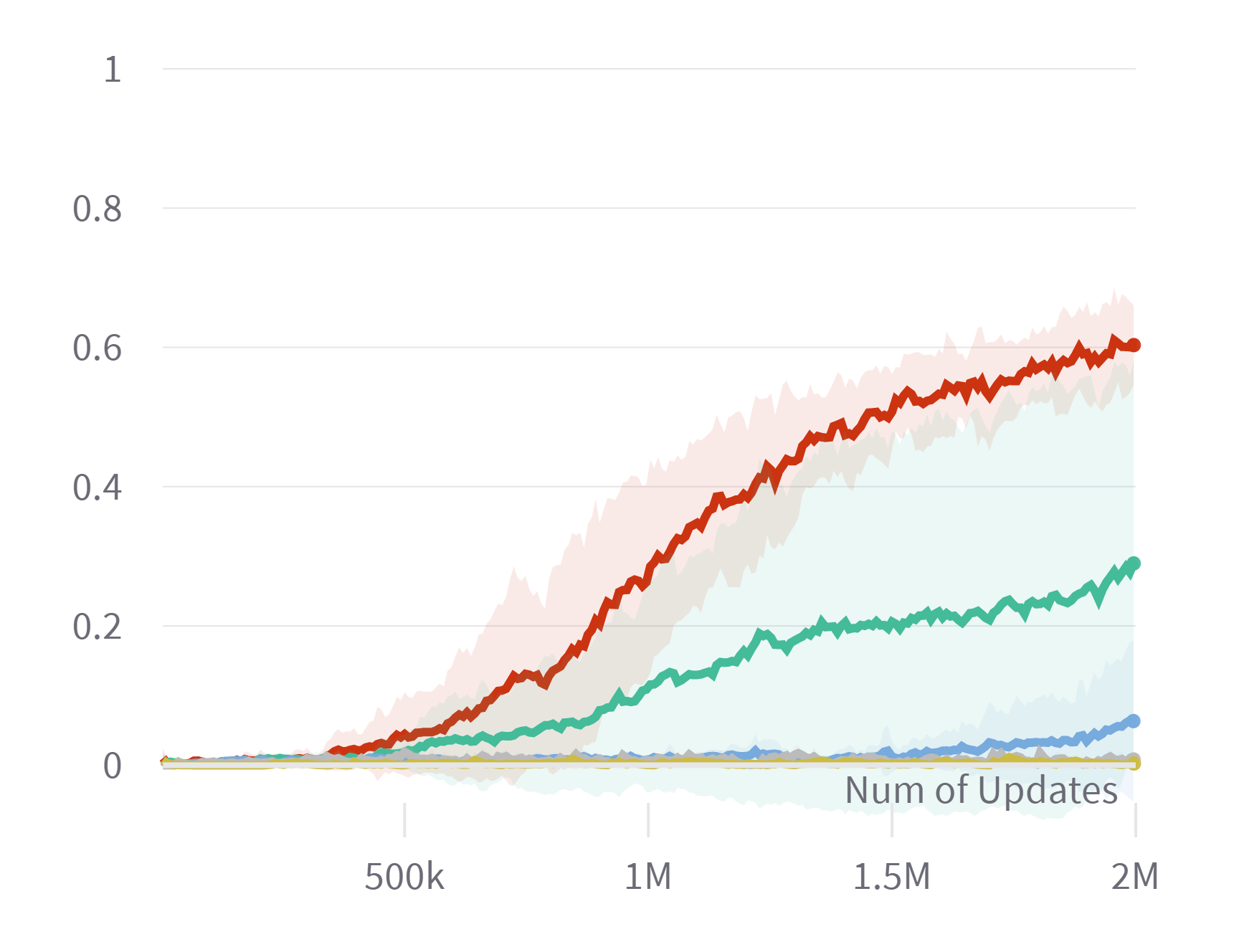}}
    \centering
    \subfigure[\footnotesize
    \label{fig:t5_6E}Target 5]{\includegraphics[width=0.3\textwidth]{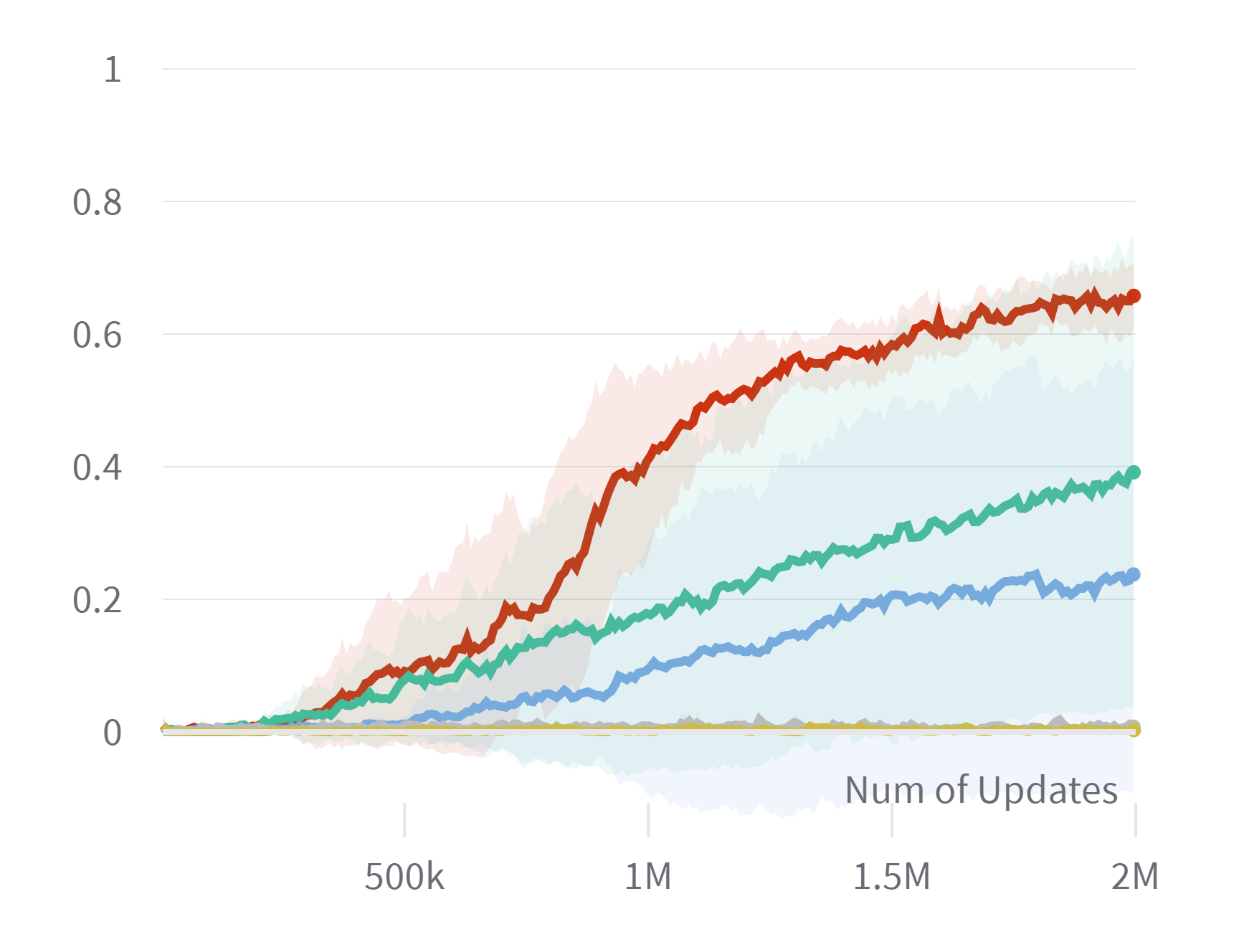}}
    \centering
    \caption{
    The success rate for each target in the Studio-6N task.
    For all targets, the red line draws high and steep learning curves.
    All baseline models show high standard deviation due to instability.
    }
\label{fig:each_target_6e}
\end{figure}

\begin{figure}[t]
\centering
    \subfigure[\footnotesize
    \label{fig:value_t0_6E}Target 0]{\includegraphics[width=0.3\textwidth]{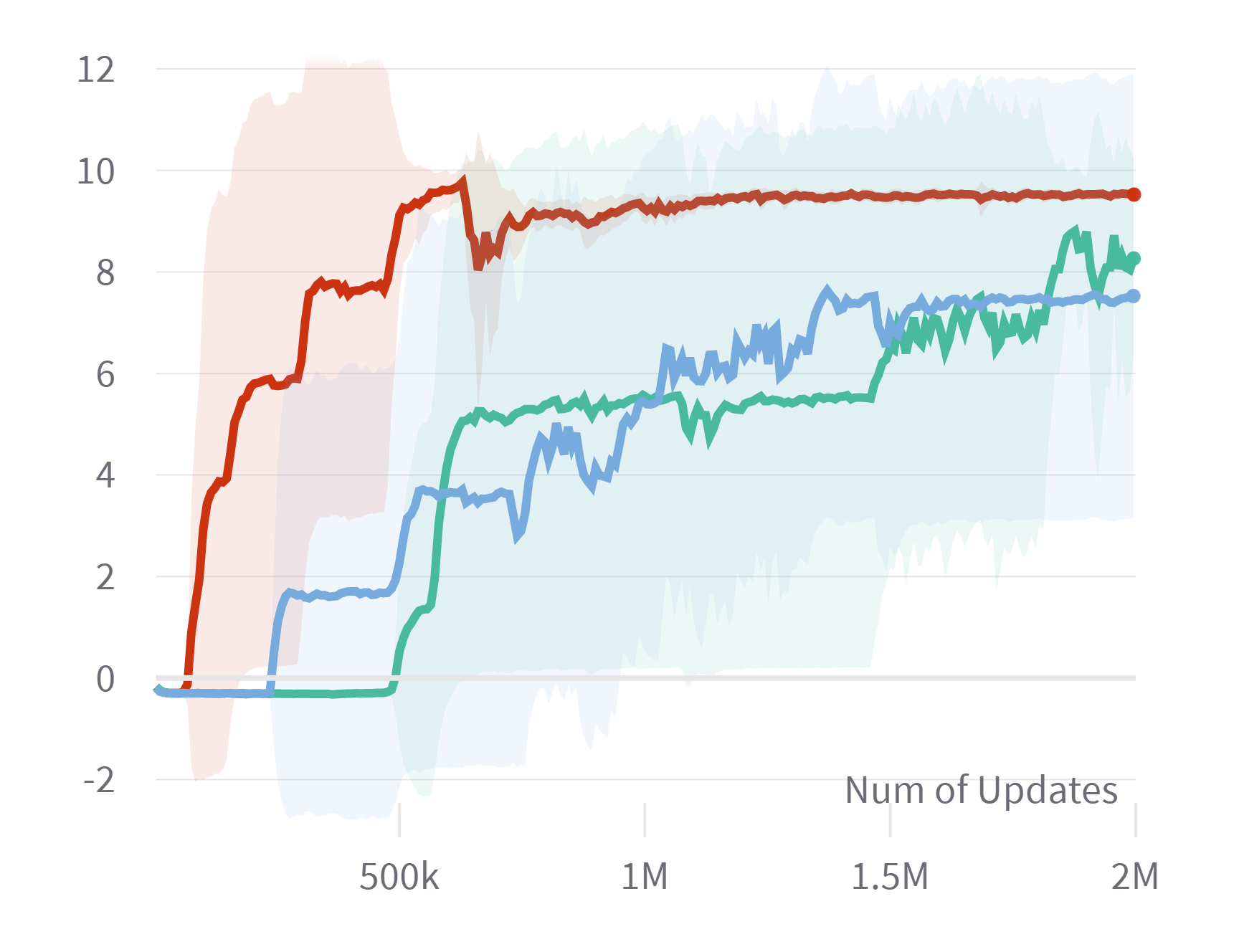}}
    \centering
    \subfigure[\footnotesize
    \label{fig:value_t1_6E}Target 1]{\includegraphics[width=0.3\textwidth]{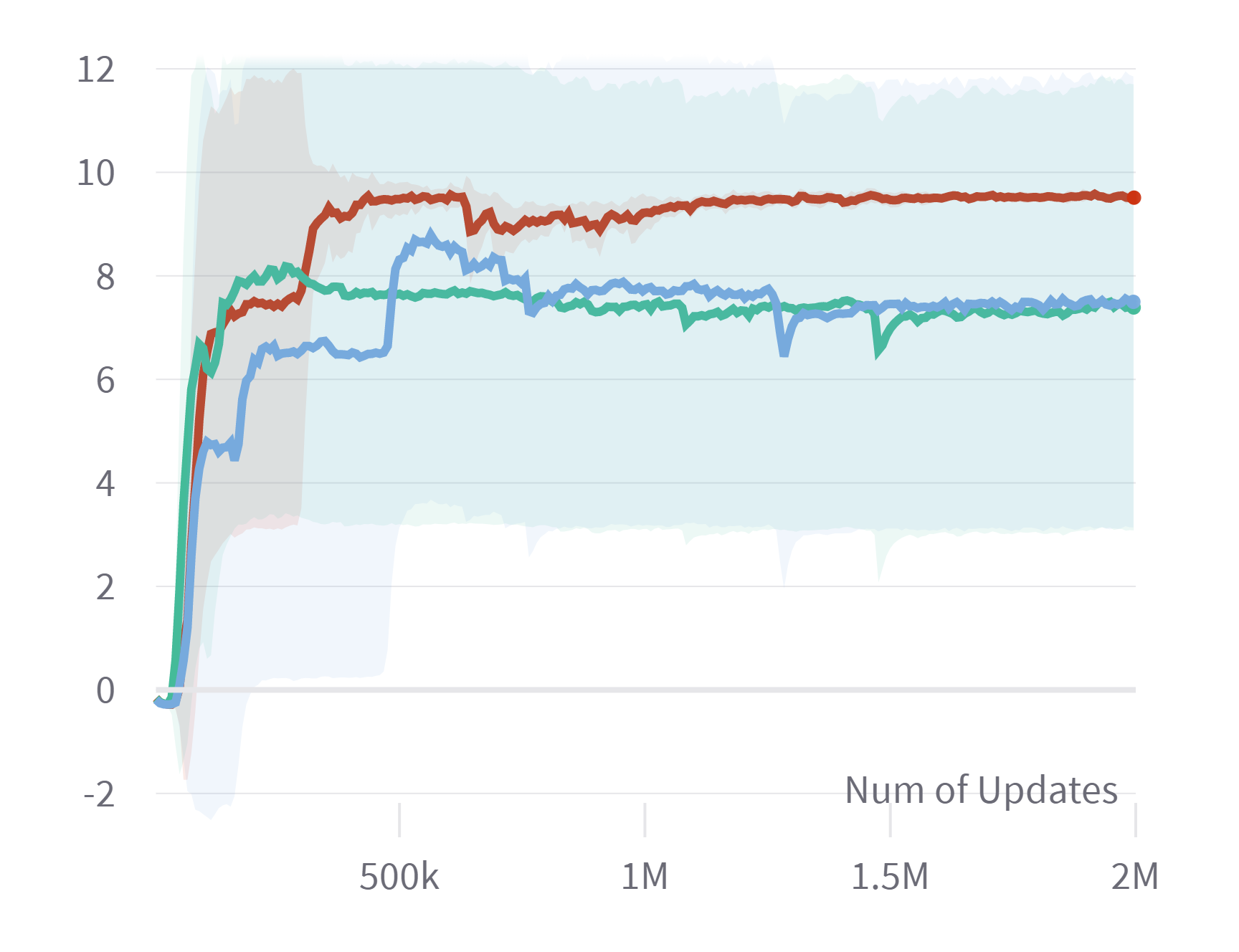}}
    \centering
    \subfigure[\footnotesize
    \label{fig:value_t2_6E}Target 2]{\includegraphics[width=0.3\textwidth]{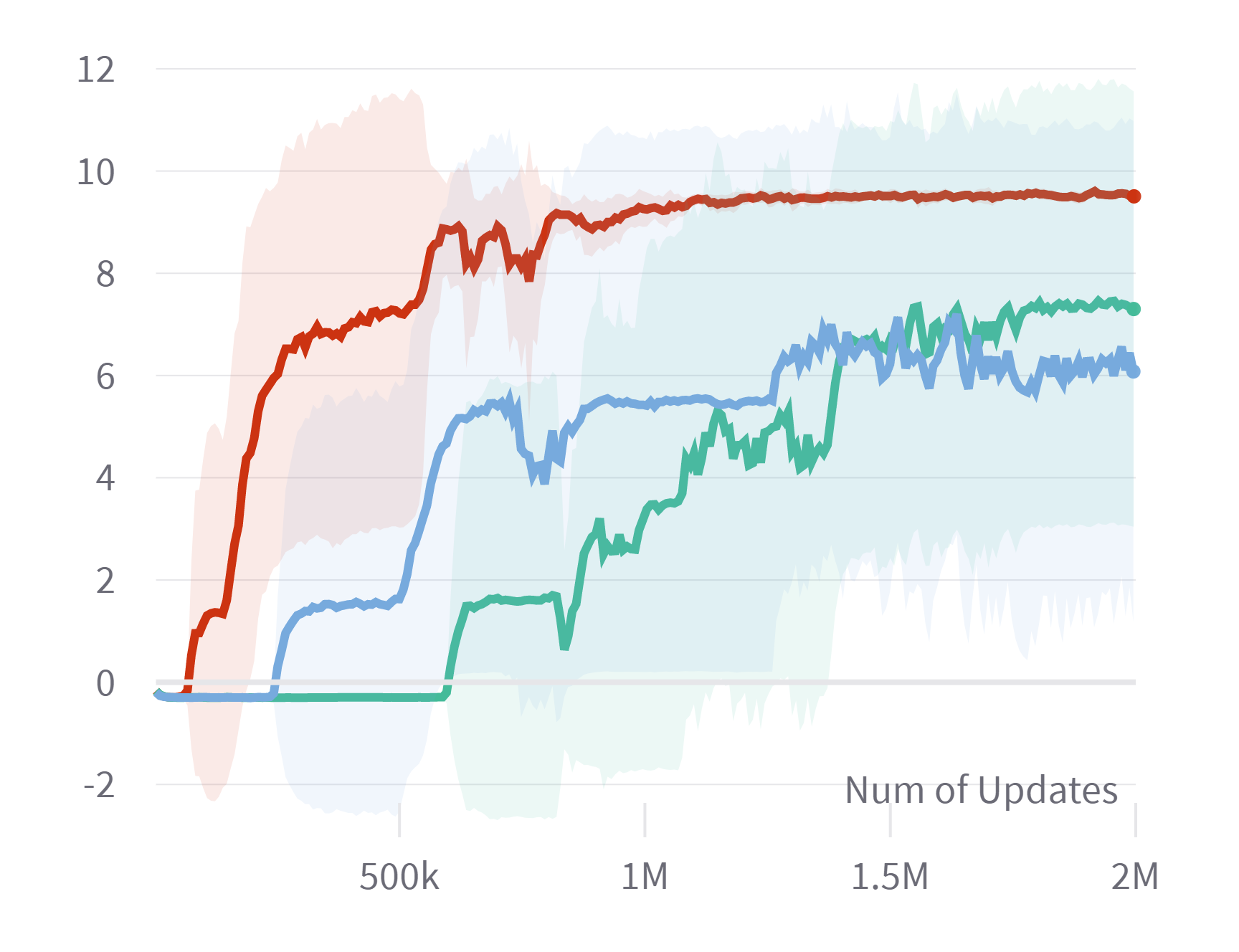}}
    \centering
    \\
    \subfigure[\footnotesize
    \label{fig:value_t3_6E}Target 3]{\includegraphics[width=0.3\textwidth]{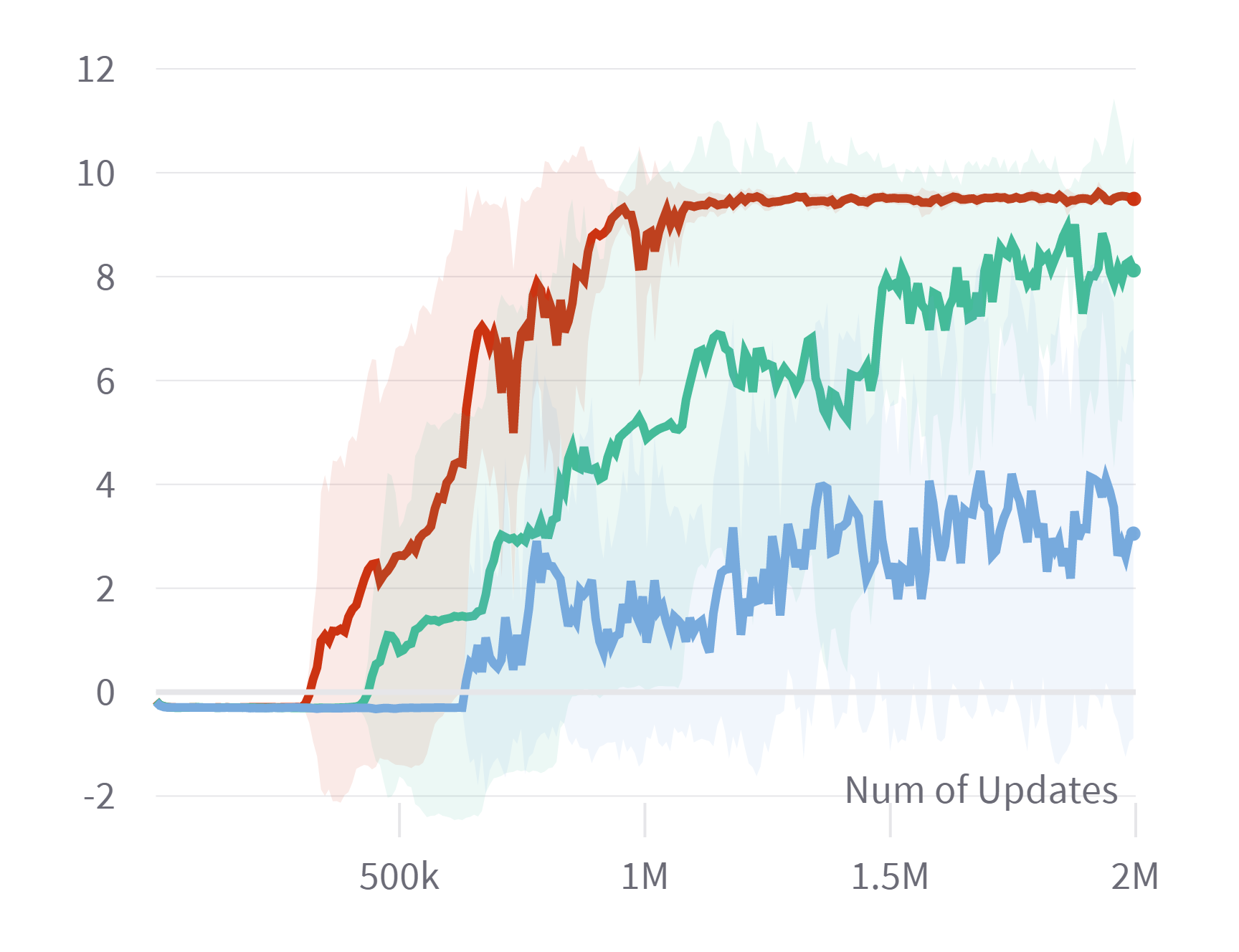}}
    \centering
    \subfigure[\footnotesize
    \label{fig:value_t4_6E}Target 4]{\includegraphics[width=0.3\textwidth]{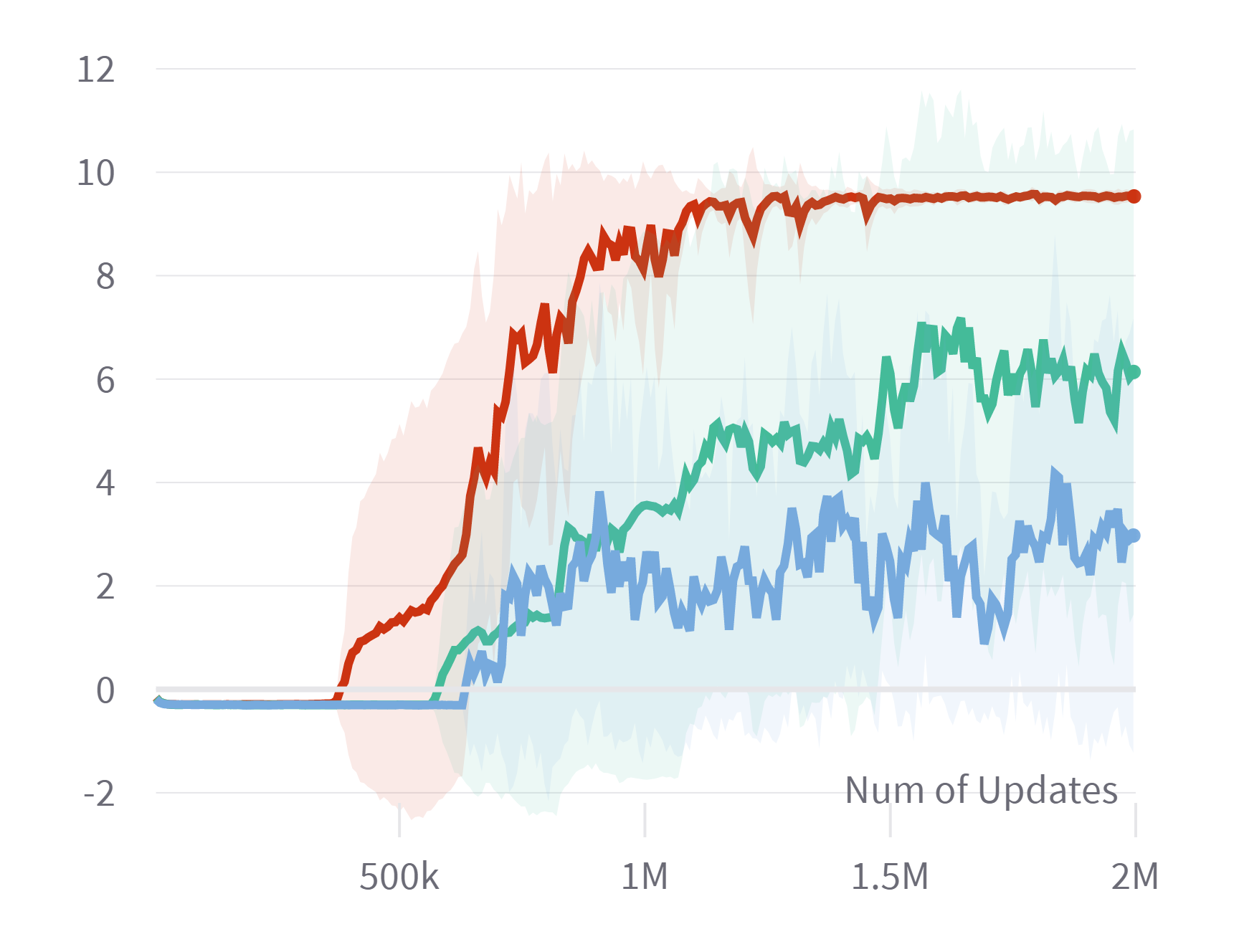}}
    \centering
    \subfigure[\footnotesize
    \label{fig:value_t5_6E}Target 5]{\includegraphics[width=0.3\textwidth]{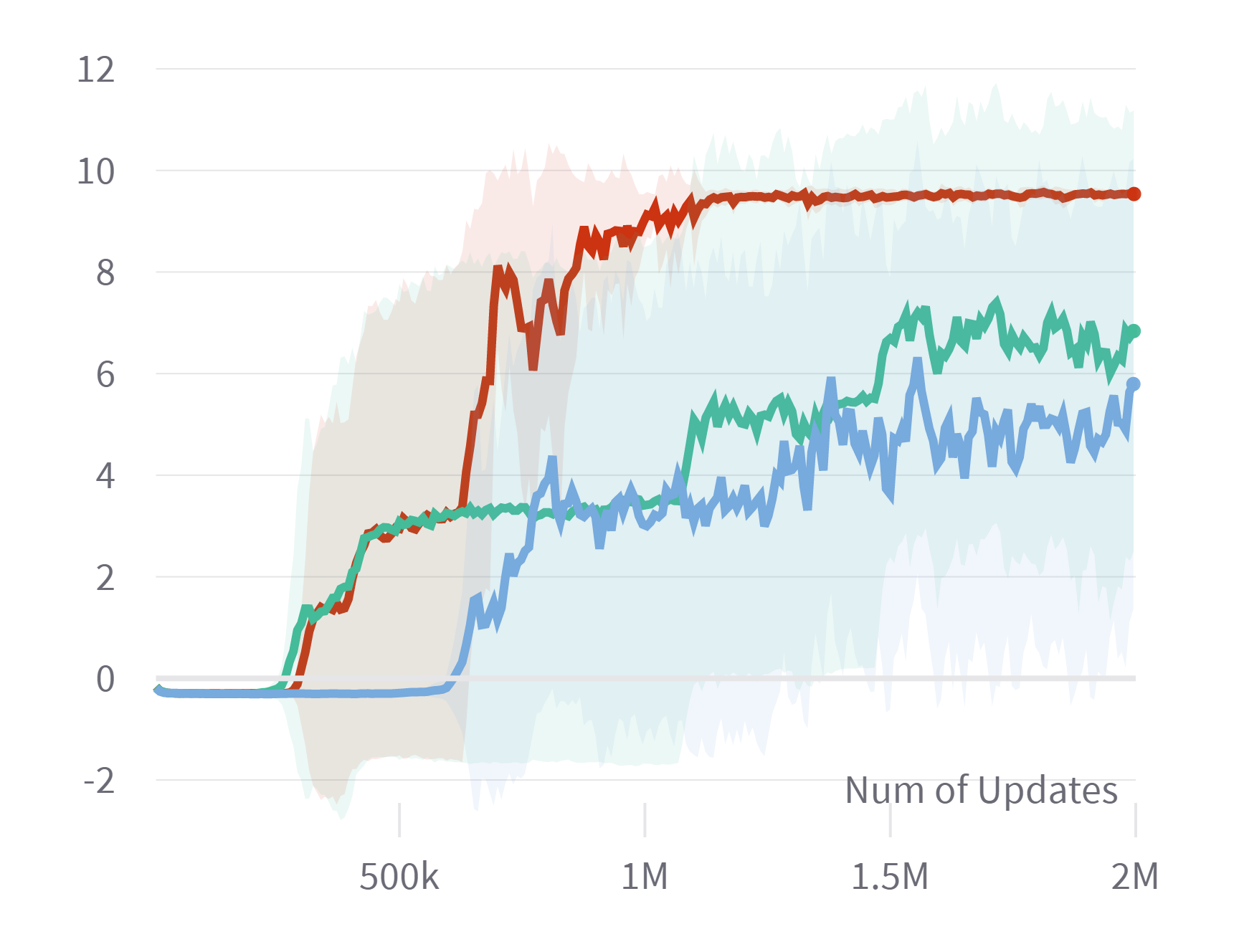}}
    \centering
    \caption{
    The value inference of each target in the Studio-6N task.
    The red line shows the highest value inference for all targets.
    L-SA framework enables accurate value inference for the collected goal states.
    Baseline models show unstable value inference curves, which slows learning.
    }
\label{fig:value_6E}
\end{figure}

\begin{figure}[hb]
\centering
    \subfigure[\footnotesize
    \label{fig:sam_stor_t0_6E}Target 0]{\includegraphics[width=0.3\textwidth]{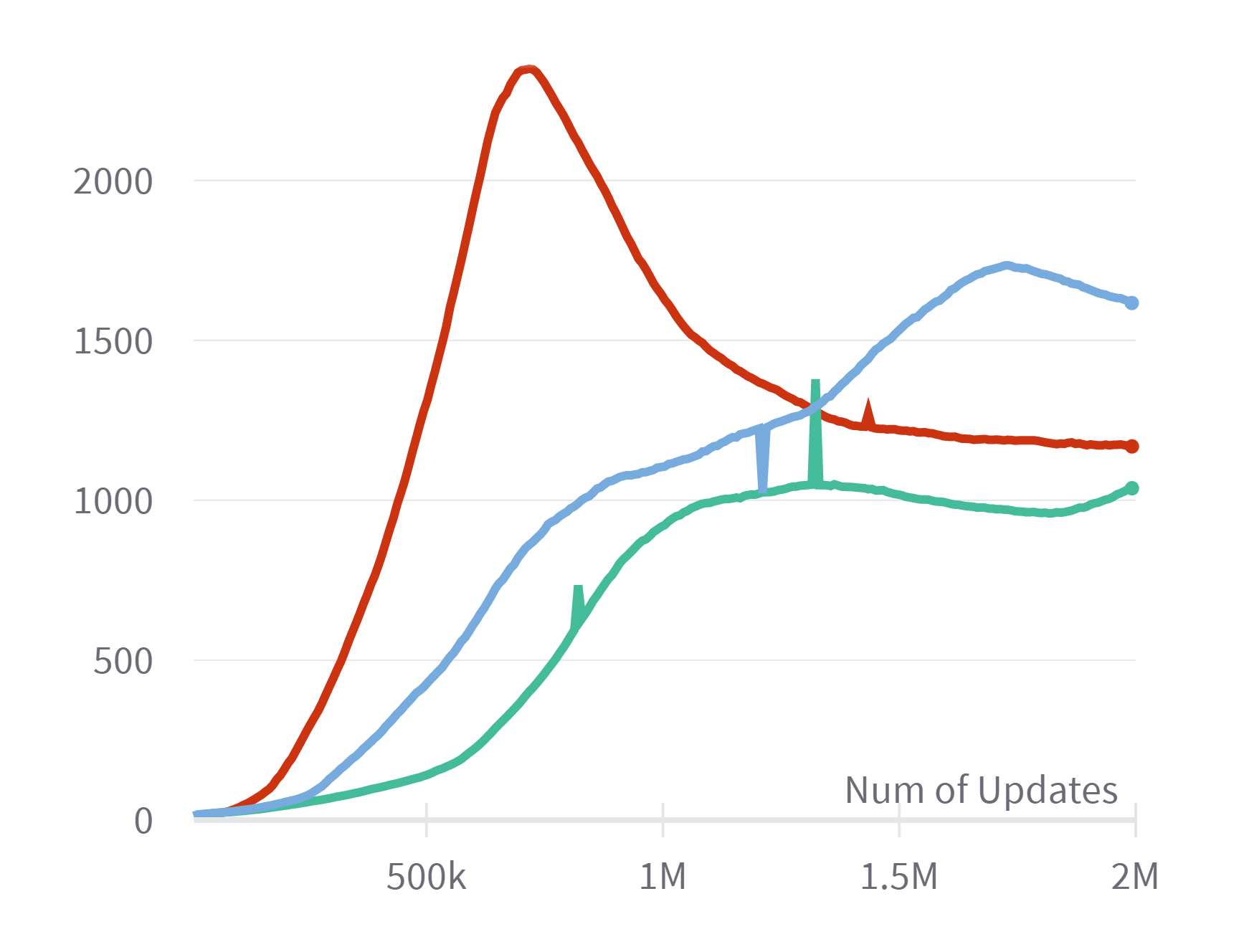}}
    \centering
    \subfigure[\footnotesize
    \label{fig:sam_stor_t1_6E}Target 1]{\includegraphics[width=0.3\textwidth]{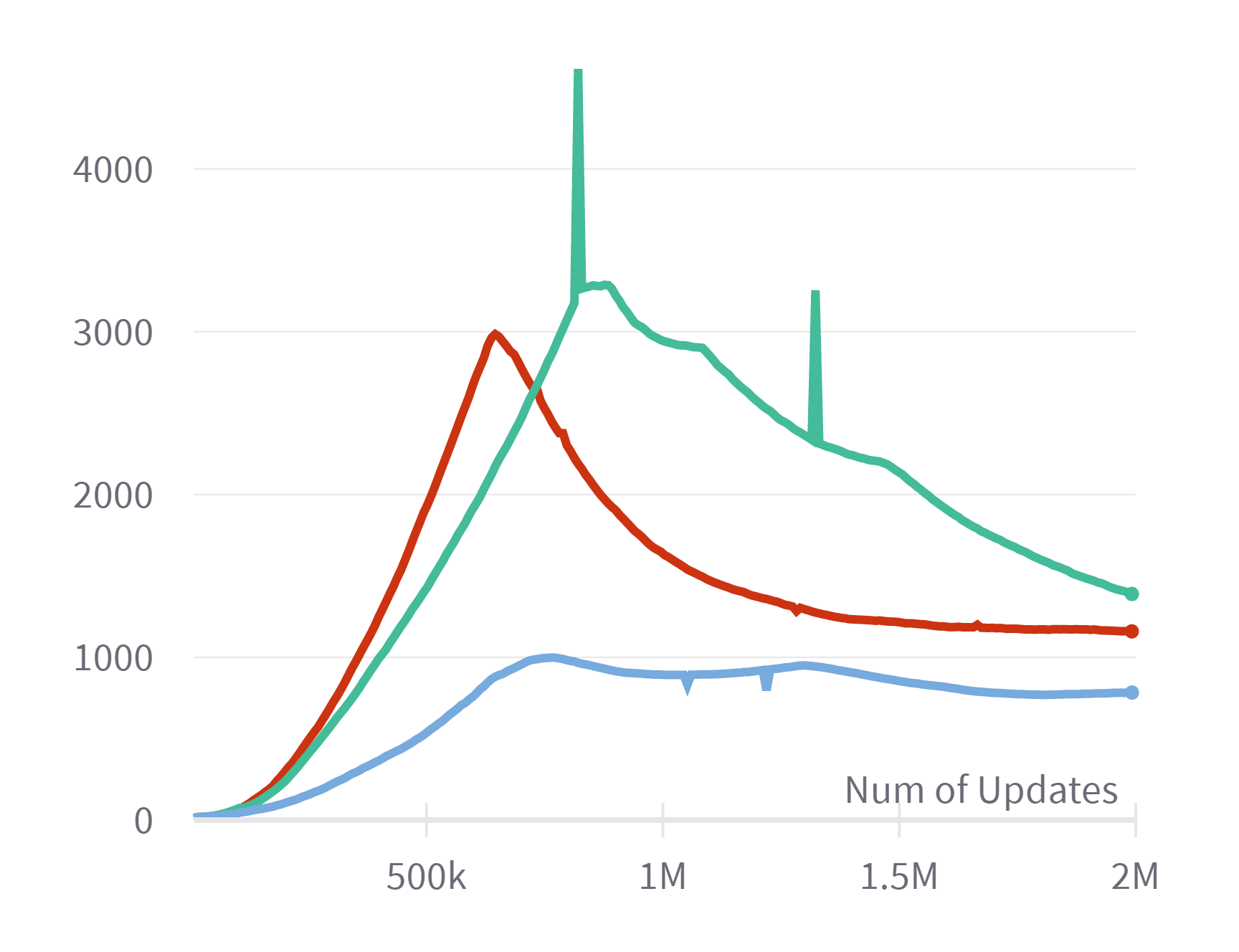}}
    \centering
    \subfigure[\footnotesize
    \label{fig:sam_stor_t2_6E}Target 2]{\includegraphics[width=0.3\textwidth]{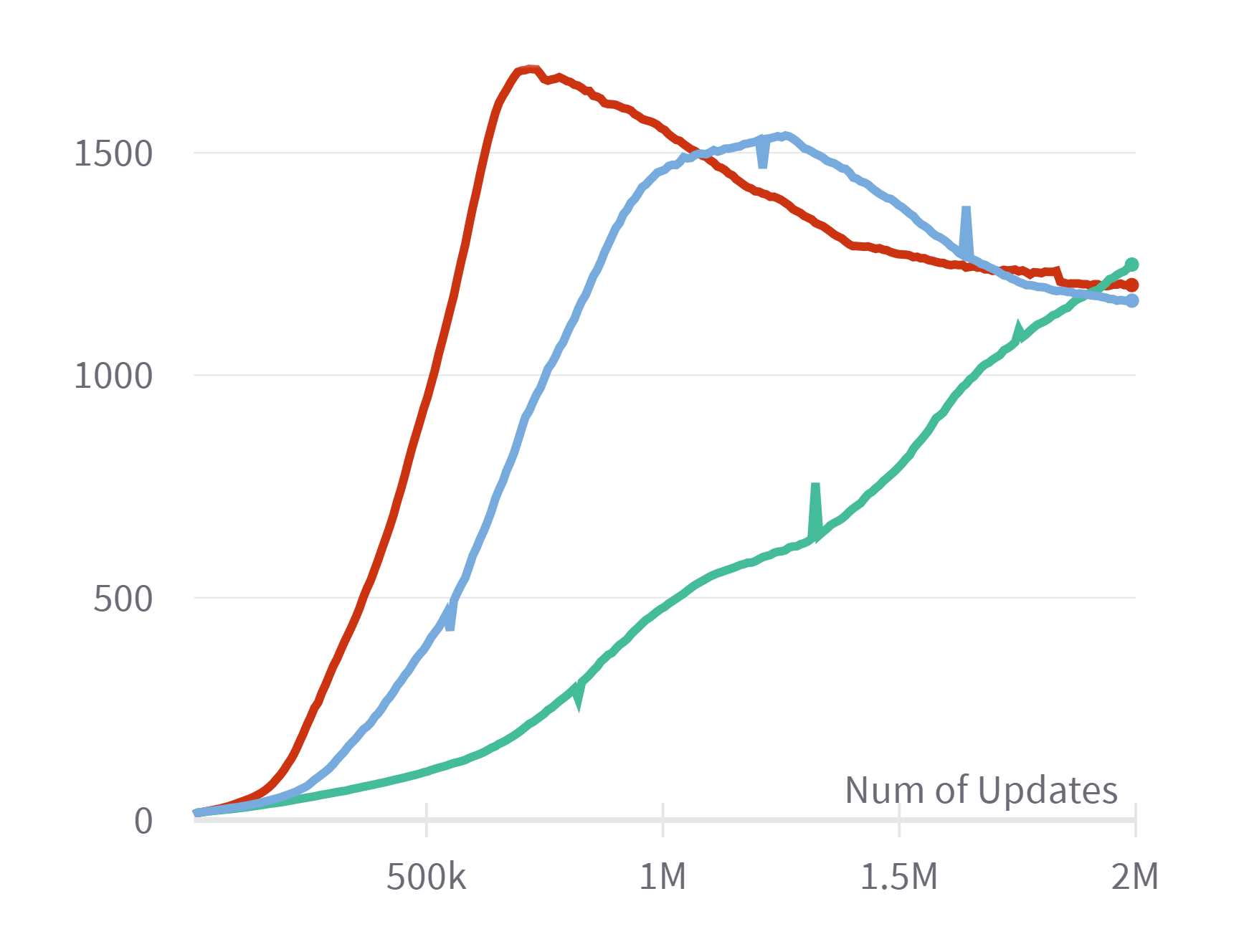}}
    \centering
    \\
    \subfigure[\footnotesize
    \label{fig:sam_stor_t3_6E}Target 3]{\includegraphics[width=0.3\textwidth]{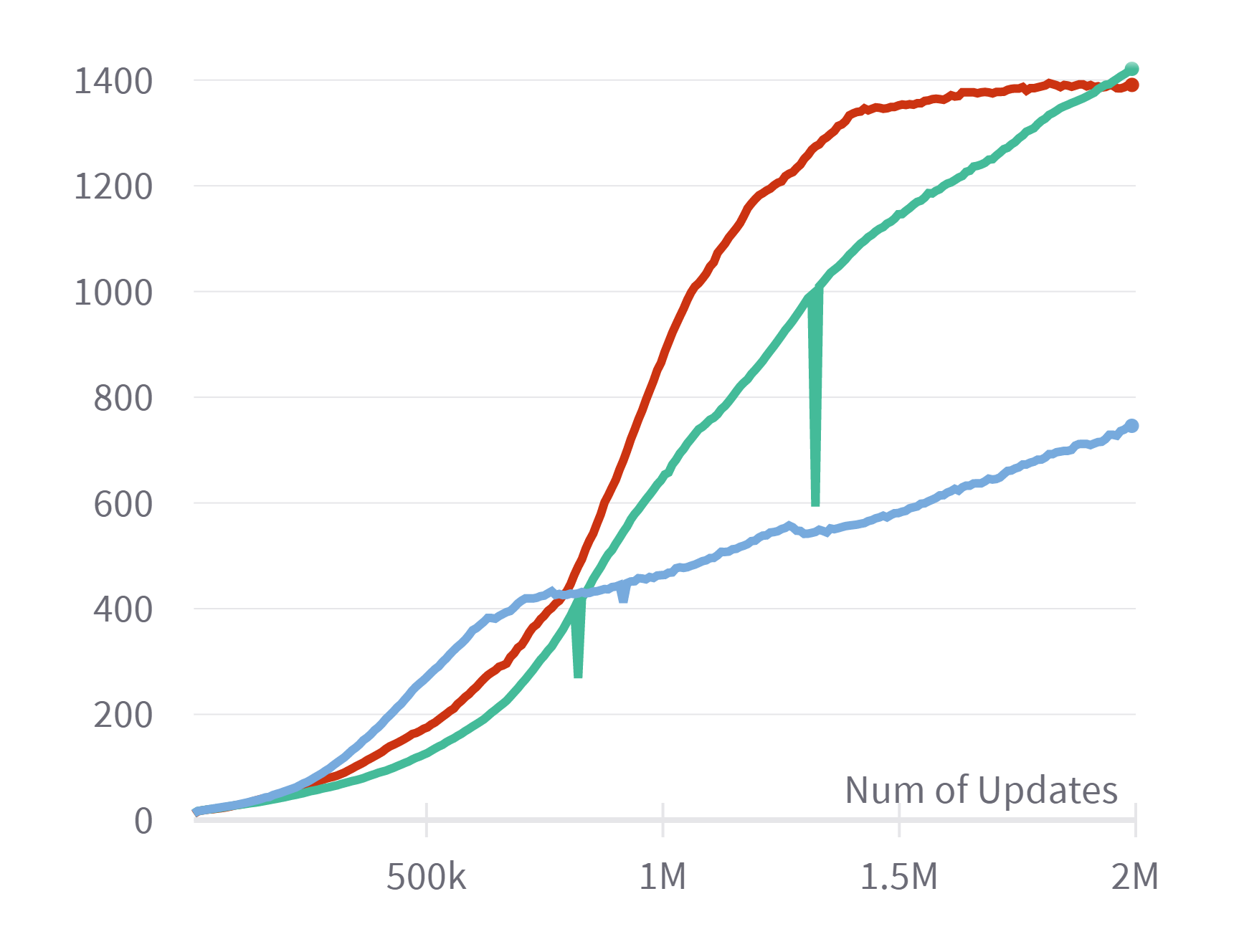}}
    \centering
    \subfigure[\footnotesize
    \label{fig:sam_stor_t4_6E}Target 4]{\includegraphics[width=0.3\textwidth]{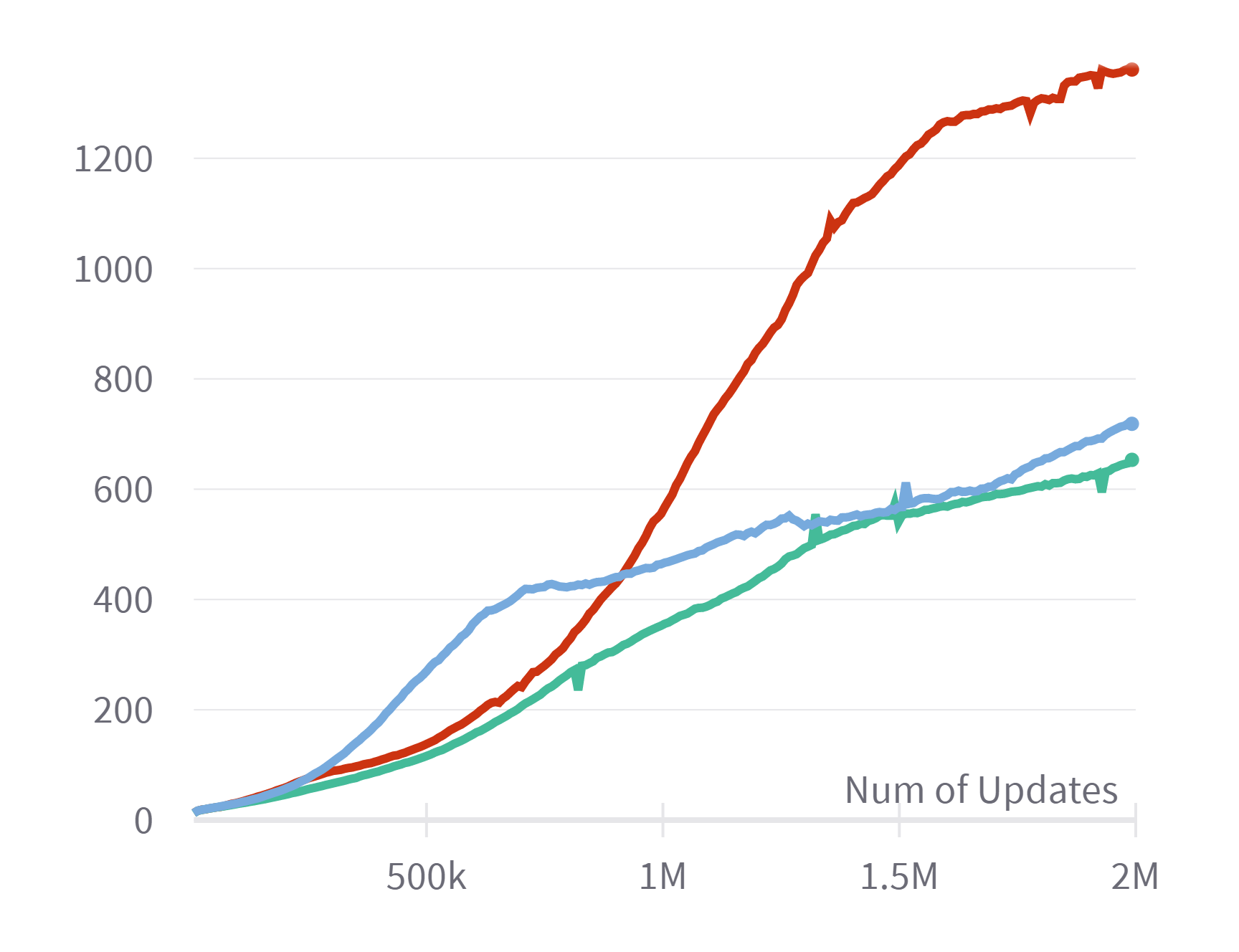}}
    \centering
    \subfigure[\footnotesize
    \label{fig:sam_stor_t5_6E}Target 5]{\includegraphics[width=0.3\textwidth]{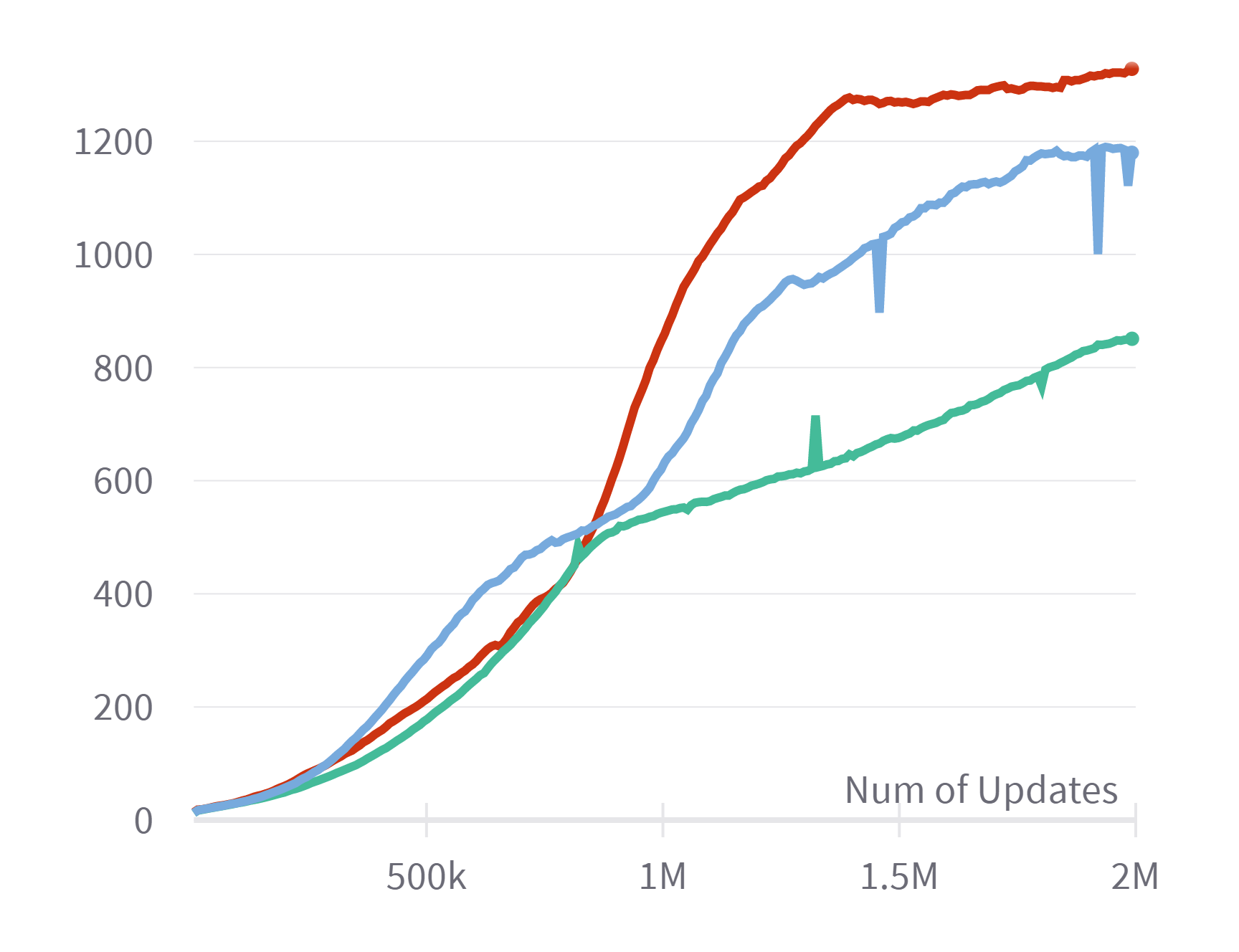}}
    \centering
    \caption{
    The sample richness of each target in the Studio-6N task.
    The higher the richness, the better, but this may also be caused by biased data collection or small amount of stored data.
    The red line rises steeply for targets 0, 1, and 2 and then descends.
    At that time, the red line rises steeply for targets 3, 4, and 5.
    Blue and green lines do not improve in the balance until the later part of the learning, compared to the red line.
    }
\label{fig:acc_sam_6E}
\end{figure}

\subsection{Additional Analyses}
We represent the success rate of each target in Figure~\ref{fig:each_target}, the cumulative query rate in Figure~\ref{fig:each_query}, the cumulative sampling rate in Figure~\ref{fig:each_sample} and the storage rate in Figure~\ref{fig:each_stor} in the Studio-2N 2H task.

For all baselines in Figure~\ref{fig:each_target}, the learning for the normal-difficulty targets is performed smoothly. 
On the other hand, only our framework (red line) and A5C (green line) show an improvement in the learning curve of hard-difficulty targets.
In particular, our method learns and saturates faster for hard-difficulty targets than A5C.
As shown by the cumulative active query rate in Figure~\ref{fig:each_query}, our framework and A5C mainly query hard-difficulty targets, but L-SA ends at the most balanced rate.
Rather, the GDAN with random query (blue line) does not learn well for hard-difficulty targets that require more experience.

The cumulative sampling rate in Figure~\ref{fig:each_sample} shows that A5C samples hard-difficulty targets at a high rate.
There are few hard-difficulty targets in the storage rate in Figure~\ref{fig:each_stor}, which leads to inefficient learning due to redundant sampling.
On the other hand, L-SA samples the normal-difficulty targets at a high rate with a high storage rate at the beginning of learning, and then the sampling rates of the hard-difficulty targets increase.
Through the virtuous cycle of the L-SA framework, the high richness of samples improves learning performance and efficiency.

Using the Studio-6N, we show the success rate for each of six normal-difficulty targets in Figure~\ref{fig:each_target_6e}, the value inference in Figure~\ref{fig:value_6E} and the richness in Figure~\ref{fig:acc_sam_6E}.
In Figure~\ref{fig:each_target_6e} and \ref{fig:value_6E}, only the L-SA framework (red line) robustly learns for all targets.
The blue line (GDAN) indicates that learning is rarely done for targets 3 and 4, and even though all targets have identical difficulty settings, the two targets are naturally excluded from learning.
The success rate for A5C (green line) is highly distributed across all targets, which indicates that A5C does not properly schedule learning.
 
Finally, we consider the sample richness for each target in the Studio-6N task, shown in Figure~\ref{fig:acc_sam_6E}.
The red line (L-SA) rises and falls for targets 0, 1, and 2, and as it descends, the richness rises steeply for targets 3, 4, and 5. This shows the order in which the targets are learned.
For the green line (A5C), the richness for target 1 is excessively high, and for other targets, the richness is low. It rises late for targets 2 and 3, and learning is not carried out sequentially.
For GDAN (blue line), as the richness rises and falls for targets 1 and 2, the richness for other targets rises slightly.
Likewise, learning does not take place sequentially, and learning is expected to become insufficient.


\begin{table}[t]
\caption{Experiment results for Studio-2N 2H and Maze-2N 2H where SEI is compared to A3C.}
\vskip 0.1in
\centering
\begin{tabular}{l||c|c|c!{\vrule width 1.5pt}c|c|c}
\toprule[1pt]
 & \multicolumn{3}{c!{\vrule width 1.5pt}}{Studio-2N 2H} & \multicolumn{3}{c}{Maze-2N 2H} \\
Algorithm & SR (\%) & \# of Updates &  SEI (\%) & SR (\%) & \# of Updates &  SEI (\%) \\
\midrule
\midrule
A3C                         &   45.0 $\pm$ 1.8 & 1.96M & 100 & 32.5 $\pm$ 3.5& 1.99M & 100\\
+NGU                        & 45.0 $\pm$ 2.2 & 1.72M & 113.95 & 28.6 $\pm$ 2.2& 1.97M & -\\
+GDAN                       & 47.1 $\pm$ 2.6 & 1.67M & 117.37 & 48.9 $\pm$ 9.5& 1.31M & 151.91\\
+A5C                        & 85.4 $\pm$ 1.8  & 0.88M  & 222.73  & 61.0$\pm$ 1.8& 0.61M & 326.23\\
+\textbf{L-SA} (ours)       & $\mathbf{86.6 \pm 1.5}$ & 0.4M  & $\mathbf{490}$ & $\mathbf{62.3 \pm 12.4}$& 0.40M & $\mathbf{497.50}$\\
\bottomrule[1pt]
\end{tabular}
\label{tab:results-2E2H}
\end{table}

\begin{table}[t]
\caption{Experiment results for Studio-3N 3H and Studio-8N where SEI is compared to GDAN.}
\vskip 0.1in
\centering
\begin{tabular}{l||c|c|c!{\vrule width 1.5pt}c|c|c}
\toprule[1pt]
 & \multicolumn{3}{c!{\vrule width 1.5pt}}{Studio-3N 3H} & \multicolumn{3}{c}{Studio-8N} \\
Algorithm & SR (\%) & \# of Updates &  SEI (\%) & SR (\%) & \# of Updates &  SEI (\%) \\
\midrule
\midrule
A3C                         & 1.1 $\pm$ 1.3 & - & - & 0.9 $\pm$ 0.3 & - & -\\
+NGU                        & 0.3 $\pm$ 2.2 & - & - & 0.1 $\pm$ 0.1 & - & - \\
+GDAN                       & 57.7 $\pm$ 1.5 & 2.0M & 100 & 12.4 $\pm$ 2.4 & 1.99M & 100 \\
+A5C                        & 67.5 $\pm$ 7.3 & 1.56M & 128.21 & 9.2 $\pm$ 13.9  & -  & -  \\
+\textbf{L-SA} (ours)       & $\mathbf{76.8 \pm 4.1}$ & 0.67M & $\mathbf{298.51}$ & $\mathbf{19.4 \pm 3.2}$ & 0.54M  & $\mathbf{368.52}$ \\
\bottomrule[1pt]
\end{tabular}
\label{tab:results-8N}
\end{table}

\subsection{Network Architecture and Hyperparameters}
We share the implementation details such as neural network architecture and hyperparameters used in the experiments.

The agent receives a 4-frame-stack of 42$\times$42 RGB-D images as an input.
The feature extractor processes this input and outputs 256 hidden features.
The feature extractor is a 4-layered Convolutional Neural Network, and batch normalization is applied to each layer.
All convolutional layers have kernel size 3, stride 2 and padding 1, and the output dimension is 256.
Afterwards, within the A3C module, the input image and the embedded instruction is processed via gated-attention, and the resulting features is fed into a Long Short-Term Memory (LSTM) module.
Finally, the action and value are output through the policy and value function, each composed of a 2-layered Multi-Layer Perceptron (MLP).

The hyperparameters used in the experiment are recorded in Table~\ref{tab:params}.
In the table, warmup refers to the amount of goal states stored in the goal storage, collected using a random-action agent before learning.
The source code we used for the experiment will be released upon publication after code cleanup.

\setlength{\tabcolsep}{4pt}
\begin{table}[t]
\caption{Hyperparameters used in our experiments.}
\vskip 0.1in
\begin{center}
\begin{tabular}{p{5.5cm}|p{2.5cm}}
\hline
Parameter Name    &   Value \\ \hline
Temperature of SupCon $\tau_s$  & 0.07 \\
Temperature of Active Querying $\tau_a$     & 60    \\
Rate of Sampling $m$   & 0.7   \\
Interval of Sampling & Every 50 updates\\
Warmup & 1,000\\
Batch Size for SupCon &   80  \\
SupCon Loss Coefficient $\eta$    &   0.5  \\
Discount $\gamma$    &   0.99  \\
Optimizer    &   Adam \\
AMSgrad    &   True  \\
Learning Rate    &   7e-5  \\
Clip Gradient Norm   &   10.0  \\
Entropy Coefficient    &   0.01  \\
Number of Training Processes    &   20  \\
Backpropagation Through Time & End of Episode\\
Non-linearity   & ReLU\\
\hline
\end{tabular}
\end{center}
\label{tab:params}
\end{table}

\begin{figure}[b]
\begin{center}
\centerline{\includegraphics[width=0.8\columnwidth]{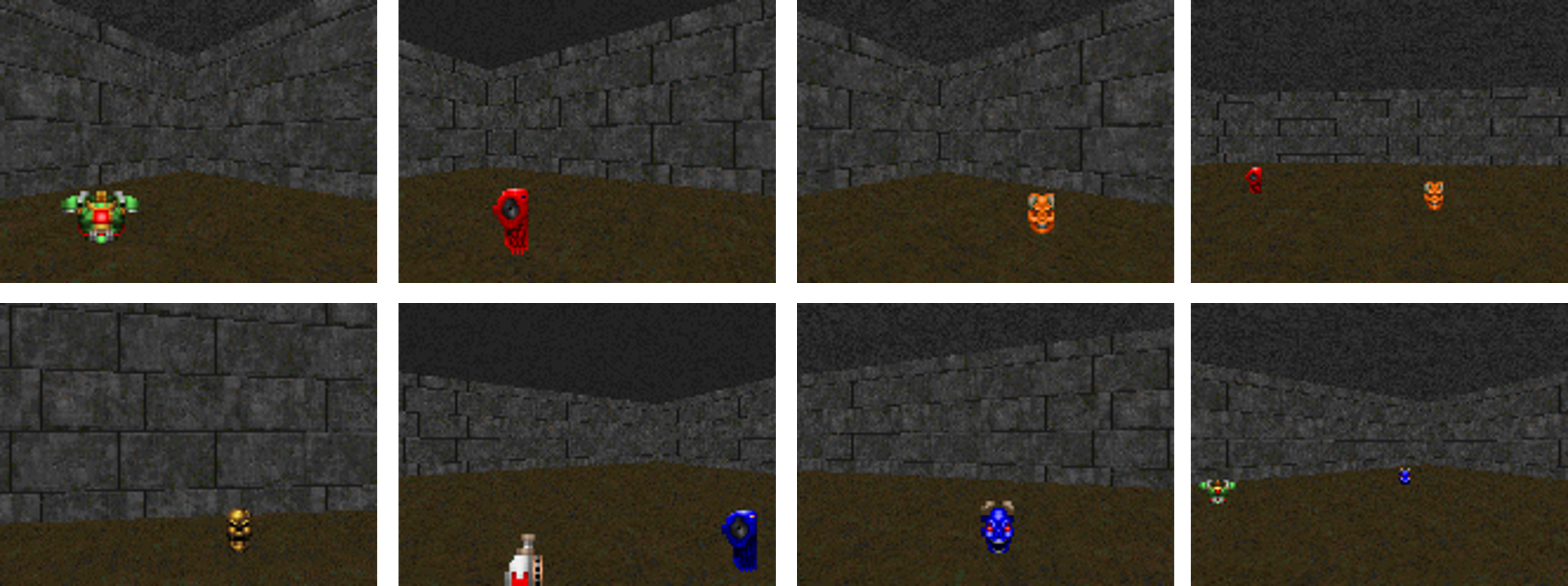}}
\caption{
    Example of visual navigation task in ViZDoom. There are the items of Armor, RedCard, Skull, HealthBonus, etc.
    }
\label{fig:app_item}
\end{center}
\end{figure}

\section{Environmental Details}
\label{app:env}
In this section, we describe the egocentric navigation tasks used for our experiments.
All of the tasks were developed using ViZDoom \cite{kempka2016vizdoom}.
The agent receives 42$\times$42 first-person perceptions of four consecutive time steps concatenated as an observation.
Likewise, every action selected by the agent is repeated for four time steps.

The four target classes used for \textbf{Maze-2N 2H}, \textbf{Studio}, and \textbf{Studio-2N 2H} maps are \{Card, Armor, Skull, Bonus\}.
The six target classes used for \textbf{Studio-6N} and \textbf{Studio-3N 3H} maps are \{RedCard, BlueCard, Armor, ChainGun, HealthBonus, ArmorBonus\}.
The eight target classes used for \textbf{Studio-8N} maps are \{RedCard, BlueCard, Armor, Skull, HealthBonus, ArmorBonus, ShotGun, ChainGun\}.
In case of four target classes, each target object may spawn as one of two variants, which differ in color or appearance.
Example views of environment and items are displayed in Figure~\ref{fig:app_item}.

The time limit $T$ is 30 for \textbf{Maze-2N 2H} map, 25 for \textbf{Studio} map, 20 for \textbf{Studio-2N 2H} map, and 23 for \textbf{Studio-6N}, \textbf{Studio-3N 3H}, and \textbf{Studio-8N} maps.
The time limit is applied after the 4-frame repeat; for instance, $T=30$ means that a total of at most $30\times 4 = 120$ in-game frames are in a single episode.
We would also like to point out that the time limit of $T=20$ for the Studio map is just barely enough for the agent to turn around to seek the instructed target and then approach it.
That is, one unnecessary step likely results in a failure of the task for that episode.

For \textbf{Studio}, \textbf{Studio-2N 2H}, and \textbf{Maze-2N 2H} maps, the map size is approximately 700 units $\times$ 700 units.
For \textbf{Studio-6N}, \textbf{Studio-3N 3H}, and \textbf{Studio-8N} maps, the map size is 1000 units $\times$ 1000 units.
The normal-difficulty and hard-difficulty targets differ according to the distances between their spawn points and the agent's initial spawn position (which is roughly the center of the map).
To be specific, hard-difficulty targets are spawned at a distance of at least 450 units away from the agent's initial spawn position for all the maps.
Normal-difficulty targets, on the other hand, are spawned at any distance.

The agent gains a reward of 10.0 for reaching the target $x$ corresponding to the instruction $I^x$, but otherwise receives a penalty of 1.0 for reaching an incorrect target, 0.1 for not reaching any target within the time limit $T$, and 0.01 every time step.

\subsection{Computational Resources}
We record the computing resources and required time for the experiments as follows:
\begin{itemize}
    \item CPU: Intel Xeon Gold 5118 CPU @ 2.30 HGz $\times$ 2
    \item RAM: 128 GB
    \item GPU: Titan V $\times$ 4
    \item Required learning time: About 13 hours for 2M updates
\end{itemize}

\section{Algorithm}
\label{app:algo}
Algorithm~\ref{alg:l-sa} shows the application of our proposed framework, L-SA, on A3C \cite{mnih2016asynchronous}.

\begin{algorithm}[t]
\caption{L-SA with A3C}
\label{alg:l-sa}
\begin{algorithmic}
\STATE Initialize actor and critic parameters $\pi$ and $\theta$
\STATE Initialize representation parameters: $\theta_s$ 
\STATE Goal Storage: $\mathcal{D}_g \leftarrow \emptyset$
\STATE Global shared counter : $T \leftarrow 0$
\STATE Thread step count: $t \leftarrow 1$
\REPEAT
        \STATE $a \sim$ Random action
        \IF{Success}
            \STATE $\mathcal{D}_g \leftarrow \mathcal{D}_g \cup \{ (s_t, I^x) \} $
        \ENDIF
    \UNTIL $|\mathcal{D}_g| > t_{warm}$
\REPEAT
    \STATE $t_{start} \leftarrow t$
    \STATE Get state $s_t$, instruction $I^x$ = Active Querying($\mathcal{D}_g$) 
    \REPEAT
        \STATE $a_t \sim \pi (a_t|s_t,I;\theta)$
        \STATE Receive reward $r_t$ and new state $s_{t+1}$
        \IF{Success}
            \STATE $\mathcal{D}_g \leftarrow \mathcal{D}_g \cup \{ (s_t, I^x) \} $
        \ENDIF
        \STATE $t \leftarrow t+1$ 
        \STATE $T \leftarrow T+1$
    \UNTIL terminal $s_t$ \OR $t - t_{start} = t_{max}$
    \FOR{$i \in \{ t-1, ...,t_{start} \}$}
        \STATE Calculate $d\theta$, $d\phi$ with Eq.~\ref{eq:actor} and ~\ref{eq:critic}
    \ENDFOR
    \STATE $B = \text{Adaptive Sampling}(w^x)$ 
    \STATE $\theta_s \leftarrow \theta_s - \eta \nabla_{\theta_s}\mathbb{E}_{(s,I) \sim B} [\nabla_{\theta_g}\mathcal{L}_{Rep}]$ with Eq.~\ref{eq:supcon}
    \STATE Perform asynchronous update of $\pi$ using $d\pi$ and of $\theta$ using $d\theta$.
\UNTIL $T > T_{max}$

\end{algorithmic}
\end{algorithm}

\end{document}